\documentclass[10pt,twocolumn,letterpaper]{article}

\usepackage[pagenumbers]{cvpr} %
\usepackage{float}

\usepackage{url}
\usepackage{titletoc}
\definecolor{cvprblue}{rgb}{0.21,0.49,0.74}
\usepackage{tcolorbox}
\usepackage{fvextra}
\usepackage{booktabs}
\usepackage{array}
\usepackage{longtable}
\usepackage{graphicx}
\usepackage{xspace}
\usepackage{placeins}
\usepackage{bbm}
\usepackage[dvipsnames]{xcolor}
\usepackage{epigraph}
\usepackage{bm}
\usepackage{amsmath}
\usepackage{multirow}

\usepackage{tikz}
\usetikzlibrary{arrows.meta,positioning,calc,shapes,fit,backgrounds,decorations.pathreplacing}
\usepackage{fontawesome5}
\usepackage{pifont}
\definecolor{vtedit}{HTML}{8E24AA}

\newcommand{\synd}[3]{\begin{tabular}[c]{@{}r@{}}#1\\{\scriptsize\textcolor{#2}{(#3)}}\end{tabular}}

\newif\ifshowallmodels
\showallmodelsfalse

\newcommand{\ourtokens}{steering tokens}
\newcommand{\Ourtokens}{Steering tokens}
\newcommand{\vlms}{VLMs}
\newcommand{\fakeparagraph}[1]{\noindent\textbf{#1}\hspace{0.5em}}

\makeatletter
\g@addto@macro\normalsize{%
  \setlength\abovedisplayskip{2pt}%
  \setlength\belowdisplayskip{2pt}%
  \setlength\abovedisplayshortskip{2pt}%
  \setlength\belowdisplayshortskip{2pt}%
}
\DeclareRobustCommand\onedot{\futurelet\@let@token\@onedot}
\def\@onedot{\ifx\@let@token.\else.\null\fi\xspace}
\def\eg{\emph{e.g}\onedot} \def\Eg{\emph{E.g}\onedot}
\def\ie{\emph{i.e}\onedot} 
 
 \def\vs{\emph{vs}\onedot}

\newcommand{\modelemph}[1]{\texttt{#1}}

\definecolor{cvprblue}{rgb}{0.21,0.49,0.74}
\usepackage[pagebackref,breaklinks,colorlinks,allcolors=cvprblue]{hyperref}

\usepackage[capitalize]{cleveref}

\def\confName{CVPR}
\def\confYear{2027}

\title{Steering What \vlms{} Should See Beyond the Limits of Prompting}

\author{
\textrm{Paul Gavrikov}$^{1}$\quad\textrm{Arnas Uselis}$^{2}$\quad\textrm{Jovita Lukasik}$^{3}$\quad\textrm{Simon Schrodi}$^{4}$\quad\textrm{Bernard Ghanem}$^{1}$ \\[0.25em]
\small
$^{1}$KAUST\quad 
$^{2}$T\"ubingen AI Center, University of T\"ubingen\quad 
$^{3}$University of Siegen\quad 
$^{4}$University of Freiburg
}
\begin{document}
\twocolumn[{%
\renewcommand\twocolumn[1][]{#1}%
\maketitle
\begin{center}
    \captionsetup{type=figure}
    \centering
    \vspace{-1.5em}
    \resizebox{\textwidth}{!}{\begin{tikzpicture}[x=1cm,y=1cm,
  font=\scriptsize,
  >={Triangle[length=3.6pt,width=3.4pt]},
  frame/.style={draw=vtline,line width=0.4pt},
  panel/.style={fill=vtpanel,draw=vtline,line width=0.4pt,rounded corners=4pt},
  ptitle/.style={anchor=north,font=\bfseries\footnotesize,inner sep=0pt},
  fam/.style={font=\scriptsize,inner sep=0pt,text height=1.6ex,text depth=0.4ex},
  famline/.style={draw=vtline,line width=0.4pt},
  cuelab/.style={anchor=west,inner sep=0pt,font=\scriptsize,text height=1.6ex,text depth=0.3ex},
  src/.style={rounded corners=1.5pt,line width=1.2pt},
  flow/.style={->,draw=vtline,line width=0.6pt},
]
\definecolor{vtline}{HTML}{000000}
\definecolor{vtpanel}{HTML}{FFFFFF}
\definecolor{ccow}{HTML}{E66100}
\definecolor{cair}{HTML}{1F6FB4}
\definecolor{cbike}{HTML}{1A8A4A}
\def\ccow#1{\textcolor{ccow}{#1}}
\def\cair#1{\textcolor{cair}{#1}}
\def\cbike#1{\textcolor{cbike}{#1}}
\definecolor{cvlm}{HTML}{2F6F73}
\definecolor{cllm}{HTML}{B5567A}
\definecolor{ctoka}{HTML}{7E57C2}%
\definecolor{ctokb}{HTML}{E0457B}
\def\ts{1.7}\def\tpitch{2.0}%
\def\stimtile#1#2#3#4{%
  \node[inner sep=0pt,outer sep=0pt,anchor=south west] (#1) at (#2,#3)
       {\includegraphics[width=\ts cm,height=\ts cm]{#4}};
  \draw[frame] (#1.south west) rectangle (#1.north east);}
\def\vsm{\textit{vs.}}
\def\conflict#1#2{\node[anchor=south,inner sep=0pt,font=\scriptsize,align=center] at ($(#1.north)+(0,0.05)$) {\strut#2\strut};}

\def\sw{1.1}%
\def\srcimg#1#2#3#4#5{%
  \foreach \d in {0.14,0.07}{%
    \draw[fill=white,draw=vtline,line width=0.4pt] (0.6+\d,#2+\d) rectangle ++(\sw,\sw);}
  \node[inner sep=0pt,outer sep=0pt,anchor=south west] (#1) at (0.6,#2)
       {\includegraphics[width=\sw cm,height=\sw cm]{#3}};
  \draw[frame] (#1.south west) rectangle (#1.north east);
  \node[rotate=90,anchor=south,font=\scriptsize,inner sep=2.5pt,text height=1.6ex,text depth=0.5ex,text=#4] at (#1.west) {#5};}
\def\yair{1.175}%
\srcimg{cow}{3.2}{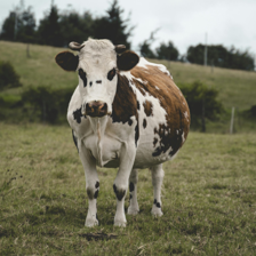}{ccow}{cow}
\srcimg{air}{\yair}{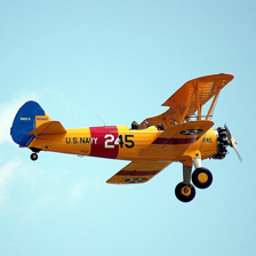}{cair}{airplane}
\srcimg{bike}{-0.85}{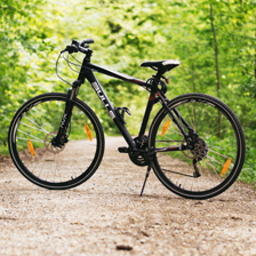}{cbike}{bicycle}
\node[anchor=north,inner sep=0pt] at ($(bike.south)+(0.07,-0.1)$) {(used for RGB only)};

\def\xa{3.0}\def\xb{\xa+\tpitch}\def\xc{\xa+2*\tpitch}\def\xd{\xa+3*\tpitch}%
\def\ya{2.2}\def\yb{-1.0}%
\stimtile{t1}{\xa}{\ya}{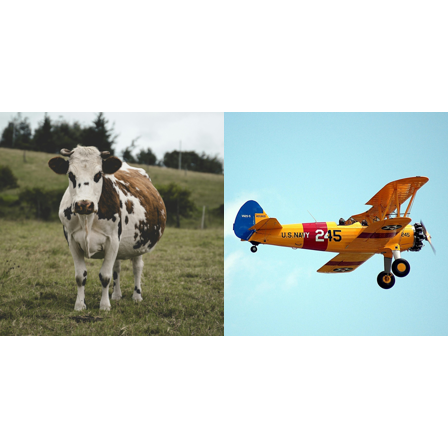}
\stimtile{t2}{\xb}{\ya}{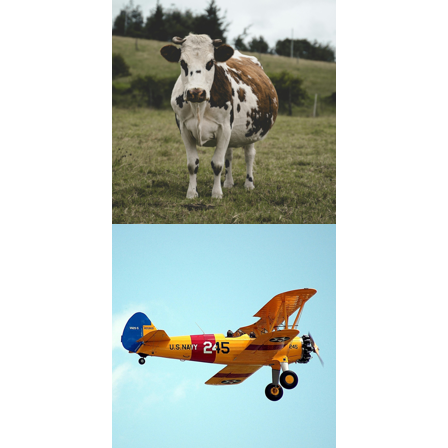}
\stimtile{t3}{\xc}{\ya}{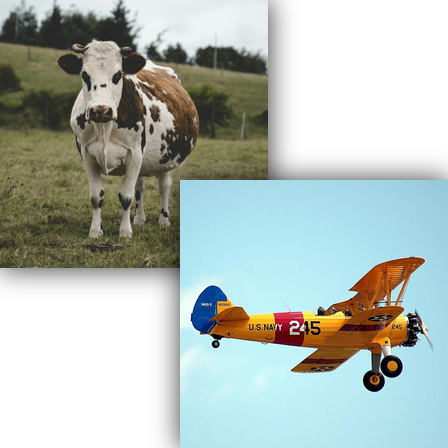}
\stimtile{t4}{\xd}{\ya}{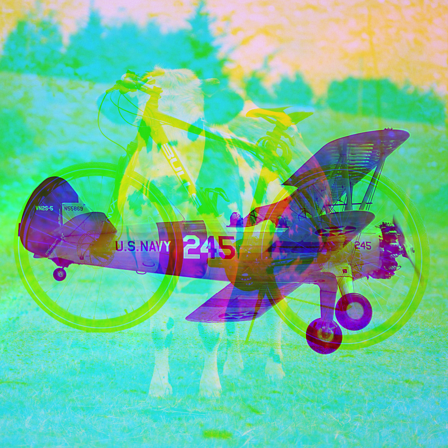}
\stimtile{t5}{\xa+0.5*\tpitch}{\yb}{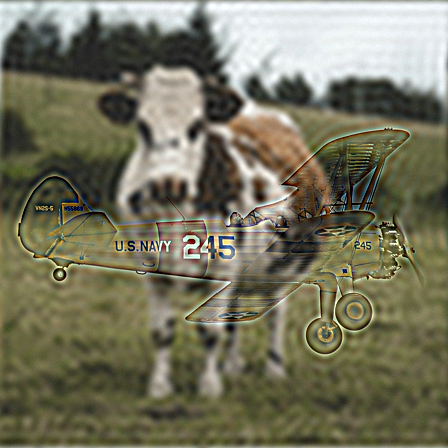}
\stimtile{t6}{\xb+0.5*\tpitch}{\yb}{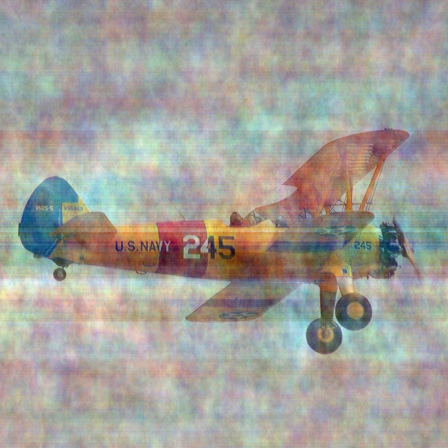}
\stimtile{t7}{\xc+0.5*\tpitch}{\yb}{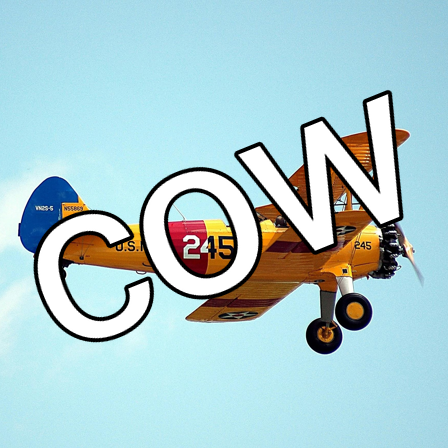}
\def\gx{2.3}\def\rx{2.75}
\node[circle,draw=vtline,line width=0.6pt,fill=white,inner sep=0pt,minimum size=0.42cm,font=\footnotesize]
  (G) at (\gx,{\yair+0.5*\sw}) {\faCog};
\draw[flow] ($(cow.east)+(0.14,0)$) -| (G.north);
\draw[flow] ($(air.east)+(0.14,0)$) -- (G.west);
\draw[flow,densely dotted] ($(bike.east)+(0.14,0)$) -| (G.south);
\draw[line width=0.6pt] (G.east) -- (\rx,{\yair+0.5*\sw});
\draw[line width=0.6pt] (\rx,\yb+0.5*\ts) -- (\rx,\ya+0.5*\ts);
\draw[flow] (\rx,\ya+0.5*\ts) -- (t1.west);
\draw[flow] (\rx,\yb+0.5*\ts) -- (t5.west);

\conflict{t1}{left \vsm\\right}
\conflict{t2}{top \vsm\\bottom}
\conflict{t3}{fore- \vsm\\background}
\conflict{t4}{red \vsm\ green\\\vsm\ blue}
\conflict{t5}{low freq.\ \vsm\\high freq.}
\conflict{t6}{amplitude \vsm\\phase}
\conflict{t7}{object \vsm\\text}

\def\nh{0.68}%
\def\famline#1#2#3{%
  \draw[famline] ($(#1.north west)+(0.08,\nh)$) -- ($(#2.north east)+(-0.08,\nh)$);
  \node[fam,anchor=south] at ($(#1.north west)!0.5!(#2.north east)+(0,\nh+0.05)$) {#3};}
\famline{t1}{t3}{\textbf{Spatial}}
\famline{t4}{t4}{\textbf{Chromatic}}
\famline{t5}{t6}{\textbf{Spectral}}
\famline{t7}{t7}{\textbf{Semantic}}

\node[anchor=south west,inner sep=0pt,font=\bfseries\footnotesize] at (0,5.2) {(a) Building forced-choice stimuli};
\tikzset{blk/.style={draw=vtline,line width=0.4pt,rounded corners=1.5pt,inner sep=2pt,
                     minimum height=0.46cm,align=left,font=\scriptsize,fill=white},
         new/.style={blk},%
         gloss/.style={anchor=north west,inner sep=0pt,font=\scriptsize\itshape},
         ans/.style={circle,draw=vtline,line width=0.6pt,inner sep=0pt,minimum size=0.5cm,font=\footnotesize},
         rlab/.style={anchor=west,inner sep=0pt,font=\scriptsize\bfseries,align=left},
         tok/.style={draw=vtline,line width=0.4pt,minimum size=0.2cm,inner sep=0pt,fill=white},
         ltok/.style={tok,top color=ctoka,bottom color=ctokb}}
\def\img#1#2{\node[inner sep=0pt,outer sep=0pt,anchor=west] (#1) at #2 {\includegraphics[width=0.64cm,height=0.64cm]{figures/stimuli/x_leftright.png}};
             \draw[frame] (#1.south west) rectangle (#1.north east);}
\def\gap{0.07}
\def\plus#1#2{\node[inner sep=0pt] (#1) at ($(#2)+(0.13,0)$) {$+$};}

\def\ox{11.4}%
\def\sx{\ox+1.45}%
\def\vlmx{\ox+6.865}%

\node[anchor=south west,inner sep=0pt,font=\bfseries\footnotesize] at (\ox,5.2) {(b) Testing biases in the VLM};
\def\y{4.32}
\def\stackbehind#1#2{\begin{pgfonlayer}{background}\foreach \d in {0.14,0.07}{%
  \draw[fill=white,draw=vtline,line width=0.4pt,#2]
    ($(#1.south west)+(\d,\d)$) rectangle ($(#1.north east)+(\d,\d)$);}\end{pgfonlayer}}

\node[inner sep=0pt,outer sep=0pt,anchor=west] (i0) at (\ox+0.05,\y) {\includegraphics[width=0.78cm,height=0.78cm]{figures/stimuli/x_rgb.png}};
\draw[frame] (i0.south west) rectangle (i0.north east);
\stackbehind{i0}{sharp corners}
\node[inner sep=0pt] (p0) at ($(i0.east)+(0.25,0)$) {$+$};
\node[blk,anchor=west,text width=3.4cm] (q0) at ($(p0.east)+(0.06,0)$)
  {$Q$\,=\,``Which option best describes\\the image? A.~\cbike{bicycle}\dots \ C.~\mbox{\cair{airplane}}\dots\ G.~\ccow{cow} \dots''};
\stackbehind{q0}{rounded corners=1.5pt}
\def\lx{\vlmx-0.68}\def\lb{\y-0.28}%
\foreach \k/\o/\h/\c in {0/A/0.14/cbike,1/B/0.07/black!35,2/C/0.3/cair,3/D/0.1/black!35,4/E/0.05/black!35,5/F/0.11/black!35,6/G/0.52/ccow}{%
  \fill[\c] (\lx+0.14*\k,\lb) rectangle ++(0.1,\h);
  \node[anchor=north,inner sep=0pt,font=\tiny] (o\o) at (\lx+0.14*\k+0.05,\lb-0.04) {\o};}
\draw[line width=0.4pt] (\lx-0.04,\lb) -- (\lx+0.98,\lb);
\coordinate (a1) at (\lx-0.08,\y);
\node[inner sep=0pt,anchor=south] at (\lx+0.47,\lb+0.6) {logits};
\node[font=\large,inner sep=1pt] (v1) at ($(q0.east)!0.5!(a1)+(0.07,0)$) {\textcolor{cvlm}{\faRobot}};
\node[inner sep=0pt,anchor=south] at ($(v1.north)+(0,0.08)$) {VLM};
\draw[flow] ($(q0.east)+(0.14,0)$) -- (v1.west);
\draw[flow] (v1.east) -- (a1);
\draw[flow] ($(oG.south)+(0,-0.04)$) -- node[anchor=east,inner sep=2pt] {argmax: G\,=\,\ccow{cow}} ++(0,-0.3) coordinate (vt);
\node[anchor=north east,inner sep=1pt] at ($(vt)+(0.3,0)$) {vote: 32.6 \% red, 47.1 \% green, 20.3 \% blue};

\def\yc{2.88}%
\node[anchor=south west,inner sep=0pt,font=\bfseries\footnotesize] at (\ox,\yc) {(c) Steering biases};

\def\y{\yc-0.98}
\node[rlab] at (\ox,\y) {optimized\\prompt};
\img{i2}{(\sx,\y)}
\plus{p2}{i2.east}
\node[new,anchor=west] (pre) at ($(p2.east)+(0.06,0)$) {prefix};
\plus{p2b}{pre.east}
\node[blk,anchor=west] (q2) at ($(p2b.east)+(0.06,0)$) {$Q$};
\node[inner sep=1pt,font=\normalsize] (llm) at ($(pre.north)+(0,0.5)$) {\textcolor{cllm}{\faBrain}};
\draw[flow] (llm.south) -- node[anchor=west,inner sep=2.5pt,font=\scriptsize\itshape] {optimizes} (pre.north);
\node[anchor=east,inner sep=1pt] at (llm.west) {LLM};
\node[inner sep=1pt,font=\small] (sv2) at (\vlmx,\y) {\textcolor{cvlm}{\faRobot}};
\draw[flow] (q2.east) -- (sv2.west);
\node[anchor=south,inner sep=1pt] (vlm) at (sv2.north) {VLM};
\draw[flow,rounded corners=2pt] (vlm.north) |- node[pos=0.75,anchor=south,inner sep=1.5pt,font=\scriptsize\itshape] {responses} (llm.east);

\def\y{\yc-2.12}
\node[rlab] at (\ox,\y) {in-context};
\node[inner sep=0pt,outer sep=0pt,anchor=west] (e3) at (\sx+0.08,\y) {\includegraphics[width=0.64cm,height=0.64cm]{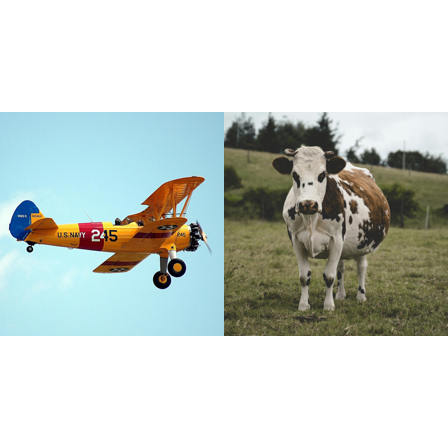}};
\draw[frame] (e3.south west) rectangle (e3.north east);
\node[new,anchor=west] (t3) at ($(e3.east)+(0.25,0)$) {``\ccow{cow}''};
\draw[flow] (e3.east) -- (t3.west);
\begin{pgfonlayer}{background}
  \node[fit=(e3)(t3),inner sep=2.5pt] (ex) {};
  \foreach \d in {0.2,0.15,0.1,0.05,0}{%
    \draw[fill=white,draw=vtline,line width=0.4pt,rounded corners=1.5pt]
      ($(ex.south west)+(\d,\d)$) rectangle ($(ex.north east)+(\d,\d)$);}
\end{pgfonlayer}
\plus{p3}{$(ex.east)+(0.2,0)$}
\node[inner sep=0pt,outer sep=0pt,anchor=west] (i3) at ($(p3.east)+(0.06,0)$) {\includegraphics[width=0.64cm,height=0.64cm]{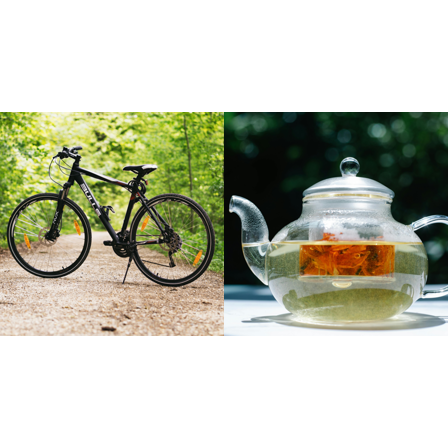}};
\draw[frame] (i3.south west) rectangle (i3.north east);
\plus{p3b}{i3.east}
\node[blk,anchor=west] (q3) at ($(p3b.east)+(0.06,0)$) {$Q$};
\node[inner sep=1pt,font=\small] (sv3) at (\vlmx,\y) {\textcolor{cvlm}{\faRobot}};
\draw[flow] (q3.east) -- (sv3.west);
\node[gloss] at ($(ex.south west)+(0,-0.07)$) {5 solved examples answered with the target cue};

\def\y{\yc-3.42}
\node[rlab] at (\ox,\y) {steering\\tokens};
\img{i4}{(\sx,\y)}
\def\tp{0.24}\def\dotp{0.2}\def\ggap{0.2}%
\tikzset{tlab/.style={inner sep=0pt,anchor=south,text height=1.6ex,text depth=0.3ex}}
\node[ltok] (la0) at ($(i4.east)+(0.48,0)$) {};
\coordinate (da) at ($(la0)+(\dotp,0)$);
\node[ltok] (la1) at ($(da)+(\dotp,0)$) {};
\foreach \k in {0,1,2}{\node[tok] (it\k) at ($(la1)+(\tp+\ggap+\tp*\k,0)$) {};}
\node[ltok] (lb0) at ($(it2)+(\tp+\ggap,0)$) {};
\coordinate (db) at ($(lb0)+(\dotp,0)$);
\node[ltok] (lb1) at ($(db)+(\dotp,0)$) {};
\foreach \c in {da,db}{\foreach \o in {-0.04,0,0.04}{\fill ($(\c)+(\o,0)$) circle (0.011);}}
\plus{p4}{$(lb1.east)+(0.04,0)$}
\node[blk,anchor=west] (q4) at ($(p4.east)+(0.06,0)$) {$Q$};
\node[inner sep=1pt,font=\small] (sv4) at (\vlmx,\y) {\textcolor{cvlm}{\faRobot}};
\draw[flow] (q4.east) -- (sv4.west);
\draw[flow] (i4.east) -- (la0.west);
\node[tlab] at ($(it1.north)+(0,0.06)$) {image};
\node[tlab] at ($(it1.north-|da)+(0,0.06)$) {\textcolor{ctoka!50!ctokb}{learned}};
\node[tlab] at ($(it1.north-|db)+(0,0.06)$) {\textcolor{ctoka!50!ctokb}{learned}};
\coordinate (bp) at ($(sv4.south)+(0,-0.3)$);
\draw[flow,rounded corners=2pt] (sv4.south) -- (bp) -| node[pos=0.25,below,inner sep=1.5pt,font=\scriptsize\itshape,align=center] {backprop tokens; VLM stays frozen} ($(it1.south-|da)+(0,-0.04)$);
\draw[flow] (bp-|db) -- ($(it1.south-|db)+(0,-0.04)$);

\end{tikzpicture}
}
    \vspace{-1.5em}
    
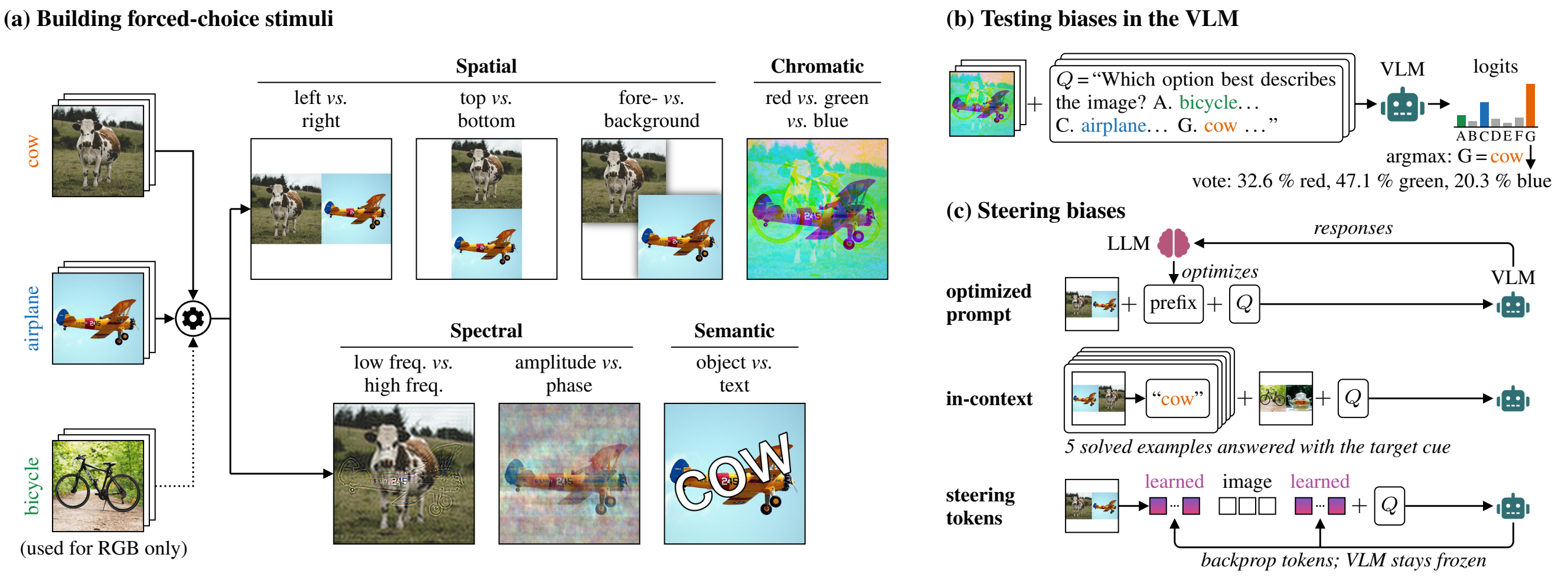
\captionof{figure}{\textbf{Measuring and steering cue biases in \vlms.} \textbf{(a)} We combine cues of different objects into a single stimulus, so that every cue supports a different object. \textbf{(b)} The model answers a forced-choice question. The picked object's cue reveals its bias, and the option logits reveal how fairly it weighs the cues. \textbf{(c)} We steer the bias toward a target cue with prompts, in-context examples, or learned \ourtokens.}
    \vspace{-0.8em}
    \label{fig:teaser}
\end{center}%
\vspace{1em}
}]
\begin{abstract}
Models inevitably pick up biases from their training, which may be misaligned with the task at hand at test time. Modern vision models are increasingly multi-modal and offer prompting potentially as a natural way to steer such biases. We investigate the opportunities and limits of this approach.
We build a controlled behavioral test bed of seven cue conflicts across spatial, chromatic, spectral, and semantic features, in which each cue supports a different class. Across open-weight \vlms, biases vary widely and often differ from those of human observers.
Targeted prompts indeed steer some biases almost perfectly in either direction, \eg, the bias toward overlaid text to over $98\%$, while failing on others. In-context learning is almost never effective.
Probes show that several of the cues that resist prompting are only weakly decodable from the visual tokens, which creates a bottleneck for steering through prompting.
To steer beyond it, we propose learnable \textit{\ourtokens} injected into the frozen vision encoder. They steer foreground--background and spatial-frequency biases to $87$--$99\%$, where prompting reaches at most $68\%$ on the same models.\\[0.27em]
\noindent\textbf{Project:} \url{paulgavrikov.github.io/cuelimits}
\end{abstract}

\section{Introduction}\label{sec:intro}

Models trained at scale learn representations that serve a wide range of downstream tasks \cite{bengio2013representation}, but they inevitably also pick up biases about which features of the input to rely on \cite{hermann2020what,lampinen2024learned}. Such biases are neither good nor bad per se: a shape bias can improve robustness \cite{geirhos2019imagenet}, reading text in an image is desired for document understanding but enables typographic attacks \cite{Goh2021Mar}, and matching human perception can be a goal in itself \cite{geirhos2018generalisation,geirhos2021partial}. Whether a bias is desirable thus depends on the task, and the task may change at test time while the model does not.

Many of these biases are well documented. ImageNet-trained CNNs prefer texture over shape \cite{geirhos2019imagenet} and rely on a broader band of spatial frequencies than humans \cite{subramanian2023spatial}. Multi-modal models such as CLIP are more shape-biased \cite{geirhos2021partial}, but readily follow text rendered into the image \cite{Goh2021Mar}. Vision-language models (\vlms{}) favor text over conflicting visual evidence \cite{deng2025words,wang2025textspeaksloudervision}, describe images from left to right \cite{chaudhary2025investigatingspatialattentionbias}, and miss shapes that humans readily see \cite{hemmat2024hidden}.

For unimodal models, changing such a bias meant retraining with different data, augmentations, or objectives \cite{geirhos2019imagenet,gavrikov024how}, which is prohibitively costly for today's foundation models. \vlms{}, in contrast, follow natural-language instructions \cite{zhou2023ifeval,lamb2025focus}, offering a way to steer biases at test time for free. However, the evidence of such steerability is mixed: prompts shift the shape bias of \vlms{}, but only within a limited range \cite{gavrikov2024can}, rewording a question changes how \vlms{} interpret bistable images \cite{panagopoulou2024evaluating}, and in-context examples do not improve abstract shape recognition \cite{hemmat2024hidden}. It remains unclear for which biases prompting works, where its limits lie, and why.

In this work, we study the opportunities and limits of test-time steering systematically. Inspired by well-known biases of human vision \cite{nuthmann2014timecourse,nuthmann2023pseudoneglect,Fernandes2021foregroundbias,tchernikov2010color,schyns1994blobs,piotrowski1982demonstration,oppenheim1981importance,tadmor1993both,guyader2004image,morgan1991relative,decesarei2021humans,geirhos2018generalisation,smith1980tracing,rayner1986graphemic}, we build a controlled behavioral test bed of seven cue conflicts across spatial, chromatic, spectral, and semantic features, in which every cue supports a different class (\cref{fig:teaser}), and measure the biases of several open-weight \vlms{} and of human observers on the same stimuli. We then ask whether prompting can steer these biases: for which cues is it effective, where are its limits, and are larger models more steerable? We compare prompts with targeted language to multi-modal in-context learning and use probes along the vision encoder and language model to locate where steering breaks down. Finally, we ask how to steer the biases that prompting cannot reach. Drawing on adversarial reprogramming \cite{elsayed2018adversarial}, visual prompts \cite{bahng2022exploring,jia2022vpt}, and soft prompts \cite{lester2021power}, we learn \textit{\ourtokens{}} that are appended to the image patches of the frozen model and steer biases well beyond the limits of prompting.

Our contributions are threefold:
\begin{itemize}
    \item We assess vision feature biases of spatial, chromatic, spectral, and semantic features in multiple \vlms{}, discovering a wide range of bias directions and strengths, a common misalignment with human vision on the same biases, and a failure to represent uncertainty in logit space (\cref{sec:experiments:default}).
    \item We show that prompt-based steering via targeted prompts can be highly effective in any direction for certain biases, while failing on others. In-context learning consistently fails to steer biases (\cref{sec:experiments:language}). Prompting failures coincide with cues that are only weakly linearly decodable from the visual tokens (\cref{sec:experiments:bottlenecks}).
    \item We train \textit{\ourtokens} that can steer prediction biases in desired directions well beyond the limits of prompting (\cref{sec:beyond}).
\end{itemize}

\section{Related work}\label{sec:related_work}

\fakeparagraph{Behavioral testing.}
A system can be characterized by how its decisions change under controlled changes of its input, without access to its internals. This behavioral view complements mechanistic interpretability \cite{bereska2024mechanistic} and has recently been argued to be necessary for AI systems that act as behavioral systems \cite{cherep2026behavioral}. A classic behavioral tool from experimental psychology is the \textit{cue conflict}, in which two sources of evidence point to different answers, so that the observer's single choice reveals which one dominates \cite{stroop1935studies,navon1977forest,schyns1994blobs,aude2006hybrid}. %
In deep learning, the paradigm became widely known through texture \vs. shape conflicts, which revealed a texture bias in ImageNet-trained CNNs \cite{geirhos2019imagenet}. It has since been carried over to \vlms, where it pits shape against texture \cite{gavrikov2024can}, written text against depicted objects \cite{Goh2021Mar,deng2025words}, or the characters of ASCII art against the picture they form \cite{wang2025textspeaksloudervision}. We apply the paradigm to a broad set of spatial, chromatic, spectral, and semantic cues, counterbalance classes across cues, record the full answer distribution in addition to the top-1 choice, and compare all models to human observers on the same stimuli. Beyond default preferences, we use the paradigm to measure how far each bias can be steered.

\fakeparagraph{Perception biases.}
Behavioral testing has exposed a range of perceptual biases in \vlms. \vlms{} are more shape-biased than their vision encoders \cite{gavrikov2024can}. On bistable images, most \vlms{} commit to one interpretation, and rewording the prompt shifts this interpretation more than manipulating the image does \cite{panagopoulou2024evaluating}; their reports also differ from those of humans \cite{takatsuki2026bistable}. \vlms{} largely miss abstract shapes formed by the arrangement of scene elements \cite{hemmat2024hidden}, and visual anagrams \cite{geng2024visualanagrams}, which keep the local parts of an image but rearrange them globally, reveal large differences in holistic shape processing across models \cite{doshi2025visual}. When text and image disagree, \vlms{} tend to follow the text \cite{deng2025words,Goh2021Mar,cheng2025unveiling,wang2025textspeaksloudervision}, and irrelevant context distracts both LLMs \cite{shi2023llmdist} and \vlms{} \cite{liu2025robustness}. Interestingly, \vlms{} describe two side-by-side images almost always from left to right, a preference that persists in a model fine-tuned on a right-to-left language \cite{chaudhary2025investigatingspatialattentionbias}.
Language offers a natural lever on such biases: prompts shift the shape bias of \vlms{} \cite{gavrikov2024can}, ``sensory words'' align the representations of a text-only LLM toward those of vision and audio encoders \cite{wang2025wordsmakelanguagemodels}, and LLMs can be fine-tuned to follow instructions about which features to rely on \cite{lamb2025focus}. In-context examples, in contrast, do not improve abstract shape recognition \cite{hemmat2024hidden}.
However, prompting can only select among features that reach the language model. Which features a network represents depends on its data, architecture, and training objective; easily decodable features can suppress harder but more predictive ones \cite{hermann2020what,hermann2024foundations}, and learned representations over-weight simple and early-learned features \cite{lampinen2024learned}. In \vlms, this bottleneck has been located both in the vision encoder, which maps visually distinct images to near-identical embeddings \cite{tong2024eyes}, and in the language model, which fails to use information that probes on the visual features recover \cite{zhang2024why,fu2025hidden}. Model interventions can bypass these shortcuts, \eg, by redirecting a small set of attention heads to the image region that should be described \cite{gandikota2026gazeheadsvlmslook}.
We extend these works from individual biases to a common test bed of seven conflicts, test for each cue whether prompting and in-context learning can steer it, and relate steering failures to the linear decodability of the cue from the visual tokens.

\fakeparagraph{Reprogramming models.} Neural networks can be ``reprogrammed'' via input-based perturbations \cite{elsayed2018adversarial}, by leveraging insights from evasion attacks through adversarial input perturbations \cite{biggio2013evasion,szegedy2014intriguing,goodfellow2015explaining}. For transformers, a learnable input token sequence changes the attention patterns of a frozen model. Prior work has studied the task adaptation of pretrained vision encoders via border- or patch-based modification of image inputs \cite{elsayed2018adversarial,bahng2022exploring}, additional learnable tokens serving as \textit{visual prompts} \cite{jia2022vpt}, or prompt-aware generation of these \cite{zhang2024prompt}, and reprogramming has been extended to black-box models with scarce data \cite{tsai20a}. In language models, the same idea appears as continuous \textit{soft prompts} that are prepended to the input embeddings \cite{lester2021power} or to the activations of every layer \cite{li2021prefix} of a frozen model. %
Our \ourtokens{} are appended to the patch sequence of the frozen vision encoder, and after projection, they act as soft prompts in the language model. In contrast to prior work, we do not use them to adapt the model to a new task, but to steer which visual cue the \vlms{} rely on while the task stays the same.

\section{Methodology}\label{sec:method}

Our methodology revolves around forced-choice experiments. Confronted with an ambiguous sample showing multiple classes in different cues, we force an observer (human or model) to make a single prediction. A repeated answer with objects from a specific cue therefore represents a preference, or \textit{bias}. In \cref{sec:method:stimuli} we introduce the source dataset and our curated bias stimuli; in \cref{sec:method:measuring}, we then introduce our metrics. An overview can be seen in \cref{fig:teaser}.

\subsection{Curating biased stimuli}\label{sec:method:stimuli}

\fakeparagraph{Source data.} To avoid confounds due to the noisy nature of existing image datasets \cite{torralba2011unbiased,recht19do,beyer2020imagenet,northcutt2021pervasive}, we carefully curate a small but controlled dataset showing common objects using images from royalty-free sources that we call \textit{MiniObjectNet}. Our dataset consists of 32 classes, which we evenly split into training and testing datasets, each containing 16 distinct classes. Each class consists of approximately 20 images, with a minimum of 12 images and a minimum resolution of $558\times 558$ px. The images in our dataset were cropped from source images to show the class object clearly centered with minimal padding. We manually filtered samples to show no other objects in the background. The selected classes are all part of the ImageNet taxonomy \cite{deng2009imagenet}. To assess potential dataset biases, we provide additional experiments with highly controlled but synthetic source images in \cref{sec:appendix:data_source}, resulting in compatible bias measurements. 

\fakeparagraph{Stimuli.} From these source images, we then generate cue-conflicted stimuli datasets targeting various biases. Each stimulus combines $N$ cues, such as the left and right half of an image or its low and high spatial frequencies, and each cue is taken from a different source image. The source images belong to $N$ distinct classes, so every cue supports a different class, and the cues are in \textit{conflict} (see \cref{fig:teaser}, \textbf{(a)}). How a cue is extracted from its source image and how the cues are combined into a single stimulus depends on the respective conflict. For every ordered combination of $N$ distinct classes, we generate $S$ stimuli from randomly drawn images of these classes, resulting in $M=S\cdot\frac{C!}{(C-N)!}$ stimuli per dataset, where $C{=}16$ is the number of classes in the respective split. Because all orderings are included, every class appears equally often in every cue, which avoids sample-specific shortcuts.

We generate the following conflicts modeling common low-level features, inspired by biases in human vision \cite{nuthmann2014timecourse,nuthmann2023pseudoneglect,Fernandes2021foregroundbias,tchernikov2010color,schyns1994blobs,piotrowski1982demonstration,oppenheim1981importance,tadmor1993both,guyader2004image,morgan1991relative,decesarei2021humans,geirhos2018generalisation,smith1980tracing,rayner1986graphemic}: 
\textbf{Spatial:} We construct three kinds of two-way spatial cue conflicts: a \emph{left \vs~right} and \emph{top \vs~bottom} conflict, where two conflicting samples are placed horizontally or vertically next to each other, while maintaining the aspect ratio. Additionally, we create a \emph{fore- \vs~background} conflict, where one image (foreground) overlaps another. We randomly position the background image in one corner and overlay the foreground on the diagonally opposite corner while adding a drop shadow. All non-image areas are filled white.
\textbf{Chromatic:} We construct a three-way cue \emph{red \vs green \vs blue (RGB)} conflict by constructing a new sample from the luminance channels of three different source samples inserted into the red, green, and blue color channels.
\textbf{Spectral:} We construct two-way hybrid images by combining all frequencies below $15\%$ of the maximum frequency along each axis from one image with all remaining frequencies from the other, followed by an inverse FFT \cite{schyns1994blobs,aude2006hybrid}. Similarly, we construct \emph{amplitude \vs~phase} conflicts by combining the Fourier amplitude spectrum of one image with the phase spectrum of the other, separately per RGB channel, followed by an inverse FFT; we keep the real part and min–max normalize each channel \cite{oppenheim1981importance,fel2024role}.
\textbf{Semantic:} We construct two-way conflicts by overlaying conflicting text onto an image \cite{smith1980tracing,rayner1986graphemic,Goh2021Mar}. We randomly vary the font, font color, and rotation of the text to increase cue variance. The font size is adjusted to fit the image with a $5 \%$ padding.

For all conflicts, we pick $S=7$, except for the chromatic one, where we use $S=3$ due to the increase in cues, resulting in a total number of stimuli $M=1{,}680$ and $M=10{,}080$, respectively. We use a resolution of $448\times 448$ px for all stimuli, which is natively supported by all tested models. 

\subsection{Measuring biases}\label{sec:method:measuring}
Given these different stimulus datasets, we aim to understand preferences of different \vlms{} 
by measuring the response rates toward specific cues in a forced-choice setting.
The models are instructed to perform zero-shot classification, following previous studies \cite{gavrikov2024can,hemmat2024hidden}. We prompt the models with a multiple-choice prompt 
providing them a shuffled set of 16 class labels in alphabetical enumerative order (\cref{sec:appendix:sens_distractor} contains an ablation of the option design). All models are instructed to only reply with the letter corresponding to their selected choice, mapping to exactly one answer token. This allows us to obtain probability estimates and reduce costs; we provide an additional ablation of the task format in \cref{sec:appendix:sens_taskformat}, showing comparable bias rates under open-ended prompting.

We generate individual responses for each image in separate chats, without batching or thinking, and utilizing greedy decoding of exactly one token. We use case-insensitive string matching of responses to map the answer. We denote by $p^{(i)}=\operatorname{softmax}(\mathbf{z}^{(i)})$ the model's probabilities over the $C$ answer options for stimulus $i$, where $\mathbf{z}^{(i)}\in\mathbb{R}^{C}$ are the next-token logits of the option letters, and by $p^{(i)}_y$ the probability of the option for class $y$.

Following previous studies \cite{geirhos2019imagenet,gavrikov2024can},
we measure the bias toward cue $c$ as the fraction of valid
responses that map to its class: \begin{equation}\label{eq:bias}
    \mathrm{Bias}_c = \frac{
        \sum_{i=1}^{M} \mathbbm{1}[\hat{y}^{(i)} = y_c^{(i)}]
    }{
        \sum_{i=1}^{M} \mathbbm{1}[\hat{y}^{(i)} \in \{y_1^{(i)}, \ldots, y_N^{(i)}\}]
    },
\end{equation}
where $y_c^{(i)}$ denotes the class shown in cue $c$ of stimulus $i$, and $\hat{y}^{(i)}$ denotes the predicted class.

Additionally, we aim to quantify the \textit{cue fairness}, \ie how evenly a model spreads its probability across the conflicting cues of an individual stimulus. Unlike the bias in \cref{eq:bias}, which only counts the top-1 answer, it tells whether a model that picks one cue was nearly undecided or fully certain. For stimulus $i$, let \(q^{(i)}=(q^{(i)}_1,\ldots,q^{(i)}_N)\) with \(q^{(i)}_c = p^{(i)}_{y^{(i)}_c} / \sum_{c'=1}^{N} p^{(i)}_{y^{(i)}_{c'}}\) denote the probabilities of its \(N\) cue classes, renormalized to sum to one. We measure the total variation (TV) distance of \(q^{(i)}\) from the uniform distribution \(u_N=(1/N,\ldots,1/N)\), normalized by its maximum value \(1-1/N\). Dropping the superscript for readability, we define
\begin{equation}\label{eq:cue_fairness}
    \mathrm{I}_{\mathrm{TV}}(q)
    =
    1-\frac{\operatorname{TV}(q,u_N)}{1-1/N}
    =
    1-\frac{N}{2(N-1)}
    \sum_{c=1}^{N}\left|q_c-\frac1N\right|.
\end{equation}
A score of 1 means that the model splits its probability evenly across all cues, and 0 means that it puts all of it on a single cue. We report the mean \(\frac1M\sum_i \mathrm{I}_{\mathrm{TV}}(q^{(i)})\) over all stimuli.
\begin{table*}[]
\centering
\caption{\textbf{Prediction bias.} Bias (\%, \cref{eq:bias}) toward each cue in our cue-conflict datasets. \emph{Neutral prompting}: default behavior under a generic VQA prompt, compared with human annotations. The cues of each conflict sum to 100\%. The remaining blocks show the bias toward each cue when that cue is targeted by \emph{explicit cue targeting} (\cref{sec:experiments:default:explicit}), \emph{in-context learning} (\cref{sec:experiments:default:icl}), or learned \ourtokens~(\cref{sec:beyond}). Each column comes from a separate run, so steered values do not have to sum to 100\%. Numbers in parentheses give the change in percentage points over the same model's neutral bias for that cue.}
\label{tab:vlm_bias}
\small
\resizebox{\linewidth}{!}{
\begin{tabular}{@{}l|rr|rr|rr|rrr|rr|rr|rr@{}}
\toprule
\multicolumn{1}{c}{}                                   & \multicolumn{6}{c}{\textbf{Spatial}}                                                                        & \multicolumn{3}{c}{\textbf{Chromatic}}                                         & \multicolumn{4}{c}{\textbf{Spectral}}                                                                                      & \multicolumn{2}{c}{\textbf{Semantic}} \\
\cmidrule(l{5pt}r{5pt}){2-7}\cmidrule(l{5pt}r{5pt}){8-10}\cmidrule(l{5pt}r{5pt}){11-14}\cmidrule(l{5pt}r{5pt}){15-16}
\multicolumn{1}{@{}l}{\textbf{Model}} & \multicolumn{1}{c}{Left} & \multicolumn{1}{c}{Right} & \multicolumn{1}{c}{Top} & \multicolumn{1}{c}{Btm.} & \multicolumn{1}{c}{Fg.} & \multicolumn{1}{c}{Bg.} & \multicolumn{1}{c}{Red} & \multicolumn{1}{c}{Grn.} & \multicolumn{1}{c}{Blue} & \multicolumn{1}{c}{LF} & \multicolumn{1}{c}{HF} & \multicolumn{1}{c}{Amp.} & \multicolumn{1}{c}{Phs.} & \multicolumn{1}{c}{Obj.} & \multicolumn{1}{c}{Text} \\
\midrule
\multicolumn{16}{c}{\textbf{Neutral prompting}}\\
\midrule

Humans & 51.2 & 48.8 & 50.5 & 49.5 & 59.5 & 40.5 & 20.0 & 71.2 & 8.8 & 79.5 & 20.5 & 2.6 & 97.4 & 67.5 & 32.5 \\

\midrule
Gemma-4-E4B & 61.9 & 38.1 & 48.9 & 51.1 & 55.3 & 44.7 & 29.1 & 60.9 & 10.0 & 42.1 & 57.9 & 0.6 & 99.4 & 89.2 & 10.8 \\
\ifshowallmodels
\quad budget 560 & 61.3 & 38.7 & 58.4 & 41.6 & 56.4 & 43.6 & 26.7 & 63.3 & 9.9 & 47.6 & 52.4 & 0.9 & 99.1 & 89.5 & 10.5 \\
\quad budget 280 & 62.9 & 37.1 & 66.3 & 33.7 & 57.3 & 42.7 & 29.9 & 60.4 & 9.7 & 60.5 & 39.5 & 0.7 & 99.3 & 89.2 & 10.8 \\
\quad budget 140 & 59.1 & 40.9 & 68.1 & 31.9 & 54.9 & 45.1 & 30.3 & 59.1 & 10.7 & 70.8 & 29.2 & 0.8 & 99.2 & 85.7 & 14.3 \\
\quad budget 70 & 55.5 & 44.5 & 56.2 & 43.8 & 55.8 & 44.2 & 29.7 & 59.2 & 11.1 & 80.5 & 19.5 & 0.9 & 99.1 & 81.3 & 18.7 \\
\fi
Gemma-4-12B & 72.2 & 27.8 & 69.1 & 30.9 & 51.9 & 48.1 & 29.1 & 57.8 & 13.1 & 44.1 & 55.9 & 0.5 & 99.5 & 93.4 & 6.6 \\
\ifshowallmodels
InternVL3.5-1B & 65.4 & 34.6 & 51.7 & 48.3 & 53.3 & 46.7 & 29.8 & 56.7 & 13.6 & 58.4 & 41.6 & 0.3 & 99.7 & 39.5 & 60.5 \\
InternVL3.5-2B & 68.5 & 31.5 & 56.9 & 43.1 & 54.0 & 46.0 & 28.9 & 56.7 & 14.4 & 56.4 & 43.6 & 0.4 & 99.6 & 30.9 & 69.1 \\
InternVL3.5-4B & 42.7 & 57.3 & 35.8 & 64.2 & 53.8 & 46.2 & 30.4 & 55.4 & 14.2 & 61.0 & 39.0 & 0.2 & 99.8 & 55.2 & 44.8 \\
\fi
InternVL3.5-8B & 47.2 & 52.8 & 34.8 & 65.2 & 54.4 & 45.6 & 30.0 & 55.3 & 14.7 & 60.4 & 39.6 & 0.3 & 99.7 & 56.7 & 43.3 \\
\ifshowallmodels
InternVL3.5-14B & 43.0 & 57.0 & 40.4 & 59.6 & 53.5 & 46.5 & 29.6 & 55.9 & 14.5 & 59.6 & 40.4 & 0.3 & 99.7 & 80.9 & 19.1 \\
InternVL3.5-38B & 51.0 & 49.0 & 61.2 & 38.8 & 56.5 & 43.5 & 28.3 & 49.3 & 22.4 & 57.6 & 42.4 & 0.1 & 99.9 & 95.2 & 4.8 \\
Qwen3.5-0.8B & 58.9 & 41.1 & 55.6 & 44.4 & 51.5 & 48.5 & 33.4 & 48.9 & 17.8 & 59.4 & 40.6 & 1.3 & 98.7 & 29.4 & 70.6 \\
Qwen3.5-2B & 50.5 & 49.5 & 55.5 & 44.5 & 54.2 & 45.8 & 31.2 & 47.9 & 20.8 & 49.7 & 50.3 & 0.1 & 99.9 & 40.6 & 59.4 \\
Qwen3.5-4B & 56.9 & 43.1 & 67.1 & 32.9 & 53.9 & 46.1 & 33.2 & 45.9 & 20.9 & 48.9 & 51.1 & 0.0 & 100.0 & 69.6 & 30.4 \\
\fi
Qwen3.5-9B & 53.3 & 46.7 & 62.3 & 37.7 & 52.9 & 47.1 & 32.6 & 47.1 & 20.3 & 55.8 & 44.2 & 0.1 & 99.9 & 88.0 & 12.0 \\
\ifshowallmodels
Qwen3.5-27B & 68.9 & 31.1 & 66.2 & 33.8 & 52.7 & 47.3 & 32.4 & 47.6 & 19.9 & 51.3 & 48.7 & 0.1 & 99.9 & 96.5 & 3.5 \\
Qwen3.6-27B & 69.9 & 30.1 & 71.8 & 28.2 & 52.7 & 47.3 & 32.4 & 49.4 & 18.3 & 56.8 & 43.2 & 0.1 & 99.9 & 99.0 & 1.0 \\
Qwen3.8-27B & 53.9 & 46.1 & 54.8 & 45.2 & 55.0 & 45.0 & 32.8 & 46.3 & 20.9 & 54.0 & 46.0 & 0.3 & 99.7 & 90.6 & 9.4 \\
\fi

\midrule
\multicolumn{16}{c}{\textbf{Explicit cue targeting}}\\
\midrule

Gemma-4-E4B & \synd{99.8}{ForestGreen}{$+$37.9} & \synd{100.0}{ForestGreen}{$+$61.9} & \synd{99.9}{ForestGreen}{$+$51.0} & \synd{100.0}{ForestGreen}{$+$48.9} & \synd{54.9}{BrickRed}{$-$0.4} & \synd{53.5}{ForestGreen}{$+$8.8} & \synd{29.6}{ForestGreen}{$+$0.5} & \synd{63.3}{ForestGreen}{$+$2.4} & \synd{12.9}{ForestGreen}{$+$2.9} & \synd{57.1}{ForestGreen}{$+$15.0} & \synd{65.3}{ForestGreen}{$+$7.4} & \synd{3.5}{ForestGreen}{$+$2.9} & \synd{99.7}{ForestGreen}{$+$0.3} & \synd{97.3}{ForestGreen}{$+$8.1} & \synd{99.6}{ForestGreen}{$+$88.8} \\
Gemma-4-12B & \synd{100.0}{ForestGreen}{$+$27.8} & \synd{100.0}{ForestGreen}{$+$72.2} & \synd{100.0}{ForestGreen}{$+$30.9} & \synd{100.0}{ForestGreen}{$+$69.1} & \synd{75.7}{ForestGreen}{$+$23.8} & \synd{68.1}{ForestGreen}{$+$20.0} & \synd{27.9}{BrickRed}{$-$1.2} & \synd{64.0}{ForestGreen}{$+$6.2} & \synd{17.9}{ForestGreen}{$+$4.8} & \synd{56.5}{ForestGreen}{$+$12.4} & \synd{66.7}{ForestGreen}{$+$10.8} & \synd{1.5}{ForestGreen}{$+$1.0} & \synd{99.5}{Gray}{$\pm$0.0} & \synd{99.0}{ForestGreen}{$+$5.6} & \synd{98.7}{ForestGreen}{$+$92.1} \\
InternVL3.5-8B & \synd{99.9}{ForestGreen}{$+$52.7} & \synd{100.0}{ForestGreen}{$+$47.2} & \synd{99.9}{ForestGreen}{$+$65.1} & \synd{100.0}{ForestGreen}{$+$34.8} & \synd{54.6}{ForestGreen}{$+$0.2} & \synd{58.7}{ForestGreen}{$+$13.1} & \synd{28.4}{BrickRed}{$-$1.6} & \synd{56.5}{ForestGreen}{$+$1.2} & \synd{14.7}{Gray}{$\pm$0.0} & \synd{67.5}{ForestGreen}{$+$7.1} & \synd{47.9}{ForestGreen}{$+$8.3} & \synd{2.0}{ForestGreen}{$+$1.7} & \synd{99.5}{BrickRed}{$-$0.2} & \synd{98.4}{ForestGreen}{$+$41.7} & \synd{98.7}{ForestGreen}{$+$55.4} \\
\ifshowallmodels
Qwen3.5-0.8B & \synd{99.7}{ForestGreen}{$+$40.8} & \synd{99.6}{ForestGreen}{$+$58.5} & \synd{99.7}{ForestGreen}{$+$44.1} & \synd{99.5}{ForestGreen}{$+$55.1} & \synd{53.1}{ForestGreen}{$+$1.6} & \synd{56.0}{ForestGreen}{$+$7.5} & \synd{32.5}{BrickRed}{$-$0.9} & \synd{49.5}{ForestGreen}{$+$0.6} & \synd{19.4}{ForestGreen}{$+$1.6} & \synd{64.1}{ForestGreen}{$+$4.7} & \synd{50.2}{ForestGreen}{$+$9.6} & \synd{3.7}{ForestGreen}{$+$2.4} & \synd{98.6}{BrickRed}{$-$0.1} & \synd{90.0}{ForestGreen}{$+$60.6} & \synd{95.0}{ForestGreen}{$+$24.4} \\
Qwen3.5-2B & \synd{99.9}{ForestGreen}{$+$49.4} & \synd{99.8}{ForestGreen}{$+$50.3} & \synd{99.9}{ForestGreen}{$+$44.4} & \synd{99.7}{ForestGreen}{$+$55.2} & \synd{57.7}{ForestGreen}{$+$3.5} & \synd{57.6}{ForestGreen}{$+$11.8} & \synd{31.2}{Gray}{$\pm$0.0} & \synd{49.5}{ForestGreen}{$+$1.6} & \synd{25.3}{ForestGreen}{$+$4.5} & \synd{69.8}{ForestGreen}{$+$20.1} & \synd{55.5}{ForestGreen}{$+$5.2} & \synd{2.5}{ForestGreen}{$+$2.4} & \synd{99.9}{Gray}{$\pm$0.0} & \synd{96.9}{ForestGreen}{$+$56.3} & \synd{99.3}{ForestGreen}{$+$39.9} \\
Qwen3.5-4B & \synd{100.0}{ForestGreen}{$+$43.1} & \synd{100.0}{ForestGreen}{$+$56.9} & \synd{100.0}{ForestGreen}{$+$32.9} & \synd{100.0}{ForestGreen}{$+$67.1} & \synd{57.7}{ForestGreen}{$+$3.8} & \synd{53.2}{ForestGreen}{$+$7.1} & \synd{33.3}{ForestGreen}{$+$0.1} & \synd{48.0}{ForestGreen}{$+$2.1} & \synd{26.3}{ForestGreen}{$+$5.4} & \synd{60.5}{ForestGreen}{$+$11.6} & \synd{63.9}{ForestGreen}{$+$12.8} & \synd{13.0}{ForestGreen}{$+$13.0} & \synd{100.0}{Gray}{$\pm$0.0} & \synd{99.8}{ForestGreen}{$+$30.2} & \synd{99.7}{ForestGreen}{$+$69.3} \\
\fi
Qwen3.5-9B & \synd{100.0}{ForestGreen}{$+$46.7} & \synd{100.0}{ForestGreen}{$+$53.3} & \synd{100.0}{ForestGreen}{$+$37.7} & \synd{100.0}{ForestGreen}{$+$62.3} & \synd{68.4}{ForestGreen}{$+$15.5} & \synd{63.2}{ForestGreen}{$+$16.1} & \synd{33.0}{ForestGreen}{$+$0.4} & \synd{48.8}{ForestGreen}{$+$1.7} & \synd{29.0}{ForestGreen}{$+$8.7} & \synd{63.1}{ForestGreen}{$+$7.3} & \synd{58.2}{ForestGreen}{$+$14.0} & \synd{0.2}{ForestGreen}{$+$0.1} & \synd{100.0}{ForestGreen}{$+$0.1} & \synd{96.5}{ForestGreen}{$+$8.5} & \synd{98.5}{ForestGreen}{$+$86.5} \\
\ifshowallmodels
Qwen3.5-27B & \synd{100.0}{ForestGreen}{$+$31.1} & \synd{100.0}{ForestGreen}{$+$68.9} & \synd{100.0}{ForestGreen}{$+$33.8} & \synd{100.0}{ForestGreen}{$+$66.2} & \synd{74.5}{ForestGreen}{$+$21.8} & \synd{72.0}{ForestGreen}{$+$24.7} & \synd{32.5}{ForestGreen}{$+$0.1} & \synd{48.0}{ForestGreen}{$+$0.3} & \synd{27.7}{ForestGreen}{$+$7.7} & \synd{63.3}{ForestGreen}{$+$12.0} & \synd{68.9}{ForestGreen}{$+$20.2} & \synd{0.3}{ForestGreen}{$+$0.3} & \synd{99.9}{Gray}{$\pm$0.0} & \synd{99.8}{ForestGreen}{$+$3.3} & \synd{99.8}{ForestGreen}{$+$96.2} \\
Qwen3.6-27B & \synd{100.0}{ForestGreen}{$+$30.1} & \synd{100.0}{ForestGreen}{$+$69.9} & \synd{100.0}{ForestGreen}{$+$28.2} & \synd{100.0}{ForestGreen}{$+$71.8} & \synd{72.2}{ForestGreen}{$+$19.5} & \synd{61.7}{ForestGreen}{$+$14.4} & \synd{32.3}{BrickRed}{$-$0.1} & \synd{49.4}{Gray}{$\pm$0.0} & \synd{25.6}{ForestGreen}{$+$7.3} & \synd{67.6}{ForestGreen}{$+$10.8} & \synd{58.1}{ForestGreen}{$+$14.9} & \synd{6.2}{ForestGreen}{$+$6.1} & \synd{99.9}{Gray}{$\pm$0.0} & \synd{99.6}{ForestGreen}{$+$0.6} & \synd{99.0}{ForestGreen}{$+$98.0} \\
Qwen3.8-27B & \synd{100.0}{ForestGreen}{$+$46.1} & \synd{100.0}{ForestGreen}{$+$53.9} & \synd{99.9}{ForestGreen}{$+$45.1} & \synd{99.9}{ForestGreen}{$+$54.7} & \synd{77.7}{ForestGreen}{$+$22.7} & \synd{69.1}{ForestGreen}{$+$24.1} & \synd{33.1}{ForestGreen}{$+$0.3} & \synd{46.4}{ForestGreen}{$+$0.1} & \synd{25.7}{ForestGreen}{$+$4.8} & \synd{58.1}{ForestGreen}{$+$4.1} & \synd{66.0}{ForestGreen}{$+$20.0} & \synd{7.9}{ForestGreen}{$+$7.6} & \synd{99.7}{Gray}{$\pm$0.0} & \synd{93.0}{ForestGreen}{$+$2.4} & \synd{95.7}{ForestGreen}{$+$86.3} \\
\fi

\midrule
\multicolumn{16}{c}{\textbf{In-context learning (5-shot)}}\\
\midrule

Gemma-4-E4B & \synd{56.1}{BrickRed}{$-$5.8} & \synd{46.6}{ForestGreen}{$+$8.5} & \synd{4.6}{BrickRed}{$-$44.3} & \synd{95.4}{ForestGreen}{$+$44.3} & \synd{51.8}{BrickRed}{$-$3.5} & \synd{48.9}{ForestGreen}{$+$4.2} & \synd{28.7}{BrickRed}{$-$0.4} & \synd{60.8}{BrickRed}{$-$0.1} & \synd{10.3}{ForestGreen}{$+$0.3} & \synd{41.5}{BrickRed}{$-$0.6} & \synd{58.1}{ForestGreen}{$+$0.2} & \synd{0.4}{BrickRed}{$-$0.2} & \synd{99.5}{ForestGreen}{$+$0.1} & \synd{94.4}{ForestGreen}{$+$5.2} & \synd{6.5}{BrickRed}{$-$4.3} \\
Gemma-4-12B & \synd{64.6}{BrickRed}{$-$7.6} & \synd{45.8}{ForestGreen}{$+$18.0} & \synd{8.4}{BrickRed}{$-$60.7} & \synd{92.5}{ForestGreen}{$+$61.6} & \synd{52.6}{ForestGreen}{$+$0.7} & \synd{47.3}{BrickRed}{$-$0.8} & \synd{28.7}{BrickRed}{$-$0.4} & \synd{57.5}{BrickRed}{$-$0.3} & \synd{14.0}{ForestGreen}{$+$0.9} & \synd{43.8}{BrickRed}{$-$0.3} & \synd{57.7}{ForestGreen}{$+$1.8} & \synd{0.7}{ForestGreen}{$+$0.2} & \synd{99.4}{BrickRed}{$-$0.1} & \synd{79.0}{BrickRed}{$-$14.4} & \synd{39.1}{ForestGreen}{$+$32.5} \\
InternVL3.5-8B & \synd{46.5}{BrickRed}{$-$0.7} & \synd{61.5}{ForestGreen}{$+$8.7} & \synd{40.7}{ForestGreen}{$+$5.9} & \synd{68.0}{ForestGreen}{$+$2.8} & \synd{53.1}{BrickRed}{$-$1.3} & \synd{47.6}{ForestGreen}{$+$2.0} & \synd{29.9}{BrickRed}{$-$0.1} & \synd{55.9}{ForestGreen}{$+$0.6} & \synd{14.6}{BrickRed}{$-$0.1} & \synd{59.4}{BrickRed}{$-$1.0} & \synd{40.8}{ForestGreen}{$+$1.2} & \synd{0.4}{ForestGreen}{$+$0.1} & \synd{99.6}{BrickRed}{$-$0.1} & \synd{83.3}{ForestGreen}{$+$26.6} & \synd{35.6}{BrickRed}{$-$7.7} \\
Qwen3.5-9B & \synd{70.8}{ForestGreen}{$+$17.5} & \synd{37.2}{BrickRed}{$-$9.5} & \synd{69.2}{ForestGreen}{$+$6.9} & \synd{43.3}{ForestGreen}{$+$5.6} & \synd{53.2}{ForestGreen}{$+$0.3} & \synd{46.7}{BrickRed}{$-$0.4} & \synd{32.5}{BrickRed}{$-$0.1} & \synd{46.8}{BrickRed}{$-$0.3} & \synd{21.8}{ForestGreen}{$+$1.5} & \synd{48.4}{BrickRed}{$-$7.4} & \synd{52.1}{ForestGreen}{$+$7.9} & \synd{0.1}{Gray}{$\pm$0.0} & \synd{99.9}{Gray}{$\pm$0.0} & \synd{99.5}{ForestGreen}{$+$11.5} & \synd{2.5}{BrickRed}{$-$9.5} \\

\midrule
\multicolumn{16}{c}{\textbf{Steering tokens}} \\
\midrule
InternVL3.5-8B & \synd{98.8}{ForestGreen}{$+$51.6} & \synd{98.9}{ForestGreen}{$+$46.1} & \synd{98.7}{ForestGreen}{$+$63.9} & \synd{98.9}{ForestGreen}{$+$33.7} & \synd{98.3}{ForestGreen}{$+$43.9} & \synd{96.9}{ForestGreen}{$+$51.3} & \synd{29.4}{BrickRed}{$-$0.6} & \synd{67.1}{ForestGreen}{$+$11.8} & \synd{17.0}{ForestGreen}{$+$2.3} & \synd{98.8}{ForestGreen}{$+$38.4} & \synd{98.7}{ForestGreen}{$+$59.1} & \synd{0.8}{ForestGreen}{$+$0.5} & \synd{99.7}{Gray}{$\pm$0.0} & \synd{99.6}{ForestGreen}{$+$42.9} & \synd{98.4}{ForestGreen}{$+$55.1} \\
Qwen3.5-9B & \synd{98.5}{ForestGreen}{$+$45.2} & \synd{100.0}{ForestGreen}{$+$53.3} & \synd{100.0}{ForestGreen}{$+$37.7} & \synd{100.0}{ForestGreen}{$+$62.3} & \synd{99.3}{ForestGreen}{$+$46.4} & \synd{97.3}{ForestGreen}{$+$50.2} & \synd{44.7}{ForestGreen}{$+$12.1} & \synd{86.2}{ForestGreen}{$+$39.1} & \synd{29.7}{ForestGreen}{$+$9.4} & \synd{99.3}{ForestGreen}{$+$43.5} & \synd{87.4}{ForestGreen}{$+$43.2} & \synd{0.1}{Gray}{$\pm$0.0} & \synd{99.9}{Gray}{$\pm$0.0} & \synd{99.6}{ForestGreen}{$+$11.6} & \synd{97.7}{ForestGreen}{$+$85.7} \\

\bottomrule
\end{tabular}
}
\end{table*}

\section{Perception biases in \vlms}

In this section, we aim to understand vision biases in \vlms. First, we assess biases that a ``neutral'' prompt causes on a global (cue biases) and local scale (cue fairness) (\cref{sec:experiments:default} and \cref{fig:teaser}, \textbf{(b)}). Then, we test if and how effective steering through prompt interventions such as optimized language and multi-modal in-context learning is (\cref{sec:experiments:language} and \cref{fig:teaser}, \textbf{(c)}). Finally, we investigate bottlenecks for prompt-based approaches in \cref{sec:experiments:bottlenecks}.

\subsection{Biases in neutral prompting}\label{sec:experiments:default}
We test the default perception behavior of multiple open-weight \vlms, such as the \modelemph{Qwen} family (\modelemph{Qwen3.5-\{0.8,2,4,9,27\}B} \cite{qwen2026qwen35}, \modelemph{Qwen3.6-27B} \cite{qwen2026qwen36_27b}, \modelemph{Qwen3.8-27B} \cite{qwen2026qwen38_27b}), \modelemph{InternVL3.5-\{1,2,4,8,14,38\}B} \cite{wang2025internvl35}, and \modelemph{Gemma-4-E4B} and \modelemph{Gemma-4-12B} \cite{gemma2026gemma4} under a \textit{neutral} VQA prompt ``\texttt{Which option best describes the image?}'' without any explicit linguistic biases to any of the tested cues. It is important to note that we are only comparing the \textit{relative} behavior of models under the same prompt and are, thus, not concerned with prompt tuning in this section. Different neutral formulations can add some noise, and we ablate them in \cref{sec:appendix:sens_prompt}. In addition to the model results, we obtain 10 human annotations per sample (54,122 unique annotators from 140 reported countries), allowing us to measure the human bias (see \cref{sec:appendix:humans} for details). The results are summarized for four models in \cref{tab:vlm_bias} (``Neutral prompting''); the full table can be found in \cref{tab:vlm_bias_full}.

\fakeparagraph{Human study.} The participants in our study behave fairly consistently with previously observed biases in other scenarios: they show a minor preference for the left side of an image \cite{nuthmann2014timecourse,nuthmann2023pseudoneglect}, are fairly unbiased to the vertical, prefer the foreground \cite{Fernandes2021foregroundbias}, have a chromatic bias \cite{tchernikov2010color} ranking consistent with the human eye's luminous sensitivity (green${>}$red${>}$blue) \cite{gibson1923visibility,wyszecki1982color}, target lower frequencies, typically indicating a large viewing distance or small stimulus \cite{schyns1994blobs}, see predominantly only the spectral phase \cite{oppenheim1981importance,piotrowski1982demonstration,guyader2004image}, and are distracted by the text, but still mostly pick the object \cite{smith1980tracing}.

\begin{figure}
    \centering
    \includegraphics[width=\linewidth]{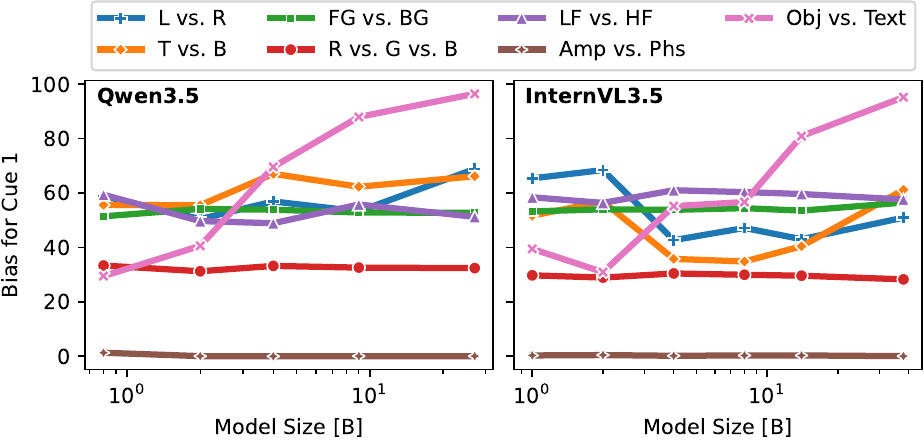}
    \caption{\textbf{Cue bias \vs.~model scale} for \modelemph{Qwen3.5} (left) and \modelemph{InternVL3.5} (right). Lines show the bias toward the first cue in each conflict (\eg `L' in the `L \vs R' cue).}
    \label{fig:vqa_cue_bias_scale}
    \vspace{-1em}
\end{figure}

\fakeparagraph{Model study.}
First, we find that despite the stimuli likely being out of distribution, all models recognize at least one cue, \ie the accuracy is typically well above 90\% (the smallest model, \modelemph{Qwen3.5-0.8B}, falls down to 86--88\% on just three cues). Regarding biases, we find that models do not share a common profile. For most conflicts, both the direction and the size of the bias change from model to model, and even between sizes of the same family (\cref{fig:vqa_cue_bias_scale}). The spread is largest for the semantic conflict. Object preference ranges from 56.7\% (\modelemph{InternVL3.5-8B}) to 93.4\% (\modelemph{Gemma-4-12B}), and from 29.4\% to 96.5\% within the \modelemph{Qwen3.5} family. The spatial conflicts come next. \modelemph{Gemma-4-12B} chooses the left (72.2\%) and top (69.1\%) cue much more often than humans do (51.2\% and 50.5\%), and \modelemph{InternVL3.5-8B} prefers the bottom (65.2\%). On the low- \vs~high-frequency conflict, the biases are moderate, but they point in opposite directions by family: \modelemph{Gemma-4} favors high frequencies, while \modelemph{InternVL3.5} and most \modelemph{Qwen3.5} models favor low frequencies.

Only three biases point the same way in every model: phase over amplitude, foreground over background, and the chromatic ranking (green${>}$red${>}$blue). These are also the only biases that match humans in direction, though not in strength. Models rely on phase almost entirely ($\geq$98.7\% \vs~97.4\%), but their preference for the foreground (51.5--56.5\% \vs~59.5\%) and for the green channel (45.9--60.9\% \vs~71.2\%) is weaker than in humans. Unlike the texture-shape bias \cite{geirhos2019imagenet,gavrikov2024can}, there is no conflict on which all models disagree with humans in direction. Still, every model differs substantially from humans on at least one conflict. The clearest case is frequency: no model comes close to the strong human preference for low frequencies (79.5\% \vs~at most 61.0\%).

In \cref{sec:appendix:sens_taskformat}, we additionally study whether the four selected models can at least name all cues through captioning. We find that even then, the recall is significantly reduced for chromatic and both spectral conflicts, while precision stays near perfect.

\fakeparagraph{Scaling laws.} Additionally, we check for potential ``scaling laws'', \ie biases that change with parameter count, in the \modelemph{Qwen3.5} and \modelemph{InternVL3.5} models (\cref{fig:vqa_cue_bias_scale}). The only consistent trend is in the semantic conflict: small models are strongly biased toward the text but shift toward the object as they grow. Object preference rises from 29.4\% to 96.5\% for \modelemph{Qwen3.5} (0.8B \vs~27B) and from 39.5\% to 95.2\% for \modelemph{InternVL3.5} (1B \vs~38B). Most other biases stay fairly stable across parameter counts. The vertical and horizontal spatial biases are the exceptions in both families, with stronger, non-monotonic trends.

\subsubsection{Cue fairness}\label{sec:experiments:default:probs}

In the previous section, we analyzed the prediction bias based on the top-1 answer (\cref{eq:bias}); now we turn to individual predictions. For instance, consider a model with a strong prediction bias for cue $c_1$ in a two-way conflict. This model might still represent the cues almost fairly, by predicting the probabilities for the cues one and two, as $q_1{=}0.5+\epsilon$ and $q_2{=}0.5-\epsilon$, respectively. Following \cref{eq:cue_fairness}, this would result in a fairness score of $\mathrm{I}_{\mathrm{TV}}((0.5{+}\epsilon, 0.5{-}\epsilon)){=}0.98$ with $\epsilon{=}10^{-2}$, while still being $100\%$ biased toward cue one.

\begin{figure}
    \centering
    \includegraphics[width=\linewidth]{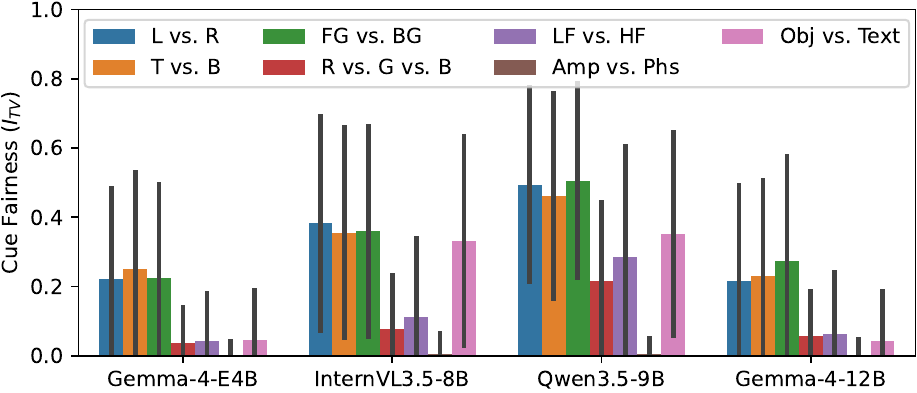}
    \caption{\textbf{Cue fairness.} Bars show mean$\pm$std. Higher is closer to uniform (fair).}
    \label{fig:vqa_cue_fairness}
\end{figure}

In \cref{fig:vqa_cue_fairness}, we show the distributions of our cue fairness metric for all cue conflicts and our four representative models. We find that all models show low cue fairness, hinting at unimodal cue probabilities. We also find large variance on all measurements, showing that fairness strongly depends on the sample. Overall, spatial biases have the highest fairness across all models, yet still far from uniform. We also find big differences between models; for instance, \modelemph{InternVL3.5-8B} and \modelemph{Gemma-4-12B} show a mean of $0.33$ and $0.04$, respectively, for the object \vs.~text conflict. 

Regarding scaling (see \cref{fig:vqa_cue_fairness_scale}), we find that for \modelemph{Qwen3.5}, fairness improves with scale up to 9B parameters, after which it stagnates. In contrast, \modelemph{InternVL3.5} models do not show a clear scaling trend. Across both families and all scales, cue confidences remain far from uniform, suggesting that scaling alone does not resolve the unimodal cue preference.

\subsection{Steering through prompting}\label{sec:experiments:language}

In this section, we attempt to steer behavior through prompting in two ways: in the first part, we explicitly optimize the language in the prompt to target the respective cue (\eg, \textit{``Which option best describes the left half of the image?''} for the left cue), and in the second part, we explore the effectiveness of multi-modal in-context learning (ICL). %

\subsubsection{Explicit cue targeting via language}\label{sec:experiments:default:explicit}

We optimize prompts using a GEPA-inspired reflective search \citep{agrawal2026gepa} implemented with \texttt{DSPy} \citep{khattab2024dspy}, using \modelemph{GPT-6 Luna (medium)} \cite{gpt6luna} to optimize the prompt prefix to target each cue in each conflict, on our four models (and a scaling study on \modelemph{Qwen3.5}). We seed the search with the neutral prompt and a manually designed targeted prompt. To measure generalization, we use disjoint class sets for prompt search and the final evaluation (defined in \cref{tab:class_splits}). Performance is evaluated on a different random sample of stimuli than the neutral prompt, so small negative changes reflect sampling noise. For more details, we refer the reader to \cref{sec:appendix:prompts:optimization}; \cref{tab:prompt_steering_prompts} lists all discovered prompts.

\begin{figure}
    \centering
    \includegraphics[width=0.8\linewidth]{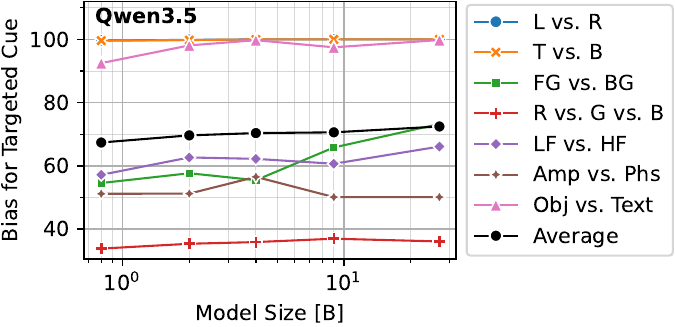}
    \caption{\textbf{Cue steering \vs.~model scale} for \modelemph{Qwen3.5}. Lines show the bias toward the targeted cue, averaged over each conflict's cues and over all cues (black).}
    \label{fig:vqa_steering_scale}
    \vspace{-1.5em}
\end{figure}

The results in \cref{tab:vlm_bias} (``Explicit cue targeting'') show that natural language prompting can be highly effective across all models in targeting vertical and horizontal spatial cues, as well as objects and text. Prompting is less effective at targeting specific frequencies or foreground/background objects, yet still shows significant improvements. Prompting seems largely ineffective at steering specific color channels, with the largest improvements on steering toward the blue channel. On the amplitude/phase bias, prompts barely move the bias: averaged over all models, amplitude answers rise only from $0.3\%$ to $2.2\%$ of all answers, while answers naming neither cue's class rise from $4.1\%$ to $28.6\%$. Across all biases, we see a positive correlation between the model scale and the steerability through prompts averaged by cues, while the strength seems to be related to the cue (\cref{fig:vqa_steering_scale}). 

Taken together, the results suggest that targeted prompting can be highly effective for some cues, and larger models are more steerable, but prompting alone can be insufficient or not effective at all to steer some biases.

\subsubsection{Multi-modal in-context learning (ICL)}\label{sec:experiments:default:icl}

Given the insufficiency of language shown in \cref{sec:experiments:default:explicit}, we explore if in-context learning (ICL) \cite{brown2020language} can be an alternative, as it provides additional multi-modal context. We test ICL on four models following the style in \cite{qin2024what}, where examples precede the actual prompt. We randomly pick five examples (5-shot) per prompt, filtering examples where the set of classes intersects with the classes in the test stimuli, to avoid shortcuts. %

While ICL has been shown to be effective for general vision tasks \cite{qin2024what}, we find ICL to be ineffective at bias steering (\cref{tab:vlm_bias}, ``In-context learning (5-shot)''). On the contrary, for all cues, we either observe that ICL has a negligible effect or even induces a strong bias, irrespective of the targeted cue. For instance, ICL targeting the bottom cue increases the bottom bias of \modelemph{Gemma-4-E4B} by 44 pp over the neutral prompt; however, targeting the top cue reduces the top bias by almost the same amount. In most cases, the biases obtained when targeting the individual cues sum to approximately $100\%$, showing a complete failure of targeting. Only a few cases show marginal steering improvements.

To find the source of this failure, we ask each model to explain the pattern in the in-context examples (100 contexts per model, conflict and cue) and use \modelemph{Qwen3.8-27B} as an LLM judge. For the foreground/background, chromatic and spectral cues, models practically never name the pattern ($\leq$5\%) and fall back to rules such as ``the most prominent object''. In the amplitude/phase case, they do not perceive the amplitude object at all. For the spatial and text cues, models often name the pattern but do not act on it. \Eg, the text rule is identified in 74--100\% of explanations, yet the text cue is selected in only 3--35\% of cases. The ineffectiveness of ICL thus stems from failing to induce the cue and, where it is induced, failing to apply it (see \cref{sec:appendix:icl_explain} for a breakdown).

This generalizes the finding of \cite{hemmat2024hidden}, that ICL cannot steer shape bias to a much broader set of biases.

\subsection{Bottlenecks for prompt-based steering}\label{sec:experiments:bottlenecks}

Any prompt-based steering intervention can only be successful if the representation generated by the vision encoder is flexible enough to (linearly) represent the cues in question. Note that this notion is independent of a representation bias: the representation may as well be biased to either cue, \eg, by assigning more magnitude, as long as all cues are represented. To better understand the feature cue representations generated by vision encoders, we apply linear probing as an example of a simplistic decoding model. To this end, we train separate linear layers on every layer of all \vlms{} to predict all $N\times C$ labels for each of our cue-conflict datasets. Details are shown in \cref{sec:appendix:probe_shapes}.

\begin{figure}[t]
    \centering
    \includegraphics[width=\linewidth]{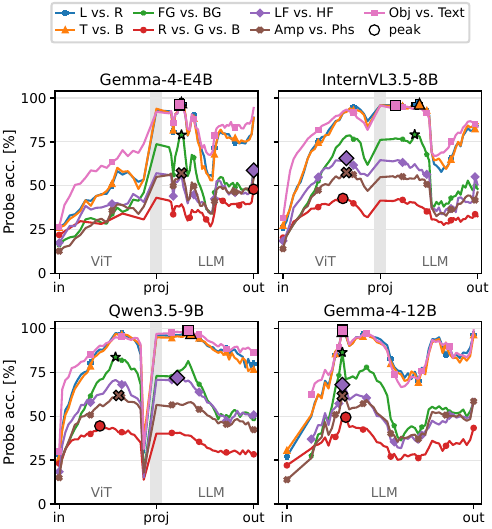}
    \vspace{-1.5em}
    \caption{\textbf{Cue decodability across depth.} Linear probe accuracy (mean over cues) at every layer, from the patch embedding (\emph{in}) through the vision encoder, the projector (\emph{proj}, shaded), and the language model (\emph{out}); large outlined markers mark each conflict's peak.}
    \label{fig:probe_depth}
\vspace{-1.5em}
    
\end{figure}

{\fakeparagraph{Decodability across depth.} \Cref{fig:probe_depth} shows the probe accuracy at every layer. Directly after patch embedding (before the first encoder block), all cues are only weakly linearly decodable ($13$--$32\%$); they become linearly decodable inside the network and peak in the late encoder or in early language-model layers; the projector is never the best place to read a cue and lies up to $11$ points below the peak (FG \vs BG in \modelemph{Qwen3.5-9B}). \modelemph{Gemma-4-12B} has no separate vision encoder: at its language-model input, the patch embedding, cues are only weakly linearly decodable ($14$--$30\%$) and peak around layer $9$ of the language model. In all models, spatial and object cues peak at $96$--$99\%$, whereas the remaining cues stay well below at every depth: FG \vs BG reaches at most $86\%$, LF \vs HF $72\%$, Amp \vs Phs $62\%$, and R \vs G \vs B $49\%$.}

Taken together, vertical/horizontal spatial and object-text cues are almost perfectly decodable from the visual tokens of every model, while color, amplitude--phase, and foreground--background cues remain only weakly decodable at every depth. For the latter, the information is not linearly accessible to the language model in the first place.
\section{Steering beyond prompts}\label{sec:beyond}

In \cref{sec:experiments:bottlenecks}, we have seen that steering through prompting can only be successful if the desired feature cue is represented in the vision tokens fed into the LLM. Yet, for some of our cues, we have observed that they are either non-linearly or not at all represented, resulting in a steering bottleneck. In this section, we now aim to explicitly loosen this bottleneck in \modelemph{Qwen3.5-9B} and \modelemph{InternVL3.5-8B} by learning optimized \ourtokens~pushing model predictions in the desired direction (\cref{fig:teaser}, \textbf{(c)}).

\fakeparagraph{Parameterization and injection.}
We learn $T{=}32$ \emph{\ourtokens} $\mathbf{R}\in\mathbb{R}^{T\times d_v}$, where $d_v$ is the hidden size of the vision encoder.
They are initialized from $\mathcal{N}(0, 0.02^2)$; the model stays frozen, and only $\mathbf{R}$ is trained.
The \ourtokens~are appended to the patch sequence of each image and take part in the encoder's bidirectional self-attention in every block.
They carry no positional information (no additive embedding and an identity rotary rotation, where applicable), while the image patches keep their 2D positions.
After the last encoder block, these tokens pass through the model's own pretrained vision-language projector like regular image tokens. 
\modelemph{Qwen3.5} and \modelemph{InternVL3.5} projectors merge groups of $2{\times}2$ tokens, therefore the $32$ \ourtokens~reduce to $8$ language-model tokens. We merge these tokens separately from the vision patches. 
To accommodate causal attention, the projected tokens are then inserted into the language-model input, half directly before and half directly after the vision tokens. They receive position ID $0$, \ie an identity rotary rotation, so they carry no positional information, and all other tokens keep their original positions.
The \ourtokens~therefore act at two levels: they rewrite the visual representation inside the encoder, and they act as soft tokens in the language model.

\fakeparagraph{Objective.}
Each example is a cue-conflict image with a multiple-choice question that lists all $C$ class names as options in random order.
Let $p=\operatorname{softmax}(\mathbf{z})$ be the option probabilities as defined in \cref{sec:method:measuring}, $\mathcal{S}$ the set of options whose classes appear as cues in the image ($|\mathcal{S}|{=}N$), and $t\in\mathcal{S}$ the option of the target cue.
We minimize
\begin{align}\label{eq:steering_loss}
\mathcal{L}(\mathbf{R}) &= -\log p_t - \lambda \log \sum_{c\in\mathcal{S}} p_c \\
&= -\log p(t \mid \mathcal{S}) - (1+\lambda)\log p(\mathcal{S}),
\end{align}
averaged over the batch, with $\lambda{=}5$ unless stated otherwise.
The first term increases the bias toward the target cue among the cues present; the second keeps probability mass on the cues and rules out trivially answering an absent class.

\fakeparagraph{Training.}
We train one set of tokens per target cue, sampling new cue-conflict examples from the training set at every step.
We use AdamW \cite{loshchilov2018decoupled} (weight decay $0$, gradient-norm clipping at $1$) with a learning rate of $10^{-3}$, $50$ linear warm-up steps followed by cosine decay, and batches of $32$ for at most $10$k updates (detailed in \cref{sec:appendix:steering_tokens}).
Every $1$k updates, we evaluate $16$ held-out classes disjoint from the training classes.
 Training stops early once the held-out target rate reaches $0.98$, or after two consecutive evaluations without an absolute gain of $0.01$; we report the tokens from the final update without checkpoint selection.
For \modelemph{InternVL3.5}, we raise the learning rate to $10^{-2}$ for the background cue, which was still improving when the schedule ended at $10^{-3}$.
On the red, green, blue, and amplitude cues, \modelemph{InternVL3.5} increasingly answers with classes that are not in the image when $\lambda{=}5$ ($12$--$64\%$ of the answers on the held-out classes). For these cues, we raise the on-cue weight to $\lambda{=}200$ (red, blue) and $\lambda{=}50$ (green, amplitude), which keeps this off-cue rate at $4.8$--$6.3\%$ at the cost of weaker steering.

\fakeparagraph{Results.}
\Cref{tab:vlm_bias} (``\Ourtokens'') reports the bias toward the target cue for \modelemph{InternVL3.5-8B} and \modelemph{Qwen3.5-9B}. On the left \vs right, top \vs bottom, and object \vs text conflicts, which prompts already steer to almost $100\%$, \ourtokens~reach  similar performance without any changes to the language.
On the conflicts where prompting stalls, \ourtokens~steer well beyond the limits of prompting (\cref{sec:experiments:language}). They raise both the foreground and the background bias to near-perfect rates, compared to $54.6$--$68.4\%$ with the optimized prompts, and the low-frequency bias to $98.8$--$99.3\%$ (prompts: $63.1$--$67.5\%$). The high-frequency bias reaches $98.7\%$ on \modelemph{InternVL3.5-8B} and $87.4\%$ on \modelemph{Qwen3.5-9B} (prompts: $47.9\%$ and $58.2\%$).
Color channels remain hard to steer. The exception is the green channel, which rises to $67.1\%$ on \modelemph{InternVL3.5-8B} and to $86.2\%$ on \modelemph{Qwen3.5-9B}, $10.6$ and $37.4$ pp above the optimized prompts. Red and blue stay close to their neutral bias, apart from red on \modelemph{Qwen3.5-9B} ($44.7\%$, $+11.7$ pp over the prompt); in all remaining cases, \ourtokens~still exceed the prompts, by $0.7$--$2.3$ pp.
The \modelemph{InternVL3.5-8B} color tokens trade steering strength for valid answers: with $\lambda{=}5$, they reach $93.0\%$ (red, blue; learning rate $10^{-2}$) and $90.3\%$ (green), but $12$--$28\%$ of their answers then name a class that is not in the image.
Finally, \ourtokens~are ineffective on the amplitude \vs phase conflict: phase is already at ceiling under the neutral prompt, and steering toward amplitude fails on both models.

In summary, \ourtokens~allow us to steer biases beyond the control that prompting provides. Inserted into the vision encoder, these tokens allow us to bypass the bottlenecks by changing the vision encoder-generated representation without breaking multi-modal alignment. However, these too have limits and cannot steer arbitrary feature cues. We hypothesize the following limit for color: the patch embedding linearly mixes the color channels with the spatial content of each patch, so color information is entangled with luminance and structure from the very first layer and may be hard to isolate downstream.

\section{Conclusion}\label{sec:conclusion}

Overall, we find that when confronted with forced choices, \vlms{} exhibit multiple vision biases, are misaligned with human vision, and do not represent ambiguous decisions fairly in logit space. Yet, \vlms{} can steer certain visual feature biases almost perfectly into any desired direction through natural language prompts, but not through in-context learning. Certain features, however, remain unaddressable by prompting due to a lack of linear separability after the vision encoder or misalignment of the vision-language modes. In these cases, \ourtokens{} can help push through the boundaries of prompt-based steering.

\clearpage

\iftoggle{cvprfinal}{%
\section*{Acknowledgements}
We thank Robert Geirhos and Abdelrahman Eldesokey for their valuable input. Additionally, we thank Rapidata for providing the human evaluation infrastructure used in this work and for supporting this research. Paul Gavrikov was supported by the KAUST Center of Excellence for Generative AI under award number 5940. For computing time, this research used Ibex managed by the Supercomputing Core Laboratory at King Abdullah University of Science \& Technology (KAUST) in Thuwal, Saudi Arabia. This work was further supported by the Tübingen AI Center. Arnas Uselis was supported by the International Max Planck Research School for Intelligent Systems (IMPRS-IS). Arnas Uselis also acknowledges support from the German Federal Ministry of Education and Research (BMBF): Tübingen AI Center, FKZ: 01IS18039A, and the International Max Planck Research School for Intelligent Systems (IMPRS-IS).
}{}

\section*{LLM Usage Statement}
As part of our method, LLMs optimize the steering prompts (\cref{sec:experiments:default:explicit}) and serve as a judge for the explanations of in-context patterns (\cref{sec:experiments:default:icl}). For our synthetic control dataset, all images were created with an image generation model (\cref{sec:appendix:dataset_synthetic}). Beyond that, we used generative AI tools to assist in writing parts of the codebase, to test design choices, to verify the correctness of results, to assist with the literature search, to create and polish figures, and to draft and polish text from outlines and drafts written by the authors. The authors reviewed all AI-assisted work: code was generated with a human in the loop and tested for correctness by the authors, all cited works were checked by the authors, and all AI-edited sections were proofread by all authors. The authors take full responsibility for the content of this work, including any text, claims, or artifacts produced with the help of generative AI.
    
{
    \small
    \bibliographystyle{ieeenat_fullname_unsrt}
    \bibliography{main}
}

\clearpage
\setcounter{page}{1}
\setcounter{section}{0}
\renewcommand{\thesection}{\Alph{section}}
\renewcommand{\thefigure}{\thesection.\arabic{figure}}
\renewcommand{\thetable}{\thesection.\arabic{table}}

\maketitlesupplementary

\begingroup
\let\clearpage\relax
\renewcommand*\numberline[1]{#1\hspace{1em}} %
\startcontents[appendix]
\printcontents[appendix]{l}{1}{\section*{Overview}\setcounter{tocdepth}{2}}
\endgroup
\FloatBarrier

\section{Reproducibility}\label{sec:appendix:reprodcuability}

\paragraph{Models.} All \vlms{} are loaded from the Hugging Face Hub at the pinned revisions listed in \cref{tab:models}, in \texttt{bfloat16} precision. We use the instruction-tuned checkpoints and disable thinking for all \modelemph{Qwen} models. Unless noted otherwise, all experiments use the random seed $42$, which also determines the shuffling of the answer options.

\paragraph{Software.} We use \texttt{CPython==3.11.14}, \texttt{torch==2.10.0}, \texttt{torchvision==0.25.0}, \texttt{accelerate==1.14.0}, and \texttt{dspy==3.3.1}. The main results are computed with \texttt{transformers==5.15.1}, which is required to run \modelemph{Gemma-4-12B}; the initial ablations in this appendix were computed with \texttt{transformers==5.6.1}.

\paragraph{Hardware.} All model evaluations are run on NVIDIA A100-SXM4-80GB GPUs. LLM judging, the training of \ourtokens{} (\cref{sec:beyond}), and the diagnosis of in-context learning (\cref{sec:experiments:default:icl}) are run on NVIDIA RTX PRO 6000 (96GB) GPUs.

\paragraph{Proprietary models.} Two parts of our pipeline rely on closed models: the prompt optimization uses \modelemph{GPT-6 Luna (medium)} \cite{gpt6luna} through the \texttt{OpenAI API} (\cref{sec:appendix:prompts:optimization}), and the synthetic control dataset was generated with \modelemph{GPT-5.6 Sol (medium)} and \modelemph{gpt-image-2} (\cref{sec:appendix:dataset_synthetic}). We will release the resulting prompts and images, so that none of our results requires access to these models.

\begin{table*}[t]
    \centering
    \caption{\textbf{Evaluated models.} Display name, Hugging Face Hub identifier, pinned revision, and license of every VLM in this study.}
    \label{tab:models}
    \resizebox{\textwidth}{!}{
        \begin{tabular}{@{}llll@{}}
             \toprule
             \multicolumn{1}{@{}l}{\textbf{Model Name}} & \multicolumn{1}{l}{\textbf{Slug}} & \multicolumn{1}{l}{\textbf{Revision}} & \multicolumn{1}{l@{}}{\textbf{License}} \\
             \midrule
             Gemma-4-12B & \texttt{google/gemma-4-12B-it} & \texttt{707f0a3b8a3c7ad586ed01e27eafbad8a27dd0f7} & Apache-2.0 \\
             \midrule
             Gemma-4-E4B & \texttt{google/gemma-4-E4B-it} & \texttt{ee0ef6023621cff504d758262d4e04895a5af4a2} & Apache-2.0 \\
             \midrule
             InternVL3.5-1B & \texttt{OpenGVLab/InternVL3\_5-1B-HF} & \texttt{9191dbccf312b537016f041b25d61c72e7c5c9f3} & Apache-2.0 \\
             InternVL3.5-2B & \texttt{OpenGVLab/InternVL3\_5-2B-HF} & \texttt{3f301ffcf3dcbb47893afae6650ea3e78d96fb6d} & Apache-2.0 \\
             InternVL3.5-4B & \texttt{OpenGVLab/InternVL3\_5-4B-HF} & \texttt{6bd4487402110ef9889ba50eb7aefeb302526fed} & Apache-2.0 \\
             InternVL3.5-8B & \texttt{OpenGVLab/InternVL3\_5-8B-HF} & \texttt{741a7d03020411e666c6109218ab71e08151ef86} & Apache-2.0 \\
             InternVL3.5-14B & \texttt{OpenGVLab/InternVL3\_5-14B-HF} & \texttt{226b96d5912e69159abc0384cefcbd51487fdce0} & Apache-2.0 \\
             InternVL3.5-38B & \texttt{OpenGVLab/InternVL3\_5-38B-HF} & \texttt{7c830fc25e874f9a861087e3c70587f2b9b555f3} & Apache-2.0 \\
             \midrule
             Qwen3.5-0.8B & \texttt{Qwen/Qwen3.5-0.8B} & \texttt{2fc06364715b967f1860aea9cf38778875588b17} & Apache-2.0 \\
             Qwen3.5-2B & \texttt{Qwen/Qwen3.5-2B} & \texttt{15852e8c16360a2fea060d615a32b45270f8a8fc} & Apache-2.0 \\
             Qwen3.5-4B & \texttt{Qwen/Qwen3.5-4B} & \texttt{851bf6e806efd8d0a36b00ddf55e13ccb7b8cd0a} & Apache-2.0 \\
             Qwen3.5-9B & \texttt{Qwen/Qwen3.5-9B} & \texttt{c202236235762e1c871ad0ccb60c8ee5ba337b9a} & Apache-2.0 \\
             Qwen3.5-27B & \texttt{Qwen/Qwen3.5-27B} & \texttt{fc05daec18b0a78c049392ed2e771dde82bdf654} & Apache-2.0 \\
             Qwen3.6-27B & \texttt{Qwen/Qwen3.6-27B} & \texttt{6a9e13bd6fc8f0983b9b99948120bc37f49c13e9} & Apache-2.0 \\
             Qwen3.8-27B & \texttt{Qwen/Qwen3.8-27B} & \texttt{1d4bf0f2ff6012fd82039f2fa52739d0dd7c60c0} & Apache-2.0 \\
             \bottomrule
        \end{tabular}
    }
\end{table*}

\section{Dataset}\label{sec:appendix:dataset}

\begin{table}[t]
    \caption{\textbf{Class splits of \textit{MiniObjectNet}.} The 32 classes are divided into two disjoint sets of 16 classes.%
    }
    \label{tab:class_splits}
    \centering
    \begin{tabular}{@{}p{0.15\linewidth} p{0.79\linewidth}@{}}
        \toprule
         \multirow{3}{*}{\textbf{train}} & chair, cheeseburger, cow, elephant, hammer, horse, lemon, mushroom, orange, pizza, shark, sheep, tank, teapot, truck, zebra\\
         \midrule
         \multirow{3}{*}{\textbf{test}} & airplane, apple, bear, bicycle, bird, boat, broccoli, car, cat, cauliflower, clock, dog, hourglass, stop sign, strawberry, teddy bear \\ 
         \bottomrule
    \end{tabular}
\end{table}

\subsection{Natural data}

\textit{MiniObjectNet} dataset consists of 32 classes, which we evenly divide into training and testing splits, each containing 16 distinct classes (shown in \cref{tab:class_splits}). Each class consists of approximately 20 images, with a minimum of 12 images and a minimum resolution of $558\times 558$ px. The images in our dataset were cropped from source images to show the class object clearly centered with minimal padding. We manually filtered samples to show no other objects in the background. The selected classes are all part of the ImageNet taxonomy \cite{deng2009imagenet}.

\paragraph{License.} Images were collected from two stock photo platforms: $630$ images from Pexels (\url{https://www.pexels.com}) and $5$ images from Unsplash (\url{https://unsplash.com}). They are used under the Pexels License (\url{https://www.pexels.com/license/}) and the Unsplash License (\url{https://unsplash.com/license}). We thank all photographers for making their work available. All natural images and stimuli shown in this paper are derived from these images.

\subsection{Synthetic variant}\label{sec:appendix:dataset_synthetic}

\begin{figure*}[t]
    \centering
    \includegraphics[width=\textwidth]{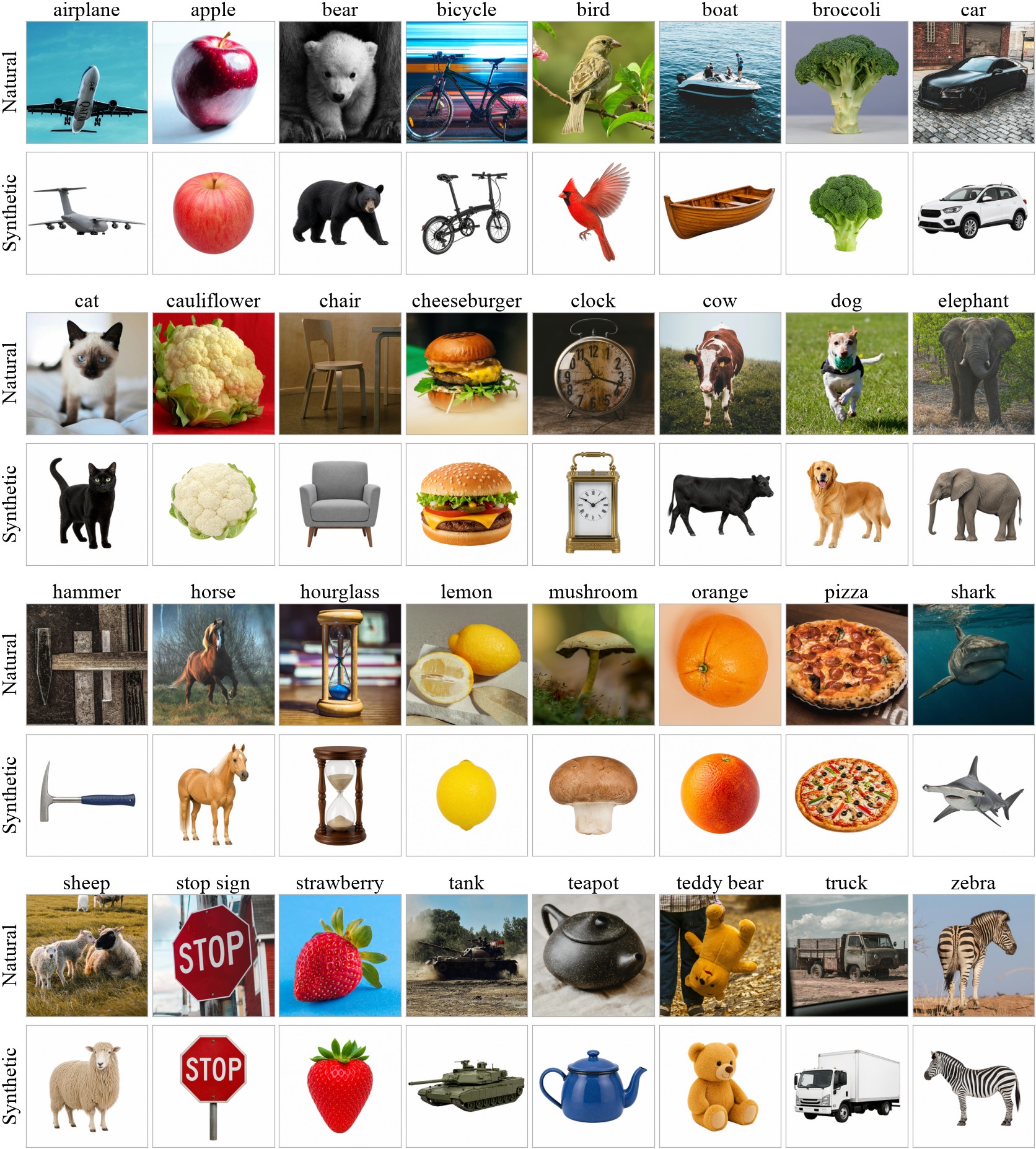}
    \caption{\textbf{Samples of \textit{MiniObjectNet}.} One image per class from the natural dataset (top rows) and its synthetic variant (bottom rows). For both, we show the first image of each class in the fixed order from which the stimuli are drawn.}
    \label{fig:dataset_samples}
\end{figure*}

Despite our careful manual curation, the natural images in \textit{MiniObjectNet} may still contain dataset biases, \eg, in the background, position, or size of the objects, which could interact with our cue conflicts. To control for such biases, we additionally create a fully synthetic variant of \textit{MiniObjectNet} with 20 images for each of the 32 classes. The images were generated by \modelemph{GPT-5.6 Sol (medium)} with \modelemph{gpt-image-2} using the prompt below, which asks for a single centered object on a plain white background and for variance in pose and kind within each class. From these images, we generate the cue-conflict stimuli with the same pipeline as for the natural images (\cref{sec:method}). \cref{fig:dataset_samples} shows one sample per class from both variants. 

\begin{Verbatim}[breaklines,breaksymbolleft={},fontsize=\footnotesize]
create a synthetic dataset of objects on white background without shadows (so I can cut them out cleanly) or any other objects. Use your image generation capabilities. 
- objects: horse, apple, boat, cauliflower, cow, hammer, mushroom, sheep, tank, zebra, bear, broccoli, chair, dog, horse, orange, stopsign, teapot, bear, bicycle, car, cheeseburger, elephant, hourglass, pizza, strawberry, teddy bear, airplane, bird, cat, clock, lemon, shark, truck
- 20 images per class

Make sure that there is some variance in pose and kind per class, eg there should be red and green apples, different cat breeds etc. The objects should be centered
\end{Verbatim}

\section{Probed representations}\label{sec:appendix:probe_shapes}
\begin{figure*}[t]
    \centering
    \includegraphics[width=\linewidth]{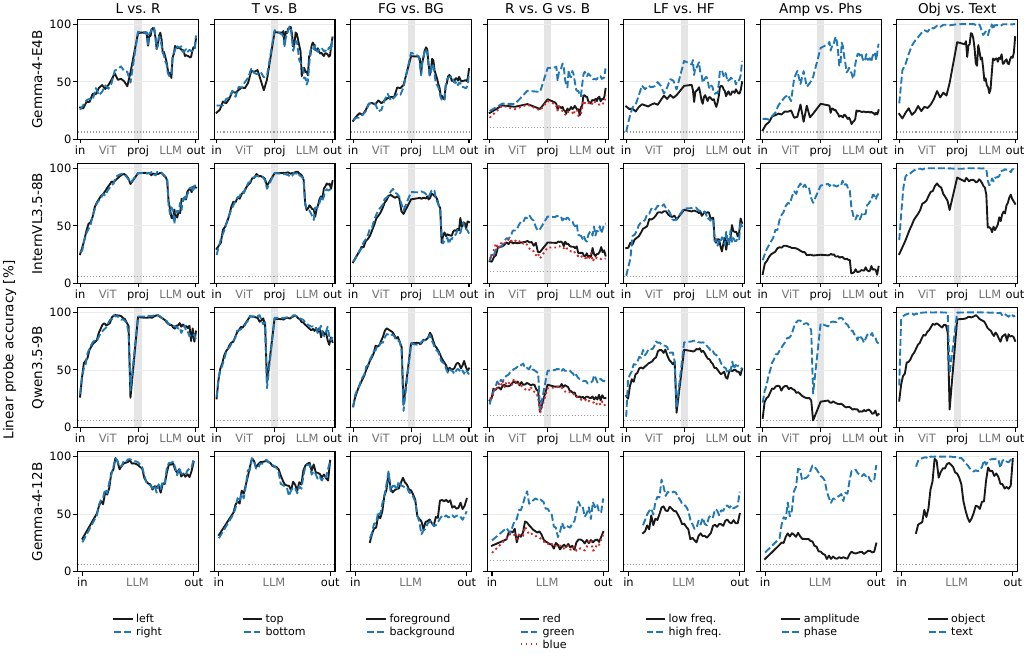}
    \caption{\textbf{Cue decodability across depth, per cue.} Linear probe accuracy for each cue of each conflict (columns) and model (rows), at every layer from the patch embedding (\emph{in}) through the vision encoder, the projector (shaded), and the language model (\emph{out}). Dotted: chance. Note that \modelemph{Gemma-4-12B} has no separate vision encoder.}
    \label{fig:probe_depth_percue}
\end{figure*}

{%
The detailed probing results are illustrated in \cref{fig:probe_depth_percue}. \cref{tab:probe_shapes} lists, for each model, the representations probed in \cref{sec:experiments:bottlenecks}.

\begin{table}
    \centering
    \caption{\textbf{Probed representations per model.} Number of layers and the shape of representation (image tokens $\times$ hidden size) in the vision encoder (ViT) and in the language model (LLM, image tokens only); we omit the projector since its output is of LLM shape. The last column gives the flattened probe input dimension used for probing.}
    \label{tab:probe_shapes}
    \footnotesize
    \setlength{\tabcolsep}{2.5pt}
    \begin{tabular*}{\linewidth}{@{\extracolsep{\fill}}lccc@{}}
    \toprule
    Model & ViT & LLM & Probe dim. \\
     & blocks, shape & layers, shape & ViT / LLM \\
    \midrule
    \modelemph{Qwen3.5-9B} & 27, $784{\times}1152$ & 32, $196{\times}4096$ & 0.90M / 0.80M \\
    \modelemph{InternVL3.5-8B} & 24, $1025{\times}1024$ & 36, $256{\times}4096$ & 1.05M / 1.05M \\
    \modelemph{Gemma-4-E4B} & 16, $10080{\times}768$ & 42, $1089{\times}2560$ & 7.74M / 2.79M \\
    \modelemph{Gemma-4-12B} & -- & 48, $1089{\times}3840$ & -- / 4.18M \\
    \bottomrule
    \end{tabular*}
\end{table}

Each model differs in how it processes visual information: \modelemph{Qwen3.5-9B} merges $2{\times}2$ neighboring patches ($28{\times}28 \to 14{\times}14$); \modelemph{InternVL3.5-8B} drops the class token and applies a $2{\times}2$ pixel shuffle ($32{\times}32 \to 16{\times}16$); \modelemph{Gemma-4-E4B} pools $3{\times}3$ patches ($99{\times}99 \to 33{\times}33$; its encoder pads every image to $10080$ tokens, of which $9801$ are patches). \modelemph{Gemma-4-12B} has no separate vision encoder: its patch embedding ($1089{\times}3840$) is the language-model input.}

\paragraph{Data.} Probes are trained on the cue-conflict stimuli built from \textit{MiniObjectNet} (seed $42$). Source photos of each class are split into disjoint train and test sets. Stimuli combine one photo per cue from distinct classes: $16$ classes for all conflicts except R \vs G \vs B, which uses $10$ classes ($3$ cues). Each ordered class combination is rendered $5$ times, giving $16{\cdot}15{\cdot}5 = 1{,}200$ train and $1{,}200$ test stimuli per two-cue conflict, and $10{\cdot}9{\cdot}8{\cdot}5 = 3{,}600$ each for R \vs G \vs B. All classes occur in both splits; the label of each cue is uniformly distributed over the classes. Each stimulus is passed through the model once, together with the multiple-choice prompt listing all classes, and the hidden states of the image tokens are recorded at every probed position (\cref{tab:probe_shapes}).

\paragraph{Probe.} The probe is a single linear layer on the flattened image tokens (tokens $\times$
hidden size) with one $C$-way output per cue, trained jointly with the mean cross-entropy over cues. We train
it with SGD and with Adam for each learning rate in \{$10^{-1}$, $5{\cdot}10^{-2}$, $10^{-2}$,
$5{\cdot}10^{-3}$, $10^{-3}$, $5{\cdot}10^{-4}$, $10^{-4}$, $5{\cdot}10^{-5}$, $10^{-5}$\}, \ie $18$ runs per
position %
We report the per-cue test accuracy of the best epoch of the best run.

\section{Prompts}\label{sec:appendix:prompts}

\cref{tab:prompts} shows the main prompts utilized in our study.

\begin{table}[t]
    \centering
    \caption{\textbf{Prompt templates.} \textbf{\texttt{<image>}} marks where the stimulus is inserted. All VQA variants list the class options in shuffled order (\cref{sec:appendix:sens_optionshuffle}) and are answered with a single option letter. In the ICL prompt, \textit{\{VQA prompt\}} is the complete 0-shot prompt including the options and the answer instruction; it is identical for all examples and the test stimulus. The examples \textbf{\texttt{<image$_k$>}} are cue-conflict stimuli of the same conflict that share no class with the test stimulus, and \textit{\{Letter$_k$\}} is the option letter of the class shown in the targeted cue of example $k$. The whole prompt is passed as a single user turn.}
    \label{tab:prompts}
    \small
    \begin{tabular}{@{}>{\raggedright\arraybackslash}p{0.2\linewidth} | >{\raggedright\arraybackslash}p{0.74\linewidth}@{}}
        \toprule
        \textbf{Task} & \textbf{Prompt} \\
         \midrule
         VQA (0-shot) (all main results) & User: \textbf{\texttt{<image>}} Which option best describes the image? Options:\linebreak
         A. \textit{\{Class 1\}}\linebreak
         B. \textit{\{Class 2\}}\linebreak
         ...\linebreak
         Answer only with the correct option letter. \\
         \midrule
         VQA (ICL, 5-shot) (\cref{sec:experiments:default:icl}) & User: \textbf{\texttt{<image$_1$>}} Question: \textit{\{VQA prompt\}} Answer: \textit{\{Letter$_1$\}}\linebreak
         ...\linebreak
         \textbf{\texttt{<image$_5$>}} Question: \textit{\{VQA prompt\}} Answer: \textit{\{Letter$_5$\}}\linebreak
         \textbf{\texttt{<image>}} Question: \textit{\{VQA prompt\}} Answer: \\
         \midrule
         Caption (\cref{sec:appendix:sens_taskformat}) & User: \textbf{\texttt{<image>}} Caption this image in a single sentence. \\
         \midrule
         Open-ended (\cref{sec:appendix:sens_taskformat}) & User: \textbf{\texttt{<image>}} Which object is shown in this image? Answer directly with the objects name. \\
         \midrule
         ICL explanation (\cref{sec:appendix:icl_explain}) & User: \textbf{\texttt{<image$_1$>}} Question: \textit{\{VQA prompt\}} Answer: \textit{\{Letter$_1$\}}\linebreak
         ...\linebreak
         \textbf{\texttt{<image$_5$>}} Question: \textit{\{VQA prompt\}} Answer: \textit{\{Letter$_5$\}}\linebreak
         Above are 5 solved examples. Each example shows visual input (one or more images), a multiple-choice question, and the correct answer letter. The correct answers follow one consistent rule that is the same for every example. Do not answer with a single letter now. Instead: 1. For each example (Example 1 to Example 5), briefly describe what you see in the visual input and explain why the given answer is the correct one. 2. Then state, in one sentence, the general rule that determines the correct answer for any new input. Begin that sentence with ``Rule:''. \\
         \bottomrule
    \end{tabular}
\end{table}

\subsection{Prompt optimization}\label{sec:appendix:prompts:optimization}
We optimized prompts using a GEPA-inspired reflective search \citep{agrawal2026gepa} implemented with \texttt{DSPy} \citep{khattab2024dspy}, optimizing a biased prompt prefix for all combinations of cues, conflicts, and models. Prompt revisions were proposed by \modelemph{GPT-6 Luna (medium)} \cite{gpt6luna} through the \texttt{OpenAI API} \citep{openai2026api}. We allowed the model to generate up to 16K tokens with default parameters. 

The search was initialized with the default zero-shot VQA prompt and the manually crafted prompt for the targeted cue (\cref{tab:prompts}). Candidate prefixes were limited to $1{,}200$ characters. The optimization then used a beam search over three rounds with four proposals per round, yielding up to 12 generated candidates. In each round, proposals were derived from the three best candidates so far, informed by their scores, misclassified examples, and the best previously tried prefixes. Each candidate was evaluated on a fixed set of $1{,}024$ and $480$ cue-conflict examples on the chromatic and the other conflicts, respectively. Candidates were ranked using a target-rate objective, defined as the prediction rate for the targeted cue $t$ over the $M_{\text{eval}}$ evaluation examples
\begin{equation}
    \text{TR}_{t} = \frac{1}{M_{\text{eval}}}\sum_{i=1}^{M_{\text{eval}}} \mathbbm{1}\left[\hat{y}^{(i)} = y_t^{(i)}\right],
\end{equation}
where $\hat{y}^{(i)}$ is the predicted option for example $i$ and $y_c^{(i)}$ is the label carried by cue $c$, so that $y_t^{(i)}$ is the label of the targeted cue. Ties were broken by the mean probability margin of the targeted cue and, then, by the prefix length. We applied early stopping once the best candidate reached a cue bias of $0.9$, \ie, $\text{TR}_{t} / \sum_c \text{TR}_c \geq 0.9$.

To measure generalization beyond the categories observed during optimization, we used disjoint class sets for prompt search (training) and final evaluation (testing) as defined in \cref{tab:class_splits}. The best-performing prompt on the search set, the manually crafted prompt, and the neutral baseline were evaluated on the full test conflict dataset.

\cref{tab:prompt_steering_prompts} at the end of the appendix lists the selected prompt for every conflict, targeted cue, and model. These are the prompts behind the ``Explicit cue targeting'' biases in \cref{tab:vlm_bias,tab:vlm_bias_full}.

\FloatBarrier

\section{Sensitivity analyses}\label{sec:appendix:sensitivity}

This section tests how sensitive our bias measurements are to the choices of our setup: the source images and the class composition (\cref{sec:appendix:data_source,sec:appendix:sens_classcompandnumsamples}), the order of the answer options and the task format (\cref{sec:appendix:sens_optionshuffle,sec:appendix:sens_taskformat}), the design of the VQA prompt (\cref{sec:appendix:sens_distractor,sec:appendix:sens_prompt}), the visual token granularity (\cref{sec:appendix:token_budget}), and the frequency cutoff of the low/high-frequency conflict (\cref{sec:appendix:lfhf_weep}).

\subsection{Data source}\label{sec:appendix:data_source}

Natural images can carry dataset biases, \eg, in the background, position, or size of the objects, which could interact with our cue conflicts and distort the measured biases. To rule this out, we repeat the neutral VQA evaluation with stimuli generated from the synthetic variant of \textit{MiniObjectNet} (\cref{sec:appendix:dataset}), which shows single centered objects on a plain white background. \cref{tab:vlm_bias_real_vs_syn} compares the biases on the synthetic and the natural stimuli on our four main models. The biases are highly consistent: across all models and cues, the mean absolute difference is $2.4$ pp, the largest difference is $7.9$ pp, and $52$ of $60$ differences are within $5$ pp. This suggests that either such biases are largely absent from our source images, or they have little influence on our cue conflicts. In either case, our bias measurements appear to be robust to the data source.

\begin{table*}[t]
\centering
\caption{\textbf{Comparison of prediction bias between natural and synthetic data.} Bias (\%, \cref{eq:bias}) toward each cue on cue-conflict stimuli generated from the synthetic source images under neutral prompting. Values in parentheses give the difference to the stimuli generated from the natural \textit{MiniObjectNet} images (synthetic $-$ natural).}
\label{tab:vlm_bias_real_vs_syn}

\small
\resizebox{\linewidth}{!}{
\begin{tabular}{@{}l|cc|cc|cc|ccc|cc|cc|cc@{}}
\toprule
\multicolumn{1}{c}{}                                   & \multicolumn{6}{c}{\textbf{Spatial}}                                                                        & \multicolumn{3}{c}{\textbf{Chromatic}}                                         & \multicolumn{4}{c}{\textbf{Spectral}}                                                                                      & \multicolumn{2}{c}{\textbf{Semantic}} \\
\cmidrule(l{5pt}r{5pt}){2-7}\cmidrule(l{5pt}r{5pt}){8-10}\cmidrule(l{5pt}r{5pt}){11-14}\cmidrule(l{5pt}r{5pt}){15-16}
\multicolumn{1}{@{}l}{\textbf{Model}} & \multicolumn{1}{c}{Left} & \multicolumn{1}{c}{Right} & \multicolumn{1}{c}{Top} & \multicolumn{1}{c}{Btm.} & \multicolumn{1}{c}{Fg.} & \multicolumn{1}{c}{Bg.} & \multicolumn{1}{c}{Red} & \multicolumn{1}{c}{Grn.} & \multicolumn{1}{c}{Blue} & \multicolumn{1}{c}{LF} & \multicolumn{1}{c}{HF} & \multicolumn{1}{c}{Amp.} & \multicolumn{1}{c}{Phs.} & \multicolumn{1}{c}{Obj.} & \multicolumn{1}{c}{Text} \\
\midrule
Gemma-4-12B & \synd{70.8}{BrickRed}{$-$1.4} & \synd{29.2}{ForestGreen}{$+$1.4} & \synd{67.9}{BrickRed}{$-$1.2} & \synd{32.1}{ForestGreen}{$+$1.2} & \synd{51.8}{BrickRed}{$-$0.1} & \synd{48.2}{ForestGreen}{$+$0.1} & \synd{26.6}{BrickRed}{$-$2.5} & \synd{59.4}{ForestGreen}{$+$1.6} & \synd{13.9}{ForestGreen}{$+$0.8} & \synd{40.8}{BrickRed}{$-$3.3} & \synd{59.2}{ForestGreen}{$+$3.3} & \synd{0.0}{BrickRed}{$-$0.5} & \synd{100.0}{ForestGreen}{$+$0.5} & \synd{97.5}{ForestGreen}{$+$4.1} & \synd{2.5}{BrickRed}{$-$4.1} \\\midrule
Gemma-4-E4B & \synd{67.7}{ForestGreen}{$+$5.8} & \synd{32.3}{BrickRed}{$-$5.8} & \synd{55.5}{ForestGreen}{$+$6.6} & \synd{44.5}{BrickRed}{$-$6.6} & \synd{52.7}{BrickRed}{$-$2.6} & \synd{47.3}{ForestGreen}{$+$2.6} & \synd{28.0}{BrickRed}{$-$1.1} & \synd{57.9}{BrickRed}{$-$3.0} & \synd{14.1}{ForestGreen}{$+$4.1} & \synd{41.2}{BrickRed}{$-$0.9} & \synd{58.8}{ForestGreen}{$+$0.9} & \synd{0.0}{BrickRed}{$-$0.6} & \synd{100.0}{ForestGreen}{$+$0.6} & \synd{92.6}{ForestGreen}{$+$3.4} & \synd{7.4}{BrickRed}{$-$3.4} \\\midrule
InternVL3.5-8B & \synd{55.1}{ForestGreen}{$+$7.9} & \synd{44.9}{BrickRed}{$-$7.9} & \synd{35.5}{ForestGreen}{$+$0.7} & \synd{64.5}{BrickRed}{$-$0.7} & \synd{55.2}{ForestGreen}{$+$0.8} & \synd{44.8}{BrickRed}{$-$0.8} & \synd{29.2}{BrickRed}{$-$0.8} & \synd{49.2}{BrickRed}{$-$6.1} & \synd{21.6}{ForestGreen}{$+$6.9} & \synd{57.6}{BrickRed}{$-$2.8} & \synd{42.4}{ForestGreen}{$+$2.8} & \synd{0.2}{BrickRed}{$-$0.1} & \synd{99.8}{ForestGreen}{$+$0.1} & \synd{55.0}{BrickRed}{$-$1.7} & \synd{45.0}{ForestGreen}{$+$1.7} \\\midrule
Qwen3.5-9B & \synd{57.1}{ForestGreen}{$+$3.8} & \synd{42.9}{BrickRed}{$-$3.8} & \synd{61.0}{BrickRed}{$-$1.3} & \synd{39.0}{ForestGreen}{$+$1.3} & \synd{51.4}{BrickRed}{$-$1.5} & \synd{48.6}{ForestGreen}{$+$1.5} & \synd{29.8}{BrickRed}{$-$2.8} & \synd{46.7}{BrickRed}{$-$0.4} & \synd{23.5}{ForestGreen}{$+$3.2} & \synd{51.7}{BrickRed}{$-$4.1} & \synd{48.3}{ForestGreen}{$+$4.1} & \synd{0.0}{BrickRed}{$-$0.1} & \synd{100.0}{ForestGreen}{$+$0.1} & \synd{88.7}{ForestGreen}{$+$0.7} & \synd{11.3}{BrickRed}{$-$0.7} \\

\bottomrule
\end{tabular}
}
\end{table*}

\subsection{Class composition and sample quantity}\label{sec:appendix:sens_classcompandnumsamples}

Our measurements rest on a fixed set of 16 test classes and a limited number of images per class. Since the classes in a stimulus determine which objects compete, a different class set or a different draw of images could change the measured biases. We test \modelemph{Qwen3.5-9B} with VQA on our train and test split, as well as 5 folds of 16 randomly sampled classes to test how sensitive our bias measurements are to the composition of classes (classes shown in \cref{tab:crossval_fold_classes}). While the results in \cref{tab:sensitivity_folds} show some fluctuations between different class compositions, the variance is fairly small across folds. The biggest outlier is the object/text conflict (specifically in the F1 fold), where certain combinations of class pairs induce an irregularly high object bias. Overall, there are only minor differences between our train and test folds, and we consider this split suitable for our analysis.

\begin{figure}[t]
    \centering
    \includegraphics[width=\linewidth]{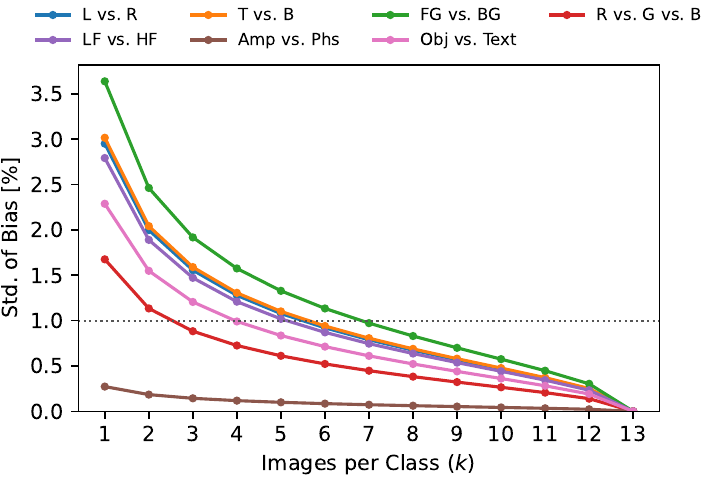}
    \caption{\textbf{Sensitivity of the bias to the number of images per class.} \modelemph{Qwen3.5-9B} on the 16 test classes with 13 images per class. For each $k$ (x-axis), we compute the bias on every subset of $k$ images per class and report the standard deviation of the bias across all subsets, averaged over the cues of each conflict. The dotted line marks 1 pp; we use $k{=}7$ ($k{=}3$ for the chromatic conflict).}
    \label{fig:crossval_test_global_error}
\end{figure}

Furthermore, we combine the class composition analysis with a sensitivity analysis of samples/class and use 12--13 images/class (corresponding to the smallest quantity of samples across all classes, depending on the class configuration). The results on the test class fold in \cref{fig:crossval_test_global_error} show that for all conflicts, the average error across subsets drops below 1 pp at 7 images per class, which we select as the hyperparameter for our analysis. Because our classes contain only 12--13 images, these subsets necessarily overlap, so the curve measures the stability of the bias within our image pool; for the same reason, the error vanishes at 13 images per class, where only one subset exists. The error under a fresh draw of images is larger by approximately the inverse finite-population correction $\sqrt{12/(13-k)}$ for $k$ images per class, \ie, by a factor of ${\approx}1.4$ at $k{=}7$. With this correction, we estimate the error under a fresh draw of 7 images per class at about $1.4$ pp for the least stable conflict (foreground/background) and below that for all others, which is small compared to the differences between models and steering methods that we discuss. This results in a total of $1{,}680$ samples per conflict. For the chromatic conflict, the only 3-way one, we select 3 samples per class, as this is sufficient to achieve the same error rate, while significantly reducing the number of samples due to the additional combinations, resulting in $10{,}080$ samples per conflict.

\begin{table*}[t]
    \caption{\textbf{Class folds.} Classes of the train and test splits and of the five random folds.}
    \label{tab:crossval_fold_classes}
    \begin{tabular}{@{}p{0.15\linewidth} | p{0.8\linewidth}@{}}
    \toprule
        \textbf{Fold} & \textbf{Classes} \\
    \midrule
         train & chair, cheeseburger, cow, elephant, hammer, horse, lemon, mushroom, orange, pizza, shark, sheep, tank, teapot, truck, zebra\\
         test & airplane, apple, bear, bicycle, bird, boat, broccoli, car, cat, cauliflower, clock, dog, hourglass, stop sign, strawberry, teddy bear \\ 
    \midrule
         F0 & bear, bicycle, broccoli, cat, cheeseburger, clock, dog, elephant, horse, lemon, orange, shark, sheep, strawberry, tank, teapot \\
F1 & airplane, apple, bicycle, bird, car, cat, chair, cheeseburger, elephant, horse, mushroom, stop sign, strawberry, teapot, teddy bear, zebra \\
F2 & bear, broccoli, car, cat, cheeseburger, hammer, horse, hourglass, lemon, pizza, sheep, stop sign, tank, teapot, teddy bear, zebra \\
F3 & airplane, bear, bicycle, cat, cheeseburger, clock, cow, horse, hourglass, lemon, pizza, shark, tank, teddy bear, truck, zebra \\
F4 & airplane, apple, bear, bird, boat, car, cat, cauliflower, cheeseburger, clock, cow, mushroom, orange, sheep, stop sign, tank \\
\bottomrule
    \end{tabular}
\end{table*}

\begin{table*}[t]
    \caption{\textbf{Sensitivity to class composition.} Accuracy and bias (\%) of \modelemph{Qwen3.5-9B} on each class fold. This ablation uses its own draw of stimuli for every fold, so the test row differs slightly from the main results; this does not affect the comparison between folds.}
    \label{tab:sensitivity_folds}
    \resizebox{\textwidth}{!}{
        \begin{tabular}{@{}lrrrrrrrrrrrrrrrrrrrrrr@{}}
        \toprule
        \multicolumn{1}{c}{}                                   & \multicolumn{9}{c}{\textbf{Spatial}}                                                                        & \multicolumn{4}{c}{\textbf{Chromatic}}                                         & \multicolumn{6}{c}{\textbf{Spectral}}                                                                                      & \multicolumn{3}{c}{\textbf{Semantic}} \\
        \cmidrule(l{5pt}r{5pt}){2-10}\cmidrule(l{5pt}r{5pt}){11-14}\cmidrule(l{5pt}r{5pt}){15-20}\cmidrule(l{5pt}r{5pt}){21-23}  
        \multicolumn{1}{@{}l}{\textbf{Fold}} & \multicolumn{1}{c}{Acc.} & \multicolumn{1}{c}{Left} & \multicolumn{1}{c}{Right} & \multicolumn{1}{|c}{Acc.} & \multicolumn{1}{c}{Top} & \multicolumn{1}{c}{Btm.} & \multicolumn{1}{|c}{Acc.} & \multicolumn{1}{c}{Fg.} & \multicolumn{1}{c}{Bg.} & \multicolumn{1}{c}{Acc.} & \multicolumn{1}{c}{Red} & \multicolumn{1}{c}{Grn.} & \multicolumn{1}{c}{Blue} & \multicolumn{1}{c}{Acc.} & \multicolumn{1}{c}{LF} & \multicolumn{1}{c}{HF} & \multicolumn{1}{c}{Acc.} & \multicolumn{1}{|c}{Amp.} & \multicolumn{1}{c}{Phs.} & \multicolumn{1}{c}{Acc.} & \multicolumn{1}{c}{Obj.} & \multicolumn{1}{c}{Text} \\
        \midrule
        train & 100.0 & 51.6 & 48.4 & 100.0 & 61.7 & 38.3 & 100.0 & 51.7 & 48.3 & 97.5 & 32.6 & 48.0 & 19.4 & 98.3 & 56.8 & 43.2 & 98.2 & 0.1 & 99.9 & 100.0 & 84.4 & 15.6 \\
        test & 100.0 & 51.2 & 48.8 & 100.0 & 60.8 & 39.2 & 100.0 & 51.2 & 48.8 & 98.8 & 32.1 & 47.8 & 20.1 & 99.4 & 54.3 & 45.7 & 97.8 & 0.2 & 99.8 & 100.0 & 87.5 & 12.5 \\
        \midrule
        F0 & 100.0 & 53.1 & 46.9 & 100.0 & 64.0 & 36.0 & 100.0 & 52.7 & 47.3 & 98.4 & 32.8 & 48.4 & 18.8 & 98.8 & 58.4 & 41.6 & 98.7 & 0.1 & 99.9 & 100.0 & 84.8 & 15.2 \\
        F1 & 100.0 & 52.4 & 47.6 & 100.0 & 61.7 & 38.3 & 100.0 & 52.3 & 47.7 & 99.8 & 32.5 & 47.8 & 19.7 & 99.7 & 53.4 & 46.6 & 99.7 & 0.0 & 100.0 & 100.0 & 94.5 & 5.5 \\
        F2 & 100.0 & 52.3 & 47.7 & 100.0 & 61.8 & 38.2 & 100.0 & 51.8 & 48.2 & 98.0 & 32.4 & 48.5 & 19.0 & 98.9 & 57.8 & 42.2 & 98.6 & 0.2 & 99.8 & 100.0 & 87.1 & 12.9 \\
        F3 & 100.0 & 53.9 & 46.1 & 100.0 & 62.6 & 37.4 & 100.0 & 52.8 & 47.2 & 99.0 & 32.3 & 48.4 & 19.2 & 98.8 & 52.4 & 47.6 & 99.6 & 0.1 & 99.9 & 100.0 & 87.6 & 12.4 \\
        F4 & 100.0 & 52.8 & 47.2 & 100.0 & 62.1 & 37.9 & 100.0 & 51.0 & 49.0 & 99.4 & 31.3 & 50.1 & 18.5 & 99.0 & 56.6 & 43.4 & 99.5 & 0.0 & 100.0 & 100.0 & 86.3 & 13.7 \\
        \bottomrule
        \end{tabular}
    }
\end{table*}

\subsection{Option shuffling}\label{sec:appendix:sens_optionshuffle}

\begin{figure*}[t]
    \centering
    \includegraphics[width=0.8\linewidth]{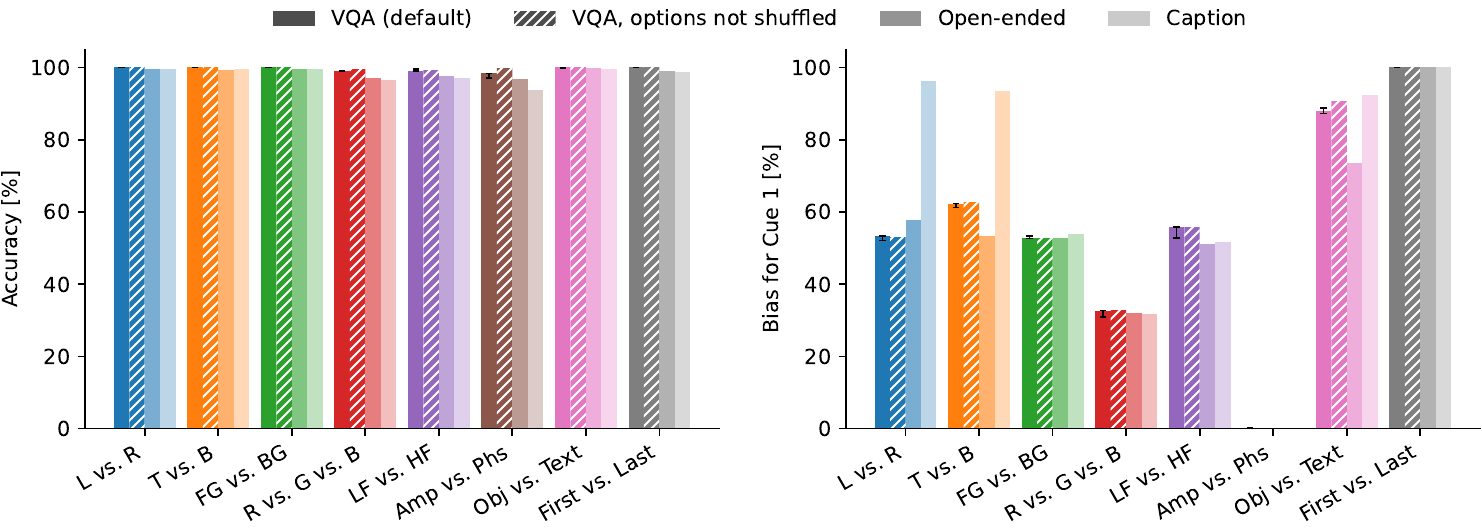}
    \caption{\textbf{Sensitivity to option shuffling and task format.} Accuracy and bias toward the first cue (\cref{eq:bias}) of \modelemph{Qwen3.5-9B} under multiple-choice VQA with shuffled options (default, seed $42$; error bars show the range over the seeds $0$, $1$, and $42$), VQA without option shuffling (\cref{sec:appendix:sens_optionshuffle}), open-ended answering, and captioning (\cref{sec:appendix:sens_taskformat}). For the free-form tasks, the prediction is the first class extracted by the LLM judge. Note that this measures a fundamentally different bias (recency instead of preference) for the caption task.}
    \label{fig:sensitivity_shuffling_task}
\end{figure*}

In our VQA prompts, we randomly shuffle options to reduce the impact of option bias. Shuffling prevents a cue from being tied to a fixed answer position, but the measured biases could still depend on whether and how the options are shuffled. We perform an additional experiment on \modelemph{Qwen3.5-9B} using the derived class composition and hyperparameter from \cref{sec:appendix:sens_classcompandnumsamples}. We record the bias measurements without shuffling and with shuffling, using 3 random seeds: 42 (our seed for all main experiments), 0, and 1. The results in \cref{fig:sensitivity_shuffling_task} (hatched bars and error bars) show minor changes, both between shuffling and no shuffling, and between seeds. Only for the chromatic and semantic conflicts is there slightly greater variation. Overall, we execute all experiments with option shuffling to avoid option biases and conclude that a single seed is sufficient to estimate the bias well on recent models.

\subsection{Task format}\label{sec:appendix:sens_taskformat}

Our main experiments use multiple-choice VQA, which forces the model to choose exactly one class from a fixed set of options. This format may itself induce specific biases. Our study does not aim to quantify biases across arbitrary recognition tasks; rather, we systematically study biases in a simple, yet commonly used task, which we keep fixed throughout. Nonetheless, we test how much our measurements depend on the task format by comparing VQA with two free-form tasks (prompts in \cref{tab:prompts}). The \emph{open-ended} task asks for the name of the shown object, which softly encourages the model to name the most salient class. As it lists no options, it is invariant to the option order by design and serves as a control for VQA-specific biases. The \emph{caption} task asks for a one-sentence caption and does not encourage the model to name a single object. It serves as a control for whether the models recognize all cues of a stimulus at all.

To map the free-form responses to classes, we use \modelemph{Qwen3.8-27B} without thinking as an LLM judge. The judge receives the response and the 16 test classes and returns all classes that the response mentions or unambiguously implies, ordered by their first mention (four in-context examples; sampling with temperature $0.7$, top-$p$ $0.8$, and top-$k$ $20$). We treat the first extracted class as the prediction, assuming that the model mentions the most salient object first, although neither task explicitly enforces this. Note that this introduces a different bias measurement for captioning: recency instead of preference. Accordingly, the accuracy is the fraction of responses whose first extracted class is shown in one of the cues, and the bias follows \cref{eq:bias}. We run the free-form tasks on the test split with 5 stimuli per class combination ($1{,}200$ stimuli per conflict and $16{,}800$ for the chromatic conflict).

\paragraph{Bias across tasks.} \cref{fig:sensitivity_shuffling_task} compares the biases of \modelemph{Qwen3.5-9B} across tasks. In all tasks, almost all answers name a class that is shown in the stimulus. The open-ended task reproduces the VQA biases: every bias keeps its direction, and most change by only a few pp. The largest deviations occur for the top/bottom and object/text conflicts, where the open-ended answers are less biased toward the top and the object, but still lean toward the same cue. Thus, the biases we measure are not an artifact of the multiple-choice format. The captions agree with VQA as well, except on the left/right and top/bottom conflicts, where they almost always mention the left and the top object first. As captions tend to describe a scene in reading order, the first mention there reflects the order of the description rather than a preference.

\begin{figure*}[t]
    \centering
    \includegraphics[width=\linewidth]{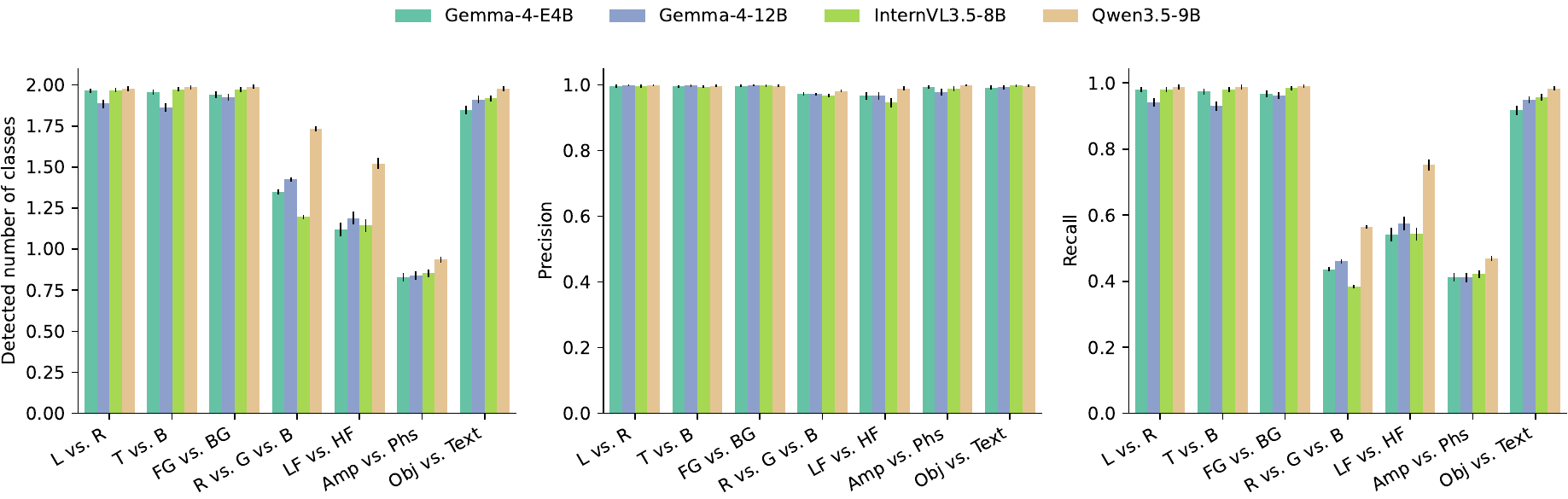}
    \caption{\textbf{Captioning performance.} Number of classes extracted by the LLM judge (left), precision (center), and recall (right) of the captions of four models for each conflict. Bars show the mean and error bars the 95\% confidence interval.}
    \label{fig:task_recall_precision}
\end{figure*}

\paragraph{Recognized cues.} To quantify how many cues the models recognize, we analyze the classes extracted from the captions of all four models in \cref{fig:task_recall_precision} by their number, precision, and recall:
\begin{align}
    \text{Precision} &= \frac{|\text{cues}\cap\text{detected classes}|}{|\text{detected classes}|}, \\
    \text{Recall} &= \frac{|\text{cues}\cap\text{detected classes}|}{|\text{cues}|}.
\end{align}
The precision is at least $0.95$ for all models and conflicts, \ie, the extracted classes are almost always shown in the stimulus, indicating that neither the models nor the judge hallucinate classes. The recall, however, depends strongly on the conflict. For the spatial and semantic conflicts, the captions of all models mention both objects in most cases (recall $\geq 0.92$). For the chromatic and spectral conflicts, it drops to $0.38$--$0.75$: the captions rarely mention all three color channels or both frequency cues, and never the amplitude object. These include the cues that prompting fails to steer (\cref{sec:experiments:language}).

Thus, even when the task does not require a single answer, the models do not mention all cues of these conflicts, suggesting that they fail to recognize some of them.

\begin{figure*}[t]
    \centering
    \includegraphics[width=\linewidth]{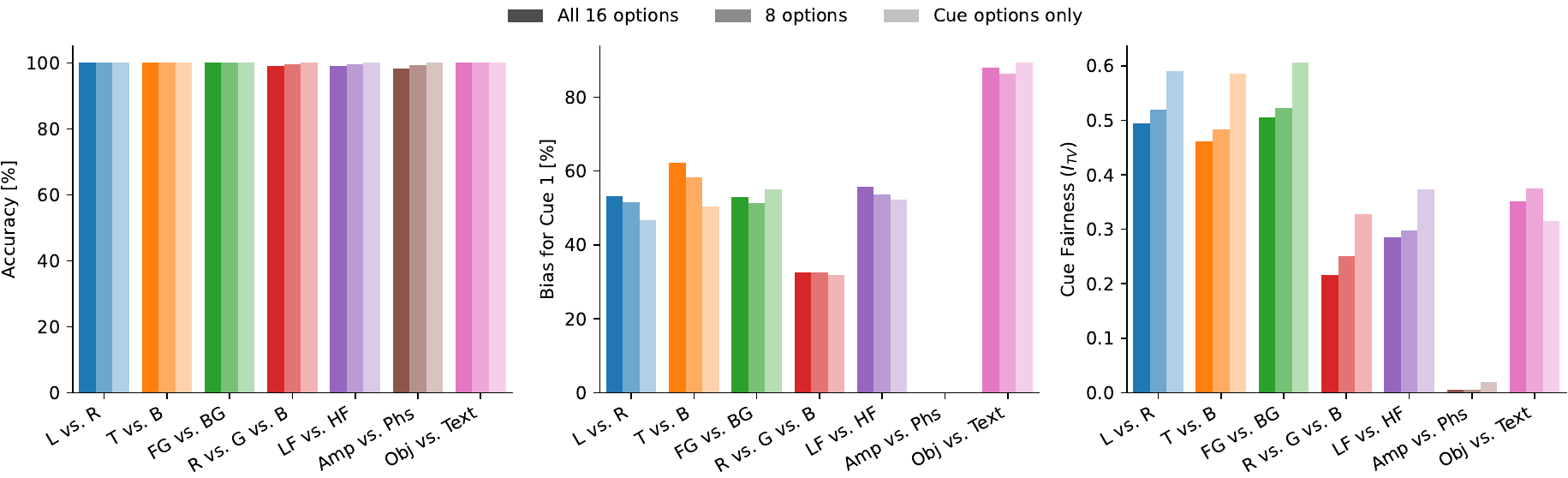}
    \caption{\textbf{Sensitivity to distractor design.} Accuracy, bias toward the first cue (\cref{eq:bias}), and cue fairness (\cref{eq:cue_fairness}) of \modelemph{Qwen3.5-9B} when the options list all $16$ test classes (default), the cue classes filled up with random distractor classes to $8$ options, or only the cue classes. The option order is shuffled in all cases. With cue options only, the accuracy is $100\%$ by construction.}
    \label{fig:sensitivity_distractor}
\end{figure*}

\subsection{Distractor design}\label{sec:appendix:sens_distractor}

In all main experiments, every question lists all $16$ test classes as answer options, \ie, the $N$ cue classes of the stimulus and $16-N$ distractors. To test how much the measured biases depend on this design, we evaluate \modelemph{Qwen3.5-9B} with two smaller option sets: the $N$ cue classes filled up with randomly drawn distractor classes to a total of $8$ options, which is also the format of our human study (\cref{sec:appendix:humans}), and only the $N$ cue classes without any distractors. Distractors are drawn from the test classes for every stimulus, and the option order is shuffled in all cases; the setup is otherwise identical to the main experiments (seed $42$). \cref{fig:sensitivity_distractor} shows the results.

The model rarely picks a distractor: at most $1.7\%$ of the answers name a class that is not in the stimulus with $16$ options (amplitude/phase) and at most $0.6\%$ with $8$ options. Without distractors, every answer names a cue by construction.

Reducing the options to $8$ barely changes the measurement. All biases stay within $1.6$ pp of the $16$-option setting, except for the top cue ($62.3\%$ \vs $58.4\%$), and the mean cue fairness rises only slightly from $0.33$ to $0.35$.
Removing all distractors has a larger effect, but only on the weak biases. Strong biases are unaffected: phase stays at $99.9\%$, the color biases change by at most $0.8$ pp, and the object bias changes from $88.0\%$ to $89.3\%$. In contrast, the weak spatial biases shift toward or past chance: the top bias drops from $62.3\%$ to $50.4\%$, and the left bias of $53.3\%$ turns into a right bias ($46.7\%$ left). The cue fairness increases from $0.33$ to $0.40$ on average, mostly for the spatial ($0.46$--$0.51$ \vs $0.59$--$0.61$) and chromatic ($0.22$ \vs $0.33$) conflicts, \ie, the model spreads its probability more evenly over the cues once they are the only options.

An option list that contains only the cue classes reveals which classes are present in the stimulus and turns the classification into a forced choice between them, which may tip near-balanced preferences. We therefore list all $16$ classes in our experiments, which does not reveal the cue classes. Since the $8$-option format yields biases within $3.9$ pp of the $16$-option format, the different option sets of the human study and the model evaluations are unlikely to have affected the differences between human and model biases.

\subsection{Prompt formulation}\label{sec:appendix:sens_prompt}

\begin{figure*}[t]
    \centering
    \includegraphics[width=\linewidth]{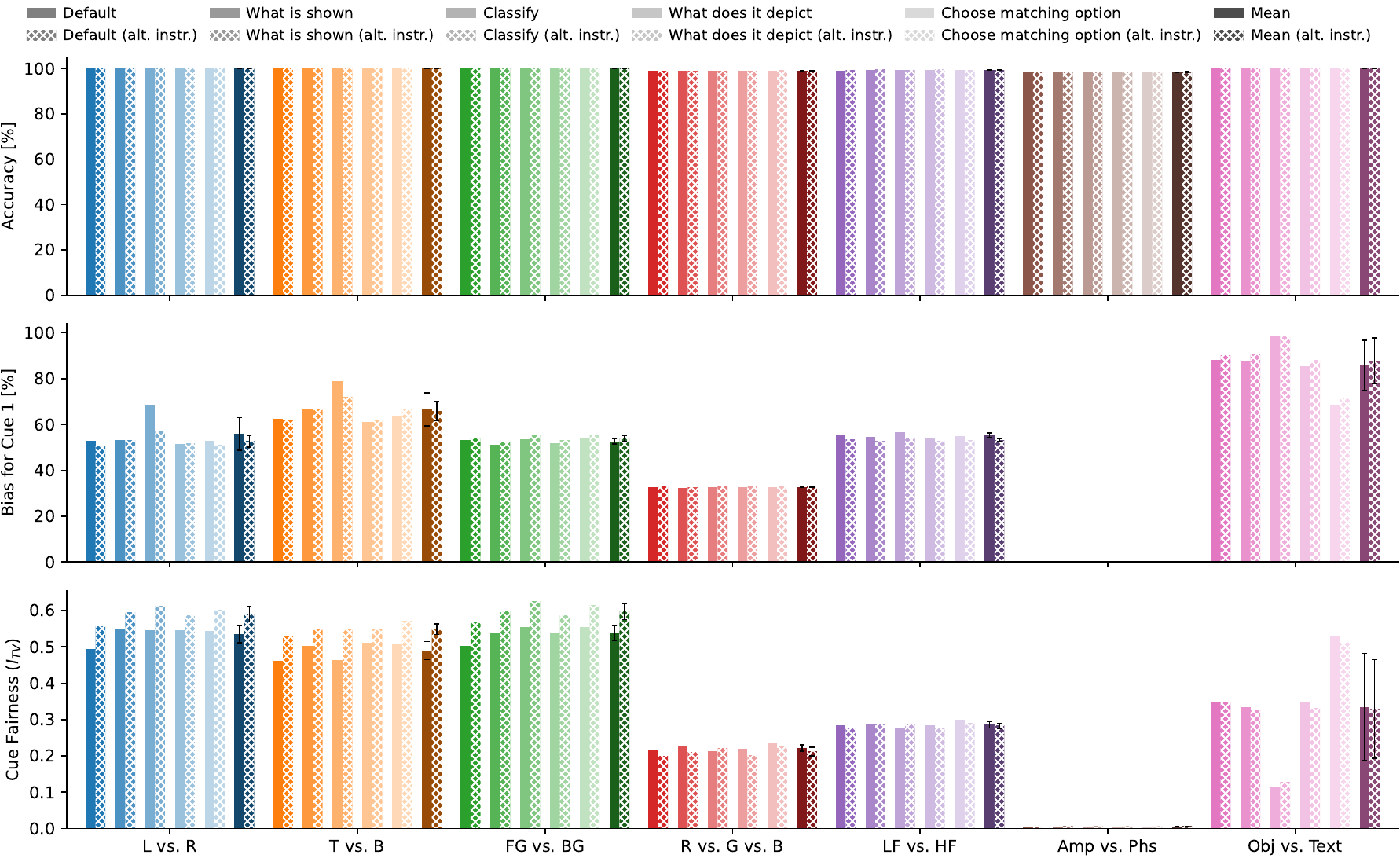}
    \caption{\textbf{Sensitivity to the prompt formulation.} Accuracy, bias toward the first cue (\cref{eq:bias}), and cue fairness (\cref{eq:cue_fairness}) of \modelemph{Qwen3.5-9B} for the five neutral questions, each with the default (solid) and the alternative (hatched) answer instruction (\cref{tab:neutral_prompts}). The darkest bars show the mean over the five questions for each instruction; error bars show the standard deviation over the questions.}
    \label{fig:sensitivity_prompt}
\end{figure*}

All bias measurements in the main paper use a single neutral prompt (\cref{tab:prompts}). To test how much these measurements depend on its exact wording, we evaluate \modelemph{Qwen3.5-9B} with four alternative neutral questions that avoid words referring to any of the tested cues (\eg, ``object'' or ``foreground''). We combine each of the five questions with the default and an alternative answer instruction, resulting in $5\times2$ prompts (\cref{tab:neutral_prompts}). Only the question and the answer instruction change; the setup is otherwise identical to the main experiments (16 test classes, all 16 options in shuffled order, seed 42). \cref{fig:sensitivity_prompt} shows the results.

\begin{table}[t]
    \centering
    \caption{\textbf{Neutral prompt formulations.} Each question is combined with both answer instructions, resulting in ten prompts. The prompts follow the VQA template in \cref{tab:prompts}: the question, followed by the shuffled options and the answer instruction. The default question and instruction are used for all other results.}
    \label{tab:neutral_prompts}
    \small
    \begin{tabular}{@{}>{\raggedright\arraybackslash}p{0.29\linewidth} | >{\raggedright\arraybackslash}p{0.65\linewidth}@{}}
        \toprule
        \textbf{Question} & \textbf{Prompt text} \\
        \midrule
        Default & Which option best describes the image? Options: \\
        What is shown & What is shown in this image? Options: \\
        Classify & Classify the image into one of the following categories: \\
        What does it depict & What does this image depict? Options: \\
        Choose matching option & Choose the option that matches this image. Options: \\
        \midrule
        \textbf{Answer instruction} & \\
        \midrule
        Default & Answer only with the correct option letter. \\
        Alternative & Answer with the option's letter from the given choices directly. \\
        \bottomrule
    \end{tabular}
    \vspace{-0.5em}%
\end{table}

\paragraph{Accuracy.} The prompt formulation does not affect the accuracy. It stays between $98.2\%$ and $100\%$ for all prompts and varies by at most $0.4$ pp within any conflict.

\paragraph{Bias.} Some formulations strongly change the strength of the bias in the left/right, top/bottom, and object/text conflicts. Across the ten prompts, the left bias ranges from $50.7\%$ to $68.6\%$, the top bias from $61.1\%$ to $78.9\%$, and the object bias from $68.8\%$ to $99.0\%$. The ``Classify'' question, which resembles an image classification task, causes the strongest bias in all three conflicts; with the alternative instruction, this effect is weaker for the spatial conflicts (left: $68.6\% \to 57.0\%$, top: $78.9\% \to 72.0\%$). ``Choose the option that matches this image'' causes the weakest object bias. However, no formulation changes the direction of a bias: in every conflict, the dominant cue is the same for all ten prompts. For the remaining conflicts, the biases are similar across questions and instructions: they vary by at most $4.5$ pp for the foreground/background and low/high-frequency conflicts, $1.3$ for the chromatic conflict, and $0.1$ for the amplitude/phase conflict.

\paragraph{Cue fairness.} The alternative instruction increases the cue fairness for the conflicts with the fairest cues: for left/right, top/bottom, and foreground/background (mean $\mathrm{I}_{\mathrm{TV}}$ of $0.49$--$0.54$ with the default instruction), the fairness increases by $0.055$--$0.060$ on average and for every question ($+0.04$ to $+0.09$). This also holds where the top-1 bias barely changes (\eg, top: $62.4\%$ \vs~$62.0\%$ for the default question), \ie, the model spreads more probability over the cues without changing its cue preference. For other conflicts, the fairness decreases slightly on average, though this varies by question.

Overall, the formulation of a neutral prompt can modulate the strength of some biases, but it does not change which cue a model prefers. We therefore evaluate every model with the same single default prompt rather than averaging over formulations. A fixed prompt means the differences between models come from the models and not from the prompt, and it matches how VLMs are queried in practice, where a user poses one question rather than an ensemble of paraphrases. Averaging would also hide the prompt-dependent variation shown above, which is itself a property of the model.

\begin{figure*}[t]
    \centering
    \includegraphics[width=\linewidth]{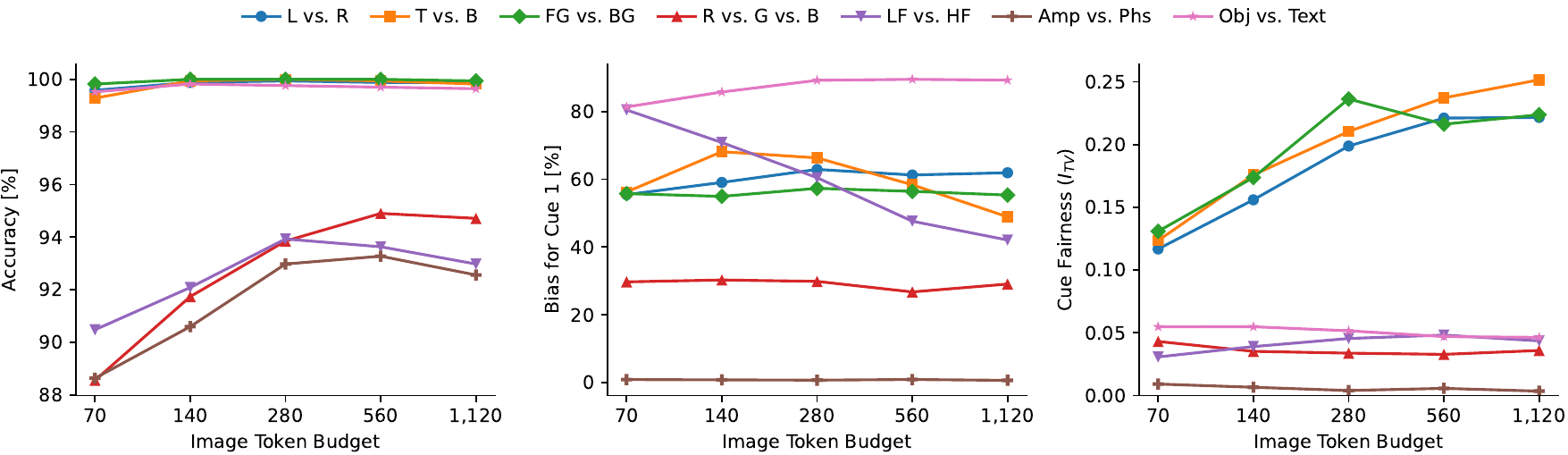}
    \caption{\textbf{Sensitivity to the visual token granularity.} Accuracy, bias toward the first cue (\cref{eq:bias}), and cue fairness (\cref{eq:cue_fairness}) of \modelemph{Gemma-4-E4B} for different budgets of visual tokens per image (log scale). $1{,}120$ tokens is the default budget used in all other experiments.}
    \label{fig:vqa_image_budget}
\end{figure*}

\subsection{Visual token granularity}\label{sec:appendix:token_budget}

Although all models receive the same $448{\times}448$ px stimuli, each divides them into a different number of visual tokens, so that one token summarizes a differently sized region of the stimulus (\cref{tab:probe_shapes}). This granularity may matter most for cues that live in fine image details, such as high frequencies. To isolate its effect from all other differences between models, we vary the number of visual tokens of a single model with fixed weights. 

\modelemph{Gemma-4} lets the user cap the number of visual tokens per image at $70$, $140$, $280$, $560$, or $1{,}120$ tokens; its image processor resizes the image so that the resulting patch grid fits this budget. A lower budget, thus, does not crop the image or shorten its context, but makes every patch, and hence every token, cover a larger region of the stimulus. All other experiments use the default budget of $1{,}120$ tokens, which yields $99{\times}99$ patches and, after $3{\times}3$ pooling, $1{,}089$ visual tokens for our stimuli; the stimuli are therefore upsampled, and one token covers roughly $14{\times}14$ stimulus pixels. Here, we evaluate \modelemph{Gemma-4-E4B} with all lower budgets under the neutral prompt and otherwise identical settings. All other models use their default image preprocessing throughout. \cref{fig:vqa_image_budget} shows the results; the biases toward all cues are part of \cref{tab:vlm_bias_full}.

\paragraph{Accuracy.} As expected, the accuracy improves slightly with a larger budget.

\begin{figure*}[t]
    \centering
    \setlength{\tabcolsep}{0.5pt}
    \resizebox{0.75\textwidth}{!}{
        \begin{tabular}{ccc}
             \includegraphics[width=0.58\linewidth]{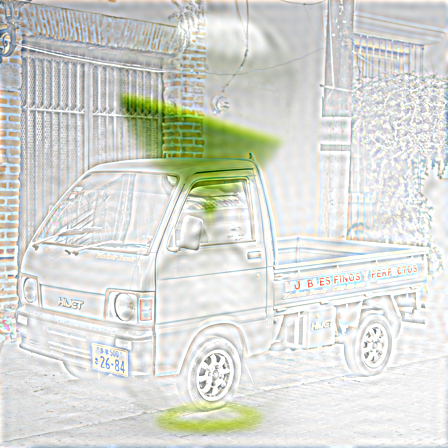} & 
             \includegraphics[width=0.28\linewidth]{figures/examples/spectral-lf-hf-hourglass-truck.png} & 
             \includegraphics[width=0.14\linewidth]{figures/examples/spectral-lf-hf-hourglass-truck.png} 
        \end{tabular}%
    }%
    \caption{\textbf{Viewing distance and frequency perception.} The same low/high-frequency stimulus at decreasing sizes, which mimics an increasing viewing distance. Up close, the high-frequency cue (truck) dominates; with increasing distance, it fades and only the low-frequency cue (hourglass) remains visible \cite{schyns1994blobs,aude2006hybrid}.}
    \label{fig:lfhf_viewingdistance}
\end{figure*}

\paragraph{Bias.} The budget mainly affects the low/high-frequency conflict. With fewer tokens, the model increasingly relies on the low-frequency cue: its low-frequency bias rises from $42.1\%$ at the default budget to $47.6\%$, $60.5\%$, $70.8\%$, and $80.5\%$ at $560$, $280$, $140$, and $70$ tokens, \ie, the dominant cue flips from the high- to the low-frequency cue. With fewer tokens, each token summarizes a larger region of the stimulus, so the finest image details are the first to be averaged out, much like an increased viewing distance in humans (\cref{fig:lfhf_viewingdistance}); at $70$ tokens, the model is about as biased toward low frequencies as our human participants ($79.5\%$). Below $280$ tokens, the object/text conflict also shifts toward the text, from $10.8\%$ at the default budget to $14.3\%$ and $18.7\%$ at $140$ and $70$ tokens. The top bias changes non-monotonically: it rises from $48.9\%$ at the default budget to $68.1\%$ at $140$ tokens and falls back to $56.2\%$ at $70$ tokens. The left bias varies less ($55.5$--$62.9\%$), and the foreground/background, chromatic, and amplitude/phase biases stay within $2.4$ pp of the default for all cues.

\paragraph{Cue fairness.} For the spatial conflicts, a lower budget reduces the cue fairness substantially: from the default budget to $70$ tokens, it roughly halves for left/right ($0.22 \to 0.12$), top/bottom ($0.25 \to 0.12$), and foreground/background ($0.22 \to 0.13$). This happens although the top-1 bias of these conflicts barely changes (\eg, left: $61.9\%$ \vs~$55.5\%$), \ie, with fewer tokens, the model is more confident in its answer without preferring a different cue. For the remaining conflicts, the fairness is low at all budgets ($\mathrm{I}_{\mathrm{TV}} \leq 0.055$). In particular, the fairness of the low/high-frequency conflict stays between $0.03$ and $0.05$ while its bias flips, \ie, at every budget, the model decides confidently for one of the two cues.

The visual token granularity is thus a confound for frequency biases in particular: within a single model, coarser tokens are enough to flip the dominant frequency cue. Comparisons of frequency biases across models should therefore take into account how large a region of the stimulus each model's visual tokens cover, although our single-model experiment does not show how much of the observed differences between models this explains. Beyond the top-1 bias, the granularity also affects how confidently the model chooses between the spatial cues.

\begin{figure*}[t]
    \centering
    \setlength{\tabcolsep}{0.5pt}
    \resizebox{\textwidth}{!}{
        \begin{tabular}{cccccc}
             \includegraphics[width=0.2\linewidth]{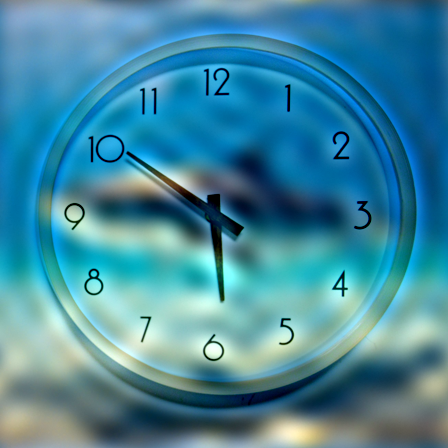} & 
             \includegraphics[width=0.2\linewidth]{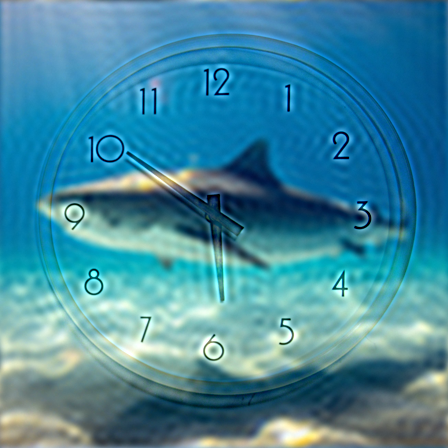} & 
             \includegraphics[width=0.2\linewidth]{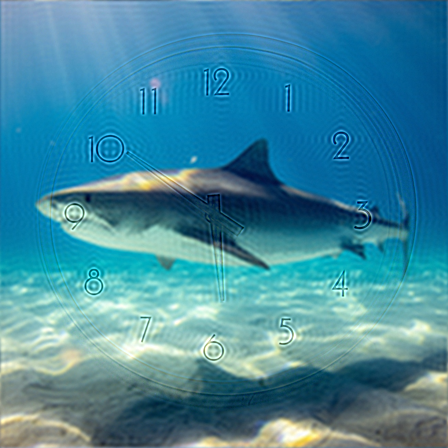} & 
             \includegraphics[width=0.2\linewidth]{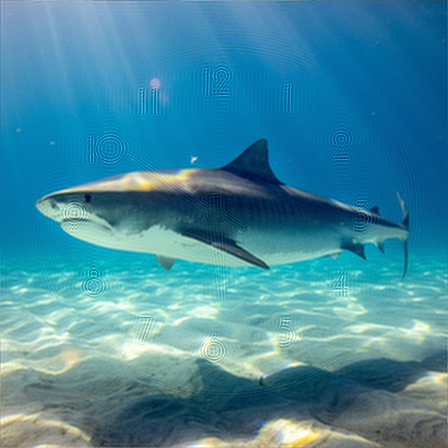} & 
             \includegraphics[width=0.2\linewidth]{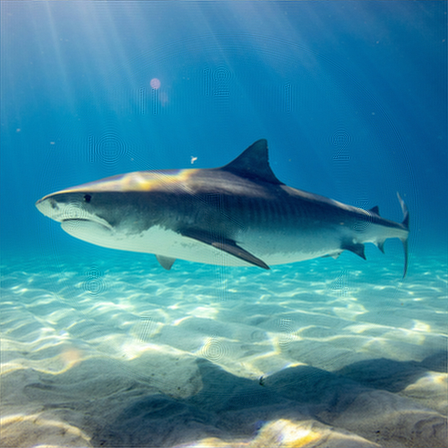} & 
             \includegraphics[width=0.2\linewidth]{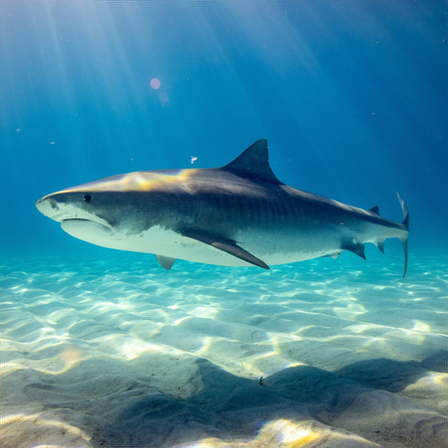} \\
             0.05 & 0.15 & 0.3 & 0.5 & 0.7 & 0.85
        \end{tabular}%
    }%
    \caption{\textbf{Frequency cutoff.} The same low/high-frequency stimulus at different cutoffs, with the low-frequency cue \textit{shark} and the high-frequency cue \textit{clock}. The clock dominates at $0.05$, both objects are visible at $0.15$, and the clock fades as the cutoff grows. The higher the cutoff, the larger the share of the spectrum taken from the low-frequency image. All main results use a cutoff of $0.15$.}
    \label{fig:lfhf_threshold}
\end{figure*}

\subsection{Frequency cutoff}\label{sec:appendix:lfhf_weep}

\begin{figure}[!tb]
    \centering
    \includegraphics[width=\linewidth]{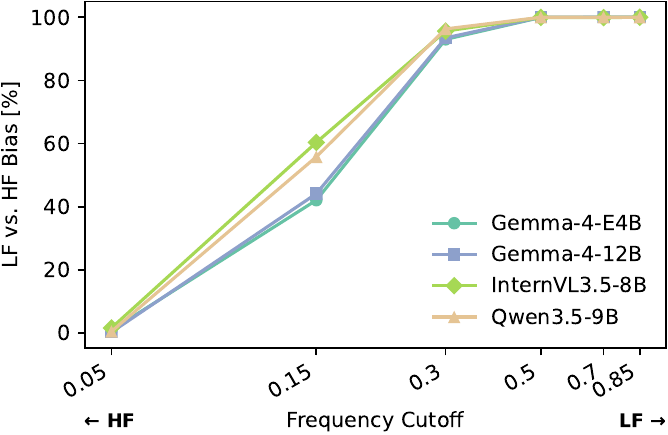}
    \caption{\textbf{Low- \vs~high-frequency bias \vs~frequency cutoff.} Bias toward the low-frequency cue (\%) on low/high-frequency stimuli generated with different cutoffs (log scale). The low-frequency cue keeps all frequencies below the cutoff (as a fraction of the maximum frequency along each axis), and the high-frequency cue keeps the rest. All main results use a cutoff of $0.15$.}
    \label{fig:vqa_lfhf_bias_vs_threshold}
\end{figure}

The low/high-frequency conflict depends on the cutoff that separates the two frequency bands. All main results use a cutoff of $0.15$, \ie, the low-frequency cue keeps all frequencies below $15\%$ of the maximum frequency along each axis, and the high-frequency cue keeps all remaining frequencies (\cref{sec:method:stimuli}). Raising the cutoff shifts a larger share of the spectrum to the low-frequency image (\cref{fig:lfhf_threshold}). Here, we sweep the cutoff over $\{0.05, 0.15, 0.3, 0.5, 0.7, 0.85\}$ for our four main models on the test classes and report results.

\cref{fig:vqa_lfhf_bias_vs_threshold} shows that the bias largely follows the share of the spectrum that each cue receives. At a cutoff of $0.05$, all models choose the high-frequency cue in at least $98.7\%$ of the valid answers; at $0.3$, they choose the low-frequency cue in $92.7$--$96.4\%$; and from $0.5$ on, in at least $99.9\%$. In between, at our cutoff of $0.15$, are the models close to undecided ($42.1$--$60.4\%$ low-frequency bias), and this is also the only tested cutoff where they differ substantially. The absolute low- \vs~high-frequency bias of a model is therefore only meaningful with respect to the cutoff. This does not affect our comparisons between models and between steering methods, which all use the same cutoff.

The same holds for humans, for whom the perceived band additionally depends on the viewing distance and the size of the stimulus \cite{schyns1994blobs,aude2006hybrid}: up close, a hybrid image is dominated by its high frequencies, and from afar by its low frequencies (\cref{fig:lfhf_viewingdistance}). Our participants viewed the stimuli on mobile devices (\cref{sec:appendix:humans}), which likely contributes to their strong preference for the low-frequency cue at the same cutoff ($79.5\%$).

\section{Cue fairness across model scale}\label{sec:appendix:fairness_scale}

\begin{figure*}[t]
    \centering
    \includegraphics[width=\linewidth]{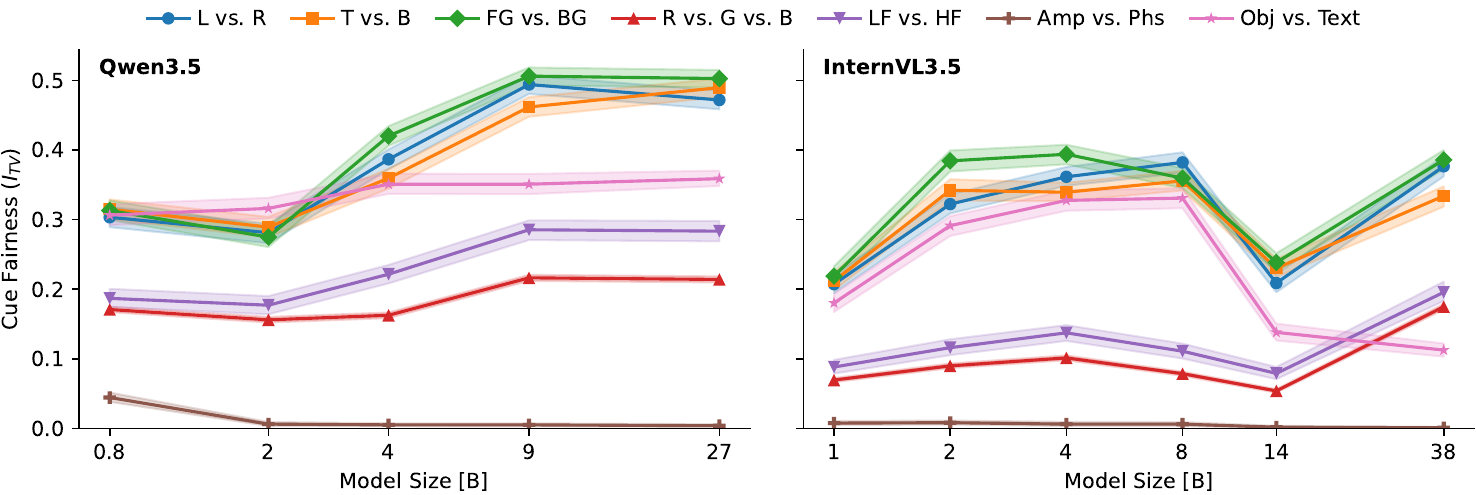}
    \caption{\textbf{Cue fairness \vs~model scale} for \modelemph{Qwen3.5} (left) and \modelemph{InternVL3.5} (right). Lines show the mean cue fairness ($\mathrm{I}_{\mathrm{TV}}$) over all samples, and shaded areas denote the 95\% confidence interval; the model size is shown on a log scale. Higher is closer to uniform (fair).}
    \label{fig:vqa_cue_fairness_scale}
\end{figure*}

In \cref{sec:experiments:default:probs}, we analyzed cue fairness (\cref{eq:cue_fairness}) for our four main models. Here, we study how fairness evolves with model scale by evaluating all available sizes of two model families: \modelemph{Qwen3.5} (0.8B, 2B, 4B, 9B, and 27B) and \modelemph{InternVL3.5} (1B, 2B, 4B, 8B, 14B, and 38B). The results are shown in \cref{fig:vqa_cue_fairness_scale}.

For \modelemph{Qwen3.5}, the mean fairness over all conflicts increases from $0.21$ at 2B to $0.33$ at 9B and stays at $0.33$ for 27B (0.8B: $0.23$). The gain comes mostly from the spatial conflicts (\eg, foreground/background: $0.28 \to 0.51$) and the low/high-frequency conflict ($0.18 \to 0.29$), whereas the chromatic conflict improves only slightly ($0.16 \to 0.22$). In contrast, \modelemph{InternVL3.5} shows no clear trend: its mean fairness ranges from $0.14$ to $0.24$, and the 14B model is about as unfair as the 1B model ($0.14$ each). For both families and all sizes, the fairness stays low for amplitude/phase ($\leq 0.05$), and no conflict comes close to uniform cue confidences ($\mathrm{I}_{\mathrm{TV}} \leq 0.51$). Scaling alone thus does not resolve the unimodal cue preference.

\section{Steering via in-context learning}\label{sec:appendix:icl_explain}

\begin{figure*}[t]
    \centering
    \includegraphics[width=\linewidth]{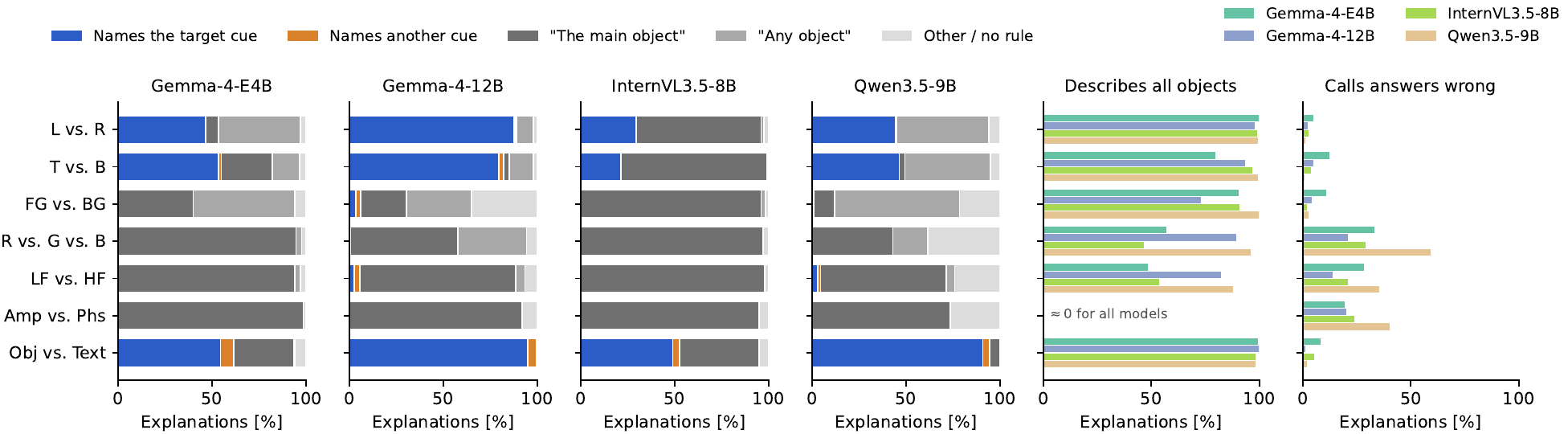}
    \caption{\textbf{What models infer from in-context examples.} Left: the rule stated by each model after describing the five in-context examples, as classified by an LLM judge (100 contexts per conflict and targeted cue, pooled over the targeted cues). Right: share of explanations that describe all conflicting objects in most examples, and that claim some of the given answers are wrong.}
    \label{fig:icl_explain_rules}
\end{figure*}

In-context learning barely steers any bias (\cref{sec:experiments:default:icl}). To find out why, we ask the models to put into words what they take away from the in-context examples, and compare the stated rules with their actual ICL behavior.

\paragraph{Setup.} For every model, conflict, and targeted cue, we take 100 evenly spaced test stimuli of the 5-shot ICL experiment and rebuild their five solved examples exactly (same stimuli, option order, and answer letters). Instead of the test stimulus, we append an instruction to describe each example, explain its answer, and state the general rule behind all answers in one sentence (\cref{tab:prompts}). Note that the instruction reveals that such a rule exists, so the explanations show whether a model \emph{can} induce the cue when asked, which is easier than inducing it implicitly. We sample one response per context (temperature $1.0$, top-$p$ $0.95$, at most $2{,}048$ new tokens). An LLM judge (\modelemph{Qwen3.8-27B} without thinking) is told how the stimuli of the conflict are built and which cue all example answers followed. It classifies the stated rule (the target cue, another cue, ``the main object'', ``any object'', or another or no rule), whether the descriptions mention all conflicting objects, and whether the model claims that some given answers are wrong.

\begin{figure}[t]
    \centering
    \includegraphics[width=\linewidth]{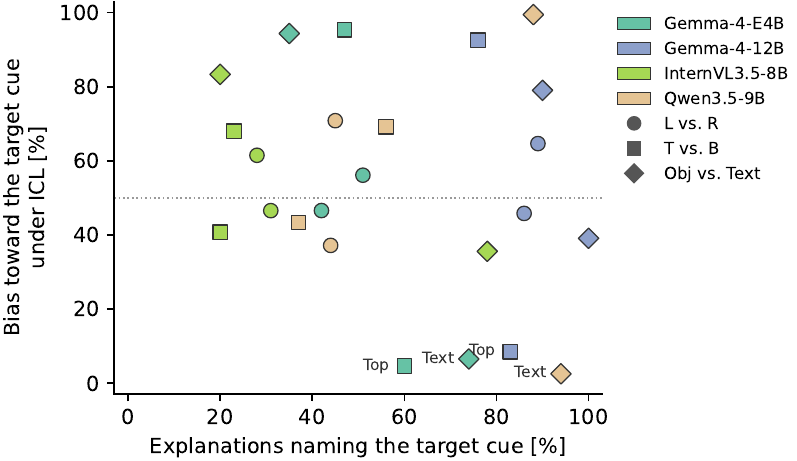}
    \caption{\textbf{Stating a rule \vs~applying it.} Each point is one model and targeted cue. Horizontal: share of explanations whose rule names the targeted cue. Vertical: bias (\cref{eq:bias}) toward the targeted cue under 5-shot ICL (\cref{tab:vlm_bias}). The dotted line marks a balanced bias; labeled points are named often but rarely selected.}
    \label{fig:icl_explain_stated_vs_applied}
\end{figure}

\paragraph{Results.} \cref{fig:icl_explain_rules} shows three distinct failure modes.
\emph{Not perceived:} for amplitude/phase, the models only ever describe the phase object. They cannot induce a cue they do not see, and when the examples follow the amplitude cue, they often conclude that the given answers are wrong.
\emph{Not inferred:} for the foreground/background, chromatic, and low/high-frequency cues, the models mostly do describe all objects, but practically never name the distinguishing cue. Instead, they fall back to generic rules, such as picking the most prominent or any of the shown objects.
\emph{Not applied:} for the left/right, top/bottom, and object/text cues, the rule is named frequently, especially by \modelemph{Gemma-4-12B}. Yet, naming the rule barely translates into behavior (\cref{fig:icl_explain_stated_vs_applied}). \Eg, \modelemph{Gemma-4-12B} names the top/bottom rule in most explanations, yet answers with the bottom cue regardless of which cue the examples target.
Thus, ICL fails for different reasons on different cues: perception limits the spectral amplitude cue, the induction of a nameable rule limits most other cues, and even an induced rule is not reliably applied.

\section{Human study}\label{sec:appendix:humans}

To compare the biases of \vlms{} to those of human observers, we collected human responses on the same cue-conflict stimuli through \url{rapidata.ai}, a platform that shows short annotation tasks to the users of partner mobile apps.

\paragraph{Stimuli.} We use all seven conflicts on the 16 test classes (\cref{tab:class_splits}), generated with the same data loader and seed as for the models: $1{,}680$ stimuli for each two-cue conflict and $10{,}080$ for the chromatic conflict, \ie, $20{,}160$ stimuli in total. The chromatic stimuli were shuffled and split into six batches of $1{,}680$ stimuli each.

\begin{figure}[t]
    \centering
    \includegraphics[width=\linewidth]{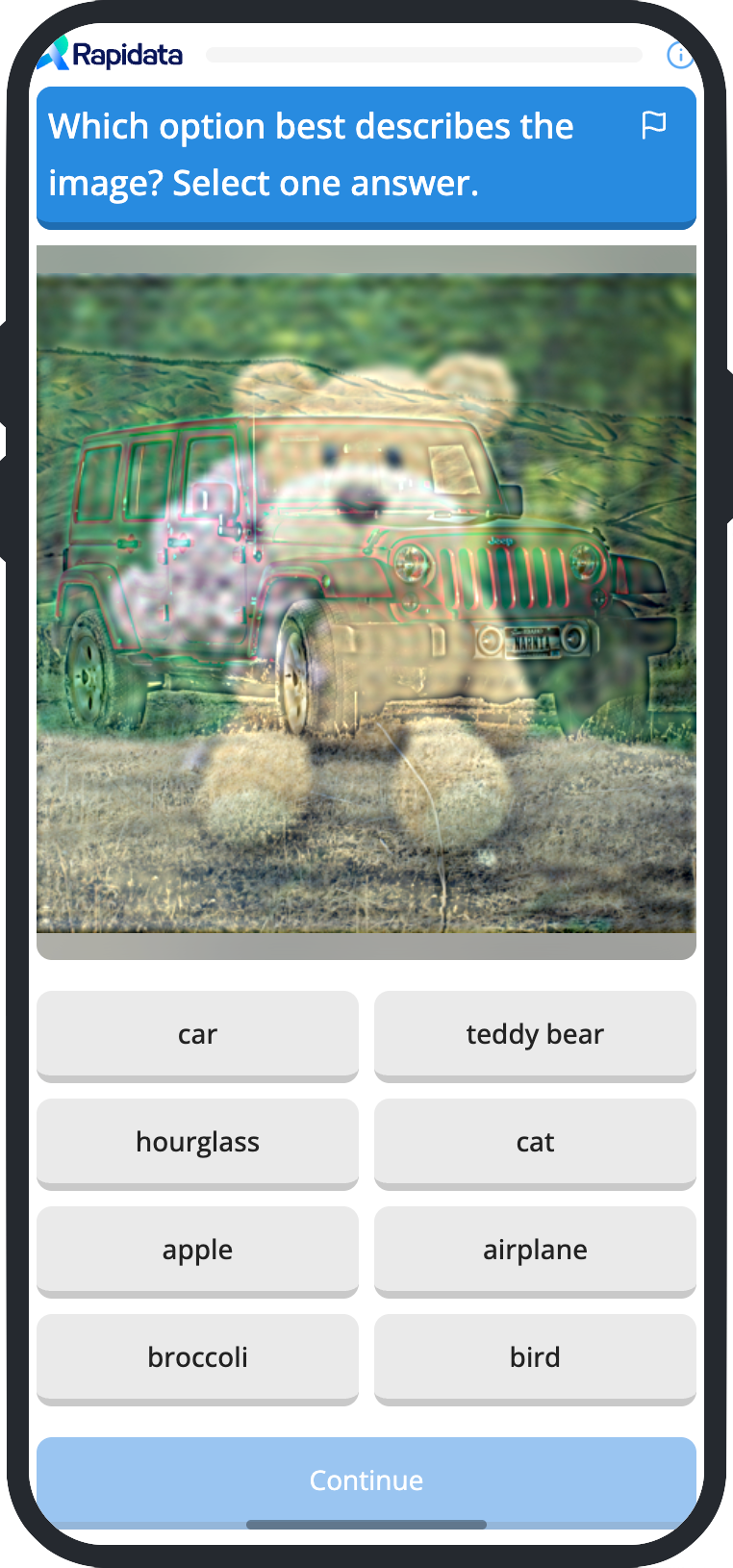}
    \caption{\textbf{Interface of the human study.} This example shows the user interface for one random low/high-frequency stimulus. }
    \label{fig:app_rapidate}
\end{figure}

\paragraph{Task.} Participants saw one stimulus at a time with the instruction ``\texttt{Which option best describes the image? Select one answer.}'' and eight answer options: the classes of all cues in the stimulus, filled up with randomly drawn distractor classes from the test classes, in shuffled order (\cref{fig:app_rapidate}). This corresponds to the eight-option variant in \cref{sec:appendix:sens_distractor}. The study was open to the platform's global audience. By default, the platform translates instructions and options into the language of each participant's device; we disabled this translation, so all participants saw the English text. Only $33.8\%$ of the responses come from devices set to English (60 device languages in total), but the options are common object names that we expect most participants to understand.

\paragraph{Responses.} We collected $10$ responses per stimulus, \ie, $201{,}600$ responses from $54{,}122$ distinct annotator IDs in $140$ reported countries and territories. IDs are platform accounts rather than verified individuals. Most IDs contribute only a few responses (median $2$), no single ID contributes more than $0.8\%$ of all responses, and $14$ stimuli received two responses from the same ID. We compute the human bias with \cref{eq:bias} over all responses; as every stimulus has exactly $10$ responses, all stimuli are weighted equally. Between $9.3\%$ and $14.7\%$ of the responses per conflict name a distractor class and are excluded by \cref{eq:bias}. Aggregating each stimulus by its majority vote instead strengthens every bias without changing its direction (\eg, $91.4\%$ instead of $79.5\%$ for the low-frequency cue and $69.0\%$ instead of $59.5\%$ for the foreground). The left bias does not depend on the reading direction of the participants' country: participants from countries with right-to-left scripts choose the left cue in $51.1\%$ of cases, compared to $51.2\%$ for all others. Country is only a coarse proxy for reading habits, so we do not draw conclusions from this comparison.

\paragraph{Limitations.} As the study runs on participants' own devices, we cannot control for visual impairments or viewing conditions, such as viewing distance, screen size, display calibration, or color filters applied by the participants. Viewing distance and stimulus size in particular affect the frequency conflict: up close, the high-frequency cue becomes more visible, whereas from afar, only the low-frequency cue remains \cite{schyns1994blobs} (\cref{fig:lfhf_viewingdistance}). Stimuli shown on a small mobile screen thus likely favor the low-frequency cue, and the human bias on this conflict is, like that of the models, only meaningful for our cutoff (\cref{sec:appendix:lfhf_weep,fig:lfhf_threshold}). Similarly, color vision deficiencies and display settings, such as blue-light filters, may affect the chromatic conflict. With $10$ responses per stimulus from a large and diverse pool of participants, we expect individual conditions to average out, but we do not claim that our human measurements are free of such biases.

\paragraph{Ethics statement.}
\iftoggle{cvprfinal}{%
The human study was reviewed and approved by the Institutional Biosafety and Bioethics Committee (IBEC) of King Abdullah University of Science and Technology (KAUST) under IBEC protocol No.~26IBEC032. IBEC is registered with the Saudi National Committee of Bioethics (NCBE) under number HAP-02-J-042.}{
The study was approved by our institution's ethics review board (protocol number withheld for anonymity).
}

\section{Steering tokens}\label{sec:appendix:steering_tokens}

\begin{figure*}[t]
    \centering
    \input{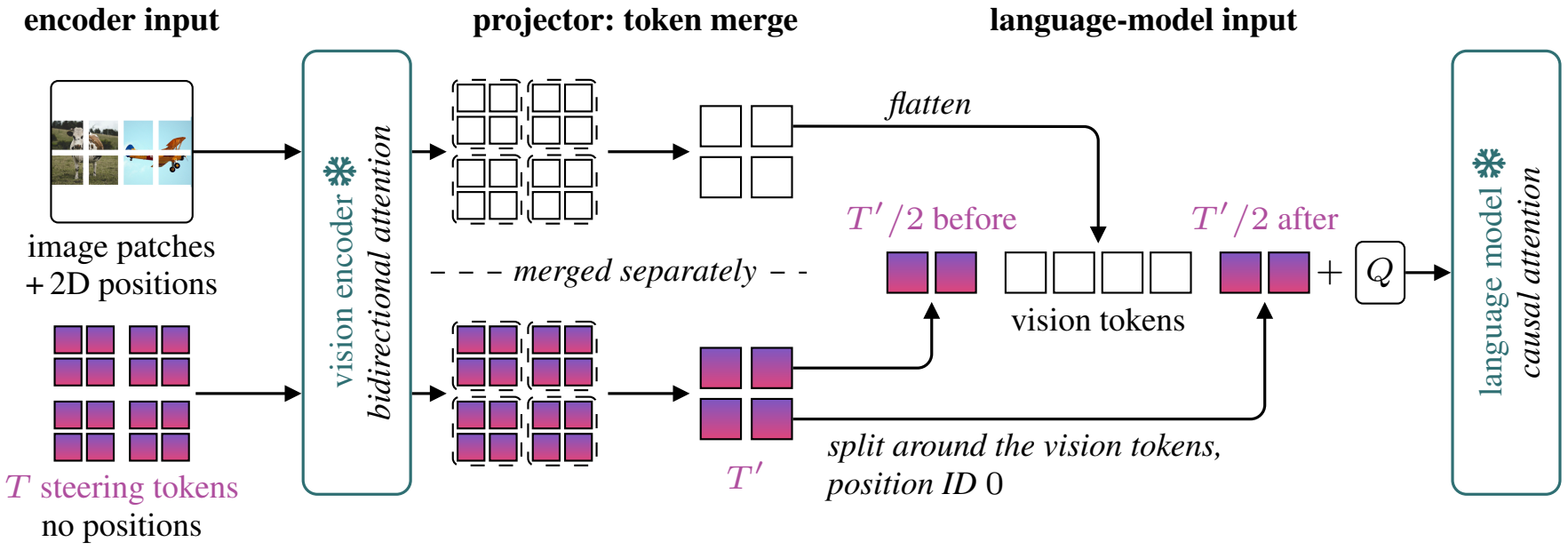}
    \caption{\textbf{Injecting \ourtokens.} $T$ learned tokens are appended to the image patches and attend jointly with them in the frozen vision encoder. The model's projector merges neighboring tokens within each stream separately, and the resulting $T'$ projected \ourtokens{} are inserted half before and half after the vision tokens in the language-model input, with position ID $0$.}
    \label{fig:steering_tokens}
\end{figure*}

\begin{figure*}[t]
    \centering
    \includegraphics[width=\linewidth]{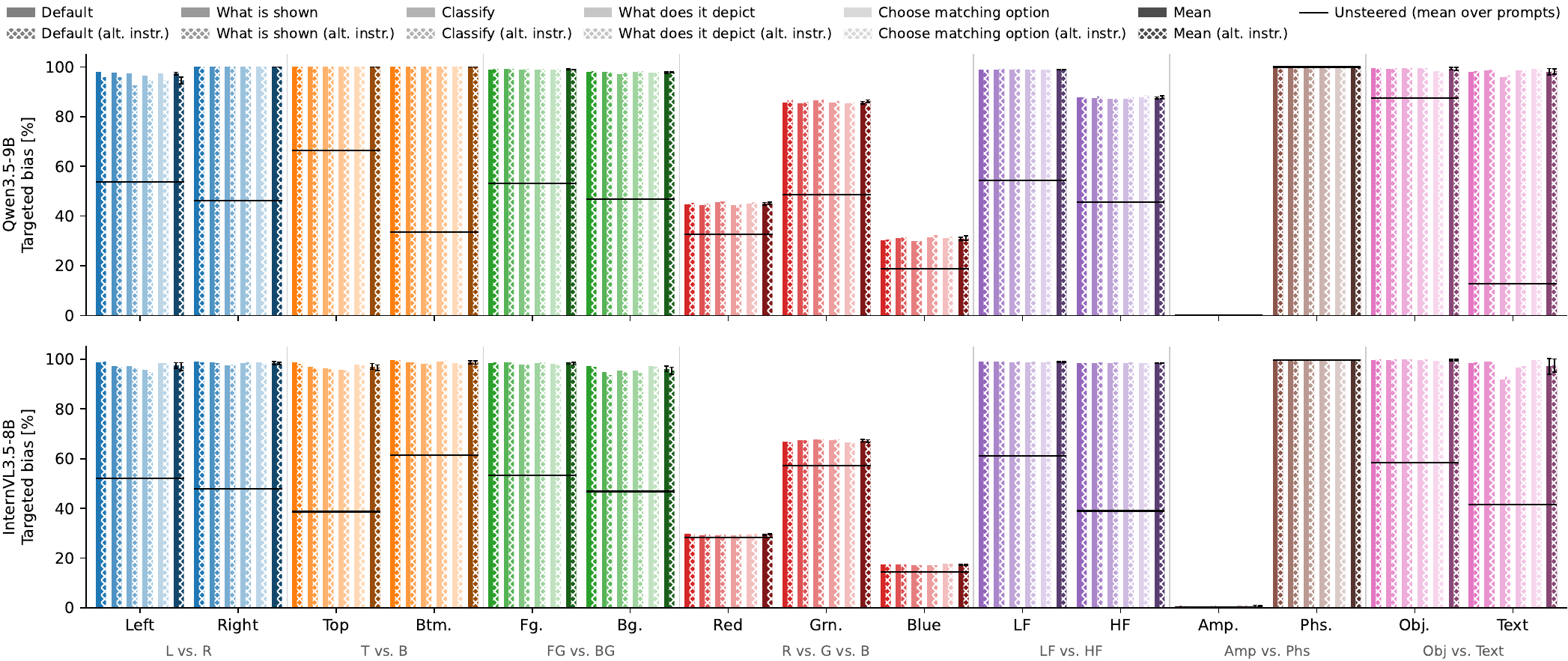}
    \caption{\textbf{Transfer of \ourtokens{} to other prompts.} Bias (\cref{eq:bias}) toward the targeted cue of the \ourtokens{} of \modelemph{Qwen3.5-9B} (top) and \modelemph{InternVL3.5-8B} (bottom) for the five neutral questions, each with the default (solid) and the alternative (hatched) answer instruction (\cref{tab:neutral_prompts}). The tokens are trained with the default prompt only. The darkest bars show the mean over the five questions for each instruction; error bars show the standard deviation over the questions. Black lines mark the bias toward the same cue without \ourtokens{}, averaged over the ten prompts.}
    \label{fig:token_prompt_transfer}
\end{figure*}

\cref{fig:steering_tokens} illustrates how the \ourtokens{} are injected into the frozen model (\cref{sec:beyond}), and \cref{tab:steering_hparams} summarizes the training setup. All settings are shared across both models and all target cues, except for a few \modelemph{InternVL3.5-8B} cues: we raise the on-cue weight $\lambda$ for the red, green, blue, and amplitude cues to keep the answers on the cues, and the learning rate for the background cue to let training converge.

\begin{table}[h!tb]
\centering
\caption{\textbf{Steering-token training setup.} Hyperparameters of the \ourtokens{} for both models and all targeted cues.}
\label{tab:steering_hparams}
\footnotesize
\begin{tabular}{@{}l>{\raggedright\arraybackslash}p{0.64\linewidth}@{}}
\toprule
\textbf{Setting} & \textbf{Value} \\
\midrule
\Ourtokens & $32$ tokens, injected in encoder and LM \\
LM tokens per image & $8$ ($2{\times}2$ merge) \\
Initialization & $\mathcal{N}(0, 0.02^2)$ \\
Objective & \cref{eq:steering_loss}, $\lambda{=}5$; InternVL3.5: $\lambda{=}200$ (red, blue), $50$ (green, amplitude) \\
Optimizer & AdamW, weight decay $0$, grad.\ clip $1.0$ \\
Learning rate & $10^{-3}$ (InternVL3.5 background: $10^{-2}$), $50$ warm-up steps, cosine decay \\
Batch size & $32$ \\
Max.\ updates & $10{,}000$ \\
Early stopping & target rate $\geq 0.98$ or 2 evals w/o $+0.01$ \\
Evaluation & every $1$k updates, $16$ held-out classes \\
Precision & frozen bf16 backbone \\
Seed & $42$ \\
\bottomrule
\end{tabular}
\end{table}

\subsection{Transfer to other prompts}\label{sec:appendix:steering_tokens_prompts}

The \ourtokens{} are trained with the default VQA prompt (\cref{tab:prompts}). To test whether they overfit to this prompt, we evaluate the tokens of \cref{tab:vlm_bias_full} with the ten neutral prompt formulations of \cref{sec:appendix:sens_prompt} (\cref{tab:neutral_prompts}), without retraining. The setup is otherwise identical to the main experiments (16 test classes, all 16 options in shuffled order, seed $42$). \cref{fig:token_prompt_transfer} shows the results.

The steering effect transfers to all prompts. For every token, the targeted bias under any of the nine other prompts stays within $5.6$ (\modelemph{Qwen3.5-9B}) and $6.7$ (\modelemph{InternVL3.5-8B}) pp of the default prompt. Where the tokens are effective, they remain so: the targeted bias stays between $91.8\%$ and $100\%$ for all spatial, frequency, and object/text cues of \modelemph{InternVL3.5-8B} and for all of these cues of \modelemph{Qwen3.5-9B} except the high-frequency cue, whose bias is $86.9$--$88.4\%$ for all prompts. The two largest drops occur with the ``Classify'' question: the text cue of \modelemph{InternVL3.5-8B} ($98.5\% \to 91.8\%$ with ``Classify'') and the left cue of \modelemph{Qwen3.5-9B} ($98.0\% \to 92.4\%$ with ``Classify'' and the alternative instruction). Where the tokens are ineffective, a different prompt does not help either: the biases toward the color cues vary by at most $2.3$ pp across the ten prompts, and the amplitude cue stays at or below $1.2\%$. The answers remain on the cues for all prompts: at most $5.0\%$ (\modelemph{Qwen3.5-9B}) and $10.2\%$ (\modelemph{InternVL3.5-8B}, amplitude token) of the answers name a class that is not shown in the stimulus.

\FloatBarrier

\begin{table*}[t]
\centering
\caption{\textbf{Prediction bias of all models.} Extended version of \cref{tab:vlm_bias} with all evaluated models. Bias (\%, \cref{eq:bias}) toward each cue under neutral prompting, and toward the targeted cue under explicit cue targeting, in-context learning, and \ourtokens. Numbers in parentheses give the change in pp over the same model's neutral bias for that cue. \modelemph{Gemma-4-E4B} uses the default budget of $1{,}120$ visual tokens per image; the indented rows below it limit this budget (\cref{sec:appendix:token_budget}).}
\label{tab:vlm_bias_full}
\showallmodelstrue

\end{table*}

\FloatBarrier

\clearpage
\onecolumn
\begingroup
\scriptsize
\setlength{\tabcolsep}{1.5pt}
\setlength{\LTcapwidth}{\textwidth}
\renewcommand{\arraystretch}{1.0}
\begin{longtable}{@{}>{\bfseries}p{1.35cm} >{\raggedright\arraybackslash}p{0.73\linewidth} *{10}{c}@{}}
\caption{\textbf{Selected steering prompts.} For each conflict and targeted cue, the prompt prefixes selected by the prompt search on the search split (\cref{sec:appendix:prompts:optimization}). A bullet marks the models for which a prefix was selected, \ie, the prompts that yield the ``Explicit cue targeting'' biases in \cref{tab:vlm_bias_full}. Each prefix is followed by the enumerated options and ``\texttt{Answer only with the correct option letter.}''}\label{tab:prompt_steering_prompts} \\
\toprule
& & \multicolumn{10}{c}{\textbf{Model}} \\
\cmidrule(l){3-12}
\textbf{Cue} & \textbf{Prompt prefix} & \rotatebox{90}{Gemma-4-E4B} & \rotatebox{90}{Gemma-4-12B} & \rotatebox{90}{InternVL3.5-8B} & \rotatebox{90}{Qwen3.5-0.8B} & \rotatebox{90}{Qwen3.5-2B} & \rotatebox{90}{Qwen3.5-4B} & \rotatebox{90}{Qwen3.5-9B} & \rotatebox{90}{Qwen3.5-27B} & \rotatebox{90}{Qwen3.6-27B} & \rotatebox{90}{Qwen3.8-27B} \\
\midrule
\endfirsthead
\caption[]{\textbf{Selected steering prompts} (continued).} \\
\toprule
& & \multicolumn{10}{c}{\textbf{Model}} \\
\cmidrule(l){3-12}
\textbf{Cue} & \textbf{Prompt prefix} & \rotatebox{90}{Gemma-4-E4B} & \rotatebox{90}{Gemma-4-12B} & \rotatebox{90}{InternVL3.5-8B} & \rotatebox{90}{Qwen3.5-0.8B} & \rotatebox{90}{Qwen3.5-2B} & \rotatebox{90}{Qwen3.5-4B} & \rotatebox{90}{Qwen3.5-9B} & \rotatebox{90}{Qwen3.5-27B} & \rotatebox{90}{Qwen3.6-27B} & \rotatebox{90}{Qwen3.8-27B} \\
\midrule
\endhead
\midrule
\multicolumn{12}{r}{\textit{Continued on next page}} \\
\endfoot
\bottomrule
\endlastfoot
\multicolumn{12}{@{}l}{\textit{Left/right}} \\*
Left & ``Which option best describes the left half of the image? Options:'' & $\bullet$ & $\bullet$ & $\bullet$ & $\bullet$ & $\bullet$ & $\bullet$ & $\bullet$ & $\bullet$ & $\bullet$ & $\bullet$ \\
\cmidrule(l){1-12}
Right & ``Which option best describes the right half of the image? Options:'' & $\bullet$ & $\bullet$ & $\bullet$ & $\bullet$ & $\bullet$ & $\bullet$ & $\bullet$ & $\bullet$ & $\bullet$ & $\bullet$ \\
\midrule
\multicolumn{12}{@{}l}{\textit{Top/bottom}} \\*
Top & ``Which option best describes the top half of the image? Options:'' & $\bullet$ & $\bullet$ & $\bullet$ & $\bullet$ & $\bullet$ & $\bullet$ & $\bullet$ & $\bullet$ & $\bullet$ & $\bullet$ \\
\cmidrule(l){1-12}
Bottom & ``Which option best describes the bottom half of the image? Options:'' & $\bullet$ & $\bullet$ & $\bullet$ & $\bullet$ & $\bullet$ & $\bullet$ & $\bullet$ & $\bullet$ & $\bullet$ & $\bullet$ \\
\midrule
\multicolumn{12}{@{}l}{\textit{Foreground/background}} \\*
Foreground & ``Which option best describes the image in the foreground? Options:'' &  &  &  & $\bullet$ & $\bullet$ & $\bullet$ &  &  &  &  \\*
 & ``Which option best describes the image? Options:'' & $\bullet$ &  & $\bullet$ &  &  &  &  &  &  &  \\*
 & ``Treat the image as two overlapping layers: which option names the object on top of the other?'' &  & $\bullet$ &  &  &  &  &  &  &  &  \\*
 & ``At the overlap, use the visible boundary to identify which object lies on top and hides part of the other; which option names that foreground object?'' &  &  &  &  &  &  & $\bullet$ &  &  &  \\*
 & ``Which option names the object whose outline stays in front at the overlap, hiding parts of the other object?'' &  &  &  &  &  &  &  & $\bullet$ &  &  \\*
 & ``Ignore which object is more recognizable; at their overlap, which option names the object whose visible surface lies on top?'' &  &  &  &  &  &  &  &  & $\bullet$ &  \\*
 & ``At the overlap, which option matches the object on the visible surface layer, covering part of the other object? Options:'' &  &  &  &  &  &  &  &  &  & $\bullet$ \\
\cmidrule(l){1-12}
Background & ``Which option best describes the image in the background? Options:'' &  &  &  & $\bullet$ & $\bullet$ &  &  &  &  &  \\*
 & ``Select the object whose visible parts lie behind and are partly obscured by the nearer object: which option names it?'' & $\bullet$ &  &  &  &  &  &  &  &  &  \\*
 & ``Which option identifies the object that continues behind the overlapping object, using its exposed parts as evidence rather than the object covering it? Options:'' &  & $\bullet$ &  &  &  &  &  &  &  &  \\*
 & ``Trace the visible portions to identify the object extending behind the overlap; which class is it? Options:'' &  &  & $\bullet$ &  &  &  &  &  &  &  \\*
 & ``Which listed class matches the object whose visible outline disappears beneath the overlapping object?'' &  &  &  &  &  & $\bullet$ &  &  &  &  \\*
 & ``Trace the object whose visible parts continue around the edges of the occluder; which option names that background object?'' &  &  &  &  &  &  & $\bullet$ &  &  &  \\*
 & ``Which option names the object whose outline is interrupted at the overlap, showing that it continues behind the other object?'' &  &  &  &  &  &  &  & $\bullet$ &  &  \\*
 & ``At the overlap, identify the object whose outline is interrupted by the other object; which option names that object? Options:'' &  &  &  &  &  &  &  &  & $\bullet$ &  \\*
 & ``Which option names the object with parts concealed by the overlapping object, indicating it is in the background? Options:'' &  &  &  &  &  &  &  &  &  & $\bullet$ \\
\midrule
\multicolumn{12}{@{}l}{\textit{Chromatic}} \\*
Red & ``Which option best describes the image? Options:'' &  &  &  & $\bullet$ & $\bullet$ &  &  & $\bullet$ & $\bullet$ &  \\*
 & ``Use the red-channel intensity pattern as evidence and match its encoded silhouette to a class, disregarding shapes indicated by green or blue values: which class is it?'' & $\bullet$ &  &  &  &  &  &  &  &  &  \\*
 & ``Use only the spatial pattern of red-channel brightness as visual evidence; which listed class does that pattern depict?'' &  & $\bullet$ &  &  &  &  &  &  &  &  \\*
 & ``Follow the object-shaped edges encoded by red-channel intensity, using its outline and parts to identify that object rather than either green- or blue-channel object; which option names it? Options:'' &  &  & $\bullet$ &  &  &  &  &  &  &  \\*
 & ``Which listed class matches the object silhouette revealed by the image's red-channel intensities, rather than the green- or blue-channel silhouettes?'' &  &  &  &  &  & $\bullet$ &  &  &  &  \\*
 & ``Which listed class matches the object represented by the red-channel signal, based on its shape while ignoring the green- and blue-channel objects?'' &  &  &  &  &  &  & $\bullet$ &  &  &  \\*
 & ``Which listed class is depicted by the shape traced in red-channel intensity, treating the green- and blue-channel shapes as distractors?'' &  &  &  &  &  &  &  &  &  & $\bullet$ \\
\cmidrule(l){1-12}
Green & ``Which option best describes the image? Options:'' & $\bullet$ &  & $\bullet$ & $\bullet$ & $\bullet$ & $\bullet$ & $\bullet$ & $\bullet$ & $\bullet$ & $\bullet$ \\*
 & ``If the image were rendered in grayscale using only green-channel intensity, which listed class would the visible object be?'' &  & $\bullet$ &  &  &  &  &  &  &  &  \\
\cmidrule(l){1-12}
Blue & ``Which listed class matches the object traced by variations in blue intensity across the image, with red and green intensity ignored?'' & $\bullet$ &  &  &  &  &  &  &  &  &  \\*
 & ``Conceptually render only each pixel's blue-channel intensity as a grayscale image; which listed object's silhouette does it reveal?'' &  & $\bullet$ &  &  &  &  &  &  &  &  \\*
 & ``Treat the image as three overlaid RGB signals: which class matches the object silhouette carried by blue intensity (the B channel), excluding silhouettes carried by red or green?'' &  &  & $\bullet$ &  &  &  &  &  &  &  \\*
 & ``Treat blue-channel intensity as a mask: which option names the object whose shape it reveals?'' &  &  &  & $\bullet$ &  &  &  &  &  &  \\*
 & ``Mentally view only the blue channel, trace the object outline it reveals, and identify its class; which listed class is it?'' &  &  &  &  & $\bullet$ &  &  &  &  &  \\*
 & ``Mentally split the image into its red, green, and blue components, inspect only the blue component's object, and classify it by its distinctive parts: which listed class is it?'' &  &  &  &  &  & $\bullet$ &  &  &  &  \\*
 & ``Inspect the image's blue-channel signal and identify the object whose shape it reveals, disregarding shapes carried by the red and green channels; which option names that object?'' &  &  &  &  &  &  & $\bullet$ &  &  &  \\*
 & ``After mentally filtering out the red and green content, which object is identifiable from the remaining blue-channel shape?'' &  &  &  &  &  &  &  & $\bullet$ &  &  \\*
 & ``Attend specifically to the blue pixels, trace the shape they form, and identify which object that shape depicts. Options:'' &  &  &  &  &  &  &  &  & $\bullet$ &  \\*
 & ``Identify the object rendered in blue, ignoring the red and green objects; which class label matches it? Options:'' &  &  &  &  &  &  &  &  &  & $\bullet$ \\
\midrule
\multicolumn{12}{@{}l}{\textit{Low/high-frequency}} \\*
Low freq. & ``Which option best describes the lowest frequencies of the image? Options:'' &  &  &  &  &  & $\bullet$ & $\bullet$ &  &  &  \\*
 & ``When viewed from far away, which listed object is identifiable from the image's largest light-and-dark regions and overall mass, rather than its crisp details?'' & $\bullet$ &  &  &  &  &  &  &  &  &  \\*
 & ``Which listed object is indicated by the image's broad light-and-dark regions and large-scale tonal pattern, ignoring fine detail?'' &  & $\bullet$ &  &  &  &  &  &  &  &  \\*
 & ``Which option names the object shown by the image's broad, coarse shapes, ignoring fine edges and details?'' &  &  & $\bullet$ &  &  &  &  &  &  &  \\*
 & ``Which class is recognizable from the image's broad, blurred shape when fine edges and texture are ignored? Options:'' &  &  &  & $\bullet$ &  &  &  &  &  &  \\*
 & ``Which listed object is represented by the image's coarse, blurred structure and broad color regions, ignoring fine edges, textures, and small details?'' &  &  &  &  & $\bullet$ &  &  &  &  &  \\*
 & ``In a spatial-frequency decomposition, which listed object is encoded by the image's coarse, long-wavelength intensity pattern rather than its rapid local changes? Options:'' &  &  &  &  &  &  &  & $\bullet$ &  &  \\*
 & ``Which option names the object represented by the image's large-scale light-and-dark variations, not its fine-scale patterns?'' &  &  &  &  &  &  &  &  & $\bullet$ &  \\*
 & ``Which option names the object indicated by the image's broad, coarse visual structure, ignoring fine details and sharp edges?'' &  &  &  &  &  &  &  &  &  & $\bullet$ \\
\cmidrule(l){1-12}
High freq. & ``Which listed class is represented by the image's finest, sharpest details, rather than its broad overall shapes?'' & $\bullet$ &  &  &  &  &  &  &  &  &  \\*
 & ``Which listed class matches the object encoded in rapid pixel-level intensity changes and fine texture, ignoring the slowly varying broad image content?'' &  & $\bullet$ &  &  &  &  &  &  &  &  \\*
 & ``Which listed object is represented by the image's finest edges and tiny details, rather than its broad, smooth structure? Options:'' &  &  & $\bullet$ &  &  &  &  &  &  &  \\*
 & ``Which option identifies the object revealed by the image's fine edges and small-scale details, rather than its broad, blurry shapes?'' &  &  &  & $\bullet$ &  &  &  &  &  &  \\*
 & ``Mentally suppress the broad, slowly changing shapes and colors; which listed object is identified by the rapid image variations that remain?'' &  &  &  &  & $\bullet$ &  &  &  &  &  \\*
 & ``Which listed object is represented by the image's high-spatial-frequency content alone, considering fine rapid variations while disregarding coarse low-frequency structure?'' &  &  &  &  &  & $\bullet$ &  &  &  &  \\*
 & ``Which option names the object revealed by the finest-scale edges and texture, rather than by the image's coarse silhouette and broad shading?'' &  &  &  &  &  &  & $\bullet$ &  &  &  \\*
 & ``Mentally suppress the broad, smooth shapes and identify the object depicted by the image's crisp, densely detailed features. Options:'' &  &  &  &  &  &  &  & $\bullet$ &  &  \\*
 & ``Which option identifies the object conveyed by the image's finest edge and texture details, rather than its broad shapes or smooth regions? Options:'' &  &  &  &  &  &  &  &  & $\bullet$ &  \\*
 & ``Which object is represented by the finest details and sharp edges in the image, rather than its broad shapes or silhouette? Choose the matching class:'' &  &  &  &  &  &  &  &  &  & $\bullet$ \\
\midrule
\multicolumn{12}{@{}l}{\textit{Amplitude/phase}} \\*
Amplitude & ``Which option names the object indicated by the image's overall colors and texture, rather than its recognizable shape or arrangement?'' & $\bullet$ &  &  &  &  &  &  &  &  &  \\*
 & ``Treat the image as a texture sample rather than a recognizable object: which listed object best matches its overall color-and-texture signature, regardless of outline or spatial layout?'' &  & $\bullet$ &  &  &  &  &  &  &  &  \\*
 & ``If spatial phase were randomized while the frequency magnitudes stayed fixed, which listed object's characteristic visual texture would remain?'' &  &  & $\bullet$ &  &  &  &  &  &  &  \\*
 & ``Treat the Fourier magnitude spectrum as a map of spatial-frequency strengths: which listed object is represented by that strength pattern, regardless of phase?'' &  &  &  & $\bullet$ &  &  &  &  &  &  \\*
 & ``Which option names the source whose information is retained when each Fourier frequency keeps its strength but its phase is randomized?'' &  &  &  &  & $\bullet$ &  &  &  &  &  \\*
 & ``Imagine the image's spatial layout scrambled while its overall frequency energy and contrast are preserved: which listed object would the resulting texture most resemble?'' &  &  &  &  &  & $\bullet$ &  &  &  &  \\*
 & ``Which listed object supplies the image's overall texture, color, and spatial-frequency energy---the amplitude source, not the object recognized from its shape and arrangement? Options:'' &  &  &  &  &  &  & $\bullet$ &  &  &  \\*
 & ``Imagine keeping only the strength of each Fourier frequency and discarding its phase; which listed object would the amplitude-only image depict?'' &  &  &  &  &  &  &  & $\bullet$ &  &  \\*
 & ``If the image's Fourier phases were randomized while its amplitude values were preserved, which listed object class would the resulting texture most resemble?'' &  &  &  &  &  &  &  &  & $\bullet$ &  \\*
 & ``Which listed object matches the image's overall texture and balance of coarse versus fine visual detail, disregarding the identity suggested by its contours?'' &  &  &  &  &  &  &  &  &  & $\bullet$ \\
\cmidrule(l){1-12}
Phase & ``Which option best describes the image? Options:'' &  &  &  & $\bullet$ & $\bullet$ & $\bullet$ &  & $\bullet$ & $\bullet$ & $\bullet$ \\*
 & ``Which option best describes the phase spectrum of the image? Options:'' & $\bullet$ & $\bullet$ & $\bullet$ &  &  &  & $\bullet$ &  &  &  \\
\midrule
\multicolumn{12}{@{}l}{\textit{Object/text}} \\*
Object & ``Which option best describes the object in the image? Options:'' & $\bullet$ & $\bullet$ &  &  &  &  & $\bullet$ & $\bullet$ & $\bullet$ & $\bullet$ \\*
 & ``Which option names the object shown by its visible shape and appearance, rather than any word written over the image?'' &  &  & $\bullet$ &  &  &  &  &  &  &  \\*
 & ``Which class does the depicted object belong to, based on its contours and component parts rather than the overlaid lettering?'' &  &  &  & $\bullet$ &  &  &  &  &  &  \\*
 & ``Using the object's visible shape and physical features rather than any words in the image, which option names the object actually depicted?'' &  &  &  &  & $\bullet$ &  &  &  &  &  \\*
 & ``Which option names the object identified by its visible shape and features, not by any text overlaid on the image? Options:'' &  &  &  &  &  & $\bullet$ &  &  &  &  \\
\cmidrule(l){1-12}
Text & ``Which option best describes the text in the image? Options:'' &  & $\bullet$ & $\bullet$ & $\bullet$ & $\bullet$ & $\bullet$ & $\bullet$ & $\bullet$ & $\bullet$ & $\bullet$ \\*
 & ``Which listed class name is spelled by the visible lettering over the picture, regardless of the object shown? Options:'' & $\bullet$ &  &  &  &  &  &  &  &  &  \\
\end{longtable}
\endgroup

\end{document}